\documentclass[Afour,sageh,times]{sagej}
\usepackage{moreverb,url}
\usepackage[colorlinks,bookmarksopen,bookmarksnumbered,citecolor=red,urlcolor=red]{hyperref}
\usepackage{amsmath,amssymb,amsfonts}
\usepackage{algorithmic}
\usepackage{graphicx}
\usepackage{textcomp}
\usepackage{xcolor}
\usepackage{booktabs}
\usepackage[weather]{ifsym}
\usepackage[dvipsnames]{xcolor}
\usepackage{moreverb,url}
\usepackage{amsmath,amsfonts,amssymb}
\usepackage{tikz,tkz-euclide}
\usepackage{balance}
\usepackage{siunitx}
\usetikzlibrary{decorations,calligraphy}
\usetikzlibrary{calc}
\tikzset{>=latex}
\usepackage{tabularx}
\newcolumntype{Y}{>{\centering\arraybackslash}X}
\newcolumntype{L}{>{\arraybackslash}X}

\usepackage{multirow}

\usepackage[colorlinks,bookmarksopen,bookmarksnumbered,citecolor=red,urlcolor=red]{hyperref}
\usepackage[nolist,nohyperlinks]{acronym}

\newcommand\edit[1]{{#1}}

\def\subfigstyle{\footnotesize}

\def\m{m}

\def\reals{\mathbb{R}}

\def\occ{o}

\def\dist{d}
\def\revert{r}

\def\gp{\mathcal{GP}}

\newcommand{\kernelraw}[1]{{{k}_{#1}}}
\newcommand{\kernel}[3]{{{k}_{#1}\left(#2,#3\right)}}
\newcommand{\kernelmat}[3]{{\mathbf{K}_{#2,#3}}}

\newcommand{\kernelvec}[3]{{\mathbf{k}_{#2,#3}}}

\renewcommand\time[1]{{t_{#1}}}
\newcommand\frametime[1]{{\tau_{#1}}}

\newcommand\pointtime[1]{{\mathfrak{t}_{#1}}}
\newcommand\refframe[2]{{\mathfrak{F}_{#1}^{#2}}}

\def\Tc{{\mathbf{T}_I^L}}

\def\state{{\mathcal{S}}}

\newcommand\trans[2]{{\mathbf{T}_{#1}^{#2}}}
\newcommand\transtilde[2]{{\tilde{\mathbf{T}}_{#1}^{#2}}}

\newcommand\rot[2]{{\mathbf{R}_{#1}^{#2}}}

\newcommand\pos[2]{{\mathbf{p}_{#1}^{#2}}}

\newcommand\vel[2]{{\mathbf{v}_{#1}^{#2}}}

\def\imu{{I}}

\def\world{{W}}
\def\lidar{{L}}
\def\map{\mathcal{M}}

\def\gravity{{\mathbf{g}}}

\def\minus{\text{-}}
\def\plus{\text{+}}

\def\acc{{f}}
\newcommand\properacc[1]{{\mathbf{\acc}_{#1}}}

\def\gyr{{\omega}}

\newcommand\biasaccbar[1]{{\bar{\mathbf{b}}_{\acc}^{#1}}}
\newcommand\biasgyrbar[1]{{\bar{\mathbf{b}}_{\gyr}^{#1}}}
\newcommand\biasacc[1]{{\mathbf{b}_{\acc}^{#1}}}
\newcommand\biasgyr[1]{{\mathbf{b}_{\gyr}^{#1}}}

\newcommand\dpos[2]{{\Delta\mathbf{p}_{#1}^{#2}}}
\newcommand\dvel[2]{{\Delta\mathbf{v}_{#1}^{#2}}}
\newcommand\drot[2]{{\Delta\mathbf{R}_{#1}^{#2}}}
\newcommand\dposbar[2]{{\bar{\Delta\mathbf{p}}_{#1}^{#2}}}
\newcommand\dvelbar[2]{{\bar{\Delta\mathbf{v}}_{#1}^{#2}}}
\newcommand\drotbar[2]{{\bar{\Delta\mathbf{R}}_{#1}^{#2}}}

\def\Exp{\text{Exp}}
\def\Log{\text{Log}}
\def\planar{\mathcal{P}}
\def\edge{\mathcal{E}}
\def\neighbours{{\mathcal{A}}}
\newcommand\residualplanar[1]{{r_{\planar_j}}}
\newcommand\residualedge[1]{{r_{\edge_j}}}
\newcommand\residualbias[1]{{\mathbf{r}_{\biasacc{}}}}
\newcommand\residualimu[1]{{\mathbf{r}_{\imu_{#1}}}}

\def\coordinate{{\mathbf{e}}}

\newcommand{\pointcloud}[2]{{\mathcal{U}_{#1}^{#2}}}

\newcommand\lidarpoint[2]{{\mathbf{x}_{#1}^{#2}}}
\newcommand\lidarpoints[2]{{\mathbf{X}_{#1}^{#2}}}

\def\spaceindex{{\mathcal{I}}}
\def\voxelmap{{\mathcal{V}}}
\def\voxel{{\mathbf{v}}}
\def\spoint{{\mathbf{s}}}

\begin{acronym}[AAAAAAAAA]
    \acro{ate}[ATE]{Absolute Trajectory Error}
    \acro{dof}[DoF]{Degree-of-Freedom}
    \acro{fmcw}[FMCW]{Frequency Modulated Continuous Wave}
    \acro{fov}[FoV]{Field-of-View}
    \acro{gp}[GP]{Gaussian Process}
    \acrodefplural{gp}[GPs]{Gaussian Processes}
    \acro{icp}[ICP]{Iterative Closest Point}
    \acro{iid}[i.i.d.]{independent and identically distributed}
    \acro{imu}[IMU]{Inertial Measurement Unit}
    \acro{lpm}[LPM]{Linear Preintegrated Measurement}
    \acro{rmse}[RMSE]{Root Mean Squared Error}
    \acro{rpe}[RPE]{Relative Pose Error}
    \acro{slam}[SLAM]{Simultaneous Localisation And Mapping}
    \acro{tsdf}[TSDF]{Truncated Signed Distance Field}
\end{acronym}

\newcommand\BibTeX{{\rmfamily B\kern-.05em \textsc{i\kern-.025em b}\kern-.08em
T\kern-.1667em\lower.7ex\hbox{E}\kern-.125emX}}
\begin{document}

\runninghead{Le Gentil et al.}

\title{2Fast-2Lamaa: Large-Scale Lidar-Inertial Localization and Mapping with Continuous Distance Fields}

\author{Cedric Le Gentil\affilnum{1}, Raphael Falque\affilnum{2}, Daniil Lisus\affilnum{1}, and Timothy D. Barfoot\affilnum{1}}

\affiliation{\affilnum{1}Robotics Institute, University of Toronto Institute for Aerospace Studies, Canada\\
\affilnum{2}Robotics Institute, University of Technology Sydney, Australia}

\corrauth{Cedric Le Gentil, Robotics Institute
University of Toronto Institute for Aerospace Studies,
North York, Ontario
M3H~5T6, Canada.}

\email{cedric.legentil@robotics.utias.utoronto.ca}

\begin{abstract}
This paper introduces 2Fast-2Lamaa, a lidar-inertial state estimation framework for odometry, mapping, and localization.
Its first key component is the optimization-based undistortion of lidar scans, which uses continuous IMU preintegration to model the system's pose at every lidar point timestamp.
The continuous trajectory over 100-200ms is parameterized only by the initial scan conditions (linear velocity and gravity orientation) and IMU biases, yielding eleven state variables.
These are estimated by minimizing point-to-line and point-to-plane distances between lidar-extracted features without relying on previous estimates, resulting in a prior-less motion-distortion correction strategy.
Because the method performs local state estimation, it directly provides scan-to-scan odometry.
To maintain geometric consistency over longer periods, undistorted scans are used for scan-to-map registration.
The map representation employs Gaussian Processes to form a continuous distance field, enabling point-to-surface distance queries anywhere in space.
Poses of the undistorted scans are refined by minimizing these distances through non-linear least-squares optimization.
For odometry and mapping, the map is built incrementally in real time; for pure localization, existing maps are reused.
The incremental map construction also includes mechanisms for removing dynamic objects.
We benchmark 2Fast-2Lamaa on over 750km of public and self-collected datasets from both automotive and handheld systems.
The framework achieves state-of-the-art performance across diverse and challenging scenarios, reaching odometry and localization errors as low as 0.22\% and 0.06 m, respectively.
The real-time implementation is publicly available at \url{https://github.com/clegenti/2fast2lamaa}.

\end{abstract}


\maketitle

\section{Introduction}

\begin{figure*}
    \centering
    \def\imgheight{4.4cm}
    \def\legenddist{0.1cm}
    \def\legendtextsize{\small}
    \def\hdist{0.05cm}
    \begin{tikzpicture}
        \node[draw = red, inner sep = 0, outer sep = 0 ](zoom){\includegraphics[clip, height=\imgheight]{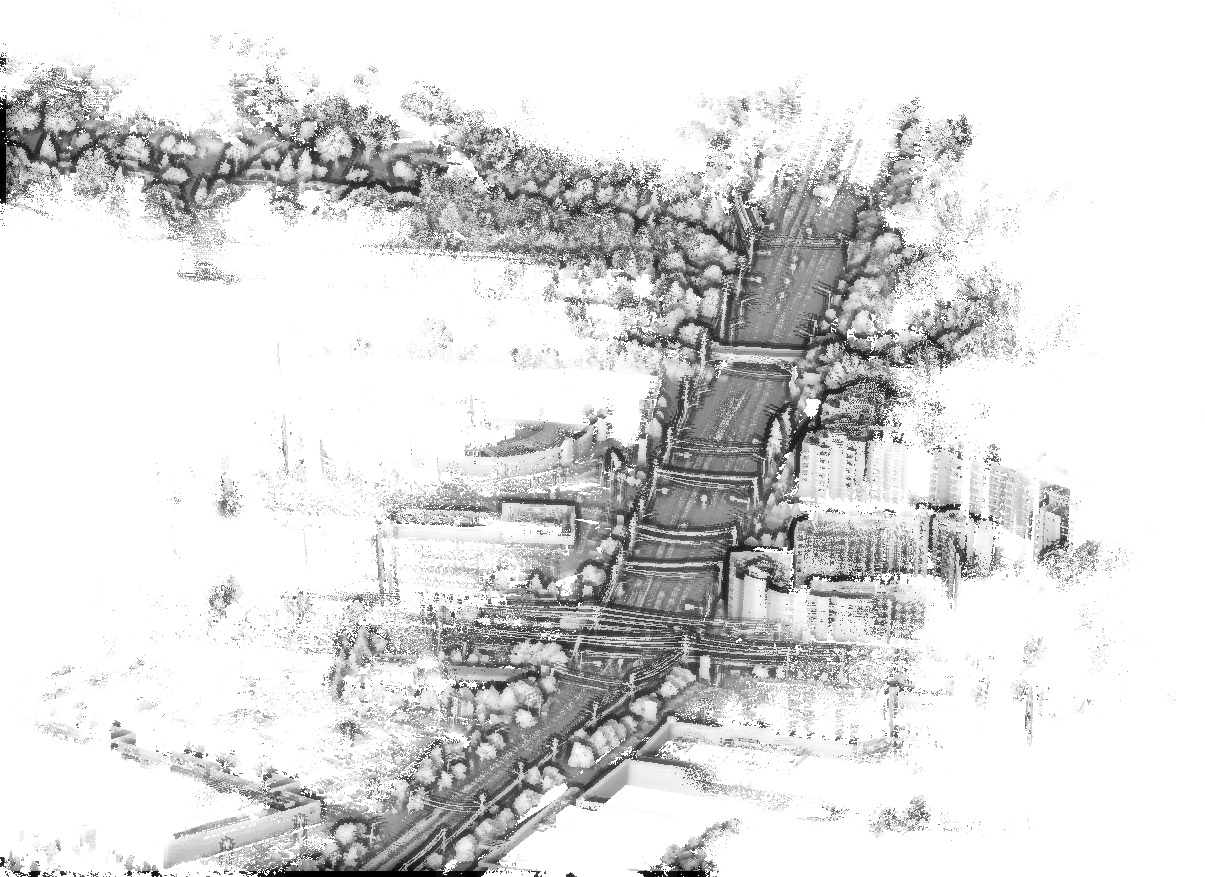}};
        \node[inner sep = 0, outer sep = 0, left=\hdist of zoom ](global){\includegraphics[clip, height=\imgheight]{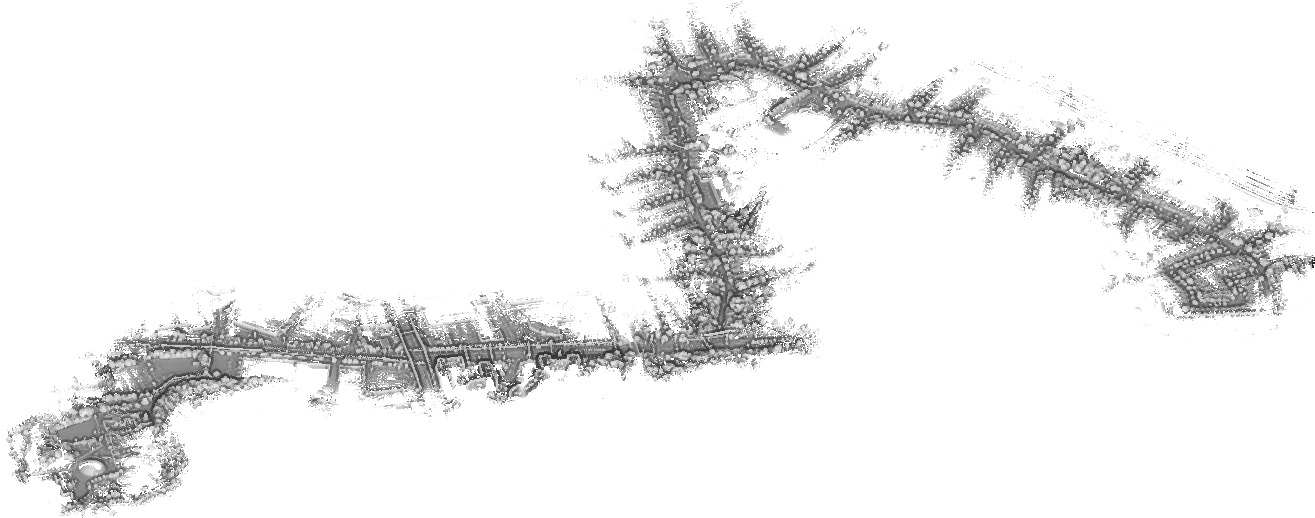}};
        \node[draw = red, inner sep = 0, outer sep = 0, minimum width=2.5cm, minimum height=1.5cm, above=0.75cm of global.south, xshift=0.4cm]{};
        \node[below = \legenddist of zoom] {(b) Zoom};
        \node[below = \legenddist of global] {(a) 10M-point map of a \SI{8}{\km} trajectory};
    \end{tikzpicture}
    \caption{2Fast-2Lamaa performs odometry, mapping, and localization over large-scale environments. It relies on GP-based distance fields for scan-to-map registration. Thanks to efficient data structures, the map can contain details of the environment's geometry while allowing large-scale operations. The images here are visualizations of the map created with an \SI{8}{\km}-long \emph{Suburbs} sequence from the Boreas-RT dataset. Despite the length of the trajectory, the map can represent all the observed geometry (a), without sacrificing details (b).}
    \label{fig:teaser}
\end{figure*}

Over the past decades, the robotics community has put a lot of effort into solving odometry, that is, estimating a system's ego-motion in unknown environments.
In our previous work \citep{legentil2025dowe}, we showed that modern sensing capabilities provide highly accurate odometry even in challenging automotive scenarios.
For the majority of real-world robot deployments, odometry alone is not sufficient to estimate the system’s state.
With the example of self-driving vehicles or autonomous conveying, the robot is expected to navigate between known locations in a previously mapped environment.
In this context, the critical enabling component for autonomous operations is localization, not odometry.
The latter is simply used as an initial guess/prior between localization steps.
In this paper, we present \textbf{2Fast-2Lamaa}, which stands for \textbf{F}ast \textbf{F}ield-based \textbf{A}gent-\textbf{s}ubtracted \textbf{T}ruly coupled \textbf{L}idar \textbf{L}ocalization \textbf{a}nd \textbf{M}apping with \textbf{A}ccelerometer and \textbf{A}ngular-rate.
This convoluted acronym refers to a lidar-inertial framework that addresses both mapping and localization by tightly coupling lidar and \ac{imu} measurements and leveraging continuous distance fields to represent the environment.

The first step of most lidar-based systems is motion-distortion correction.
Commonly used lidars do not capture instantaneous snapshots of the environment; instead, they sweep one or more laser beams through the scene to collect 3D scans.
Accordingly, any motion of the sensor during a scan's duration creates motion distortion in the data.
Many lidar-based state estimation algorithms leverage data from an \ac{imu} to address this issue \citep{lee2024lidar}.
The most naive approach consists of using the latest pose and velocity estimates and integrating the inertial measurements to approximate the system's trajectory for the duration of the incoming scan.
With that movement prediction, the scan can be undistorted before being used in a non-linear optimization for scan-to-scan or scan-to-map rigid registration, alongside inertial constraints between consecutive scans.
While this approach has been qualified as `tightly-coupled' lidar-inertial estimation, for example, by \cite{Ye2019}, this undistortion strategy is a `one-off' operation that can be thought of as an `open-loop' process.
Because it freezes the scans based on prior information, it decouples the problem of lidar data undistortion from the estimation of scan-to-scan motion.
This can lead to unrecoverable errors if the prior state estimate is not accurate enough due to accumulated drift or outlier data.
Other approaches to lidar-inertial state estimation consider the lidar points individually through some continuous motion representation.
There, the \ac{imu} measurements can be used as residuals to constrain the continuous state \citep{talbot2025continuous} or directly used to locally parameterize the trajectory \citep{LeGentil2018}.
These methods estimate the system's trajectory at the same time as correcting motion distortion in a \emph{truly coupled} manner.
Building upon our previous work \citep{legentil2024undistortion}, 2Fast-2Lamaa falls into the latter category by fully characterizing the trajectory during a scan based on \ac{imu} measurements without the need for any explicit motion model.

Once a scan is undistorted, 2Fast-2Lamaa performs localization by estimating a rigid transformation through scan-to-map registration.
Note that when used for odometry or mapping, 2Fast-2Lamaa builds the map incrementally, whereas it leverages a previously built map when performing pure localization.
A key component of a localization framework is the choice of map representation.
The majority of lidar-based state estimation frameworks model the environment with point clouds (with or without normal vectors).
In such a case, scan registration is generally performed with a variant of the \ac{icp} algorithm \citep{besl1992icp}.
2Fast-2Lamaa differs from this paradigm as it leverages distance fields.
These are continuous functions that can be queried at any location and that return an approximation of the Euclidean distance to the closest object/surface in the environment.
Accordingly, the registration process consists of directly minimizing distance queries in a non-linear least-squares formulation, neither requiring explicit geometric primitives to model the environment nor any data association steps.
The distance field presented in this work is based on \ac{gp} regression \citep{Rasmussen2006}, similarly to \cite{legentil2024accurate}.
While the concept of a \ac{gp}-based distance field is not new, 2Fast-2Lamaa is the first real-time state estimation framework that successfully builds upon this idea for large-scale operations (sequences over \SI{10}{\km}-long), as illustrated in Fig.~\ref{fig:teaser}.
Note that this integration is not trivial, as standard \ac{gp} regression suffers from cubic computational complexity.
By leveraging efficient data structures and computing \acp{gp} locally, 2Fast-2Lamaa's novel distance field approximation displays a $\log(N)$ computational complexity.
The efficient data structures also enable real-time dynamic object removal through simple ray tracing.

Regardless of the choice of geometric representation (point cloud, mesh, distance field, etc.), mapping and localization frameworks can opt for different high-level mapping strategies. 
The most natural option is to use a single globally consistent map.
However, it has been demonstrated that accurate localization does not require geometrically consistent global maps \citep{brooks1987teachandrepeat,baumgartner1994teachandrepeat}.
Using a topometric approach that only requires local geometric consistency, \cite{furgale2010vtr} performs navigation by moving in a set of topologically connected submaps, alleviating the need for perfect odometry/trajectory estimation when building the map. 
This approach is especially suited for \emph{Teach and Repeat} tasks, where the robot is manually piloted to map the environment (teach/mapping) before performing autonomous path tracking to follow the original trajectory (repeat/localization).
Our proposed framework can operate with either a globally consistent map or in a topometric manner, utilizing a succession of submaps to localize the robot's repeating trajectories.

As any odometry pipeline is bound to drift, building maps purely on odometry estimates lead to inconsistencies at the global scale when considering trajectories that include loops that revisit previously explored areas.
While not the core focus of the present work, 2Fast-2Lamaa can also be used for globally consistent trajectory estimation by performing loop-closure detection and batch pose-graph optimization, similarly to \ac{slam} frameworks.
This process is performed \edit{online} when running odometry with submaps (topometric mapping mode).
The proposed loop-closure detection and correction pipeline relies on the projection of each submap into 2D image-like data structures that represent the environment's elevation changes.
Similar to the work by~\cite{giubilato2022gpgm}, visual features are extracted from these image-like structures, matched to detect loop-closures, and then used to provide the associated rough SE(2) geometric transformation between submaps.
After SE(3) submap-to-submap pose refinement with the aforementioned \ac{gp}-based distance fields, the global pose of each submap is estimated in a batch pose-graph optimization.

To summarise, our contributions are:
\begin{itemize}
    \item The integration of truly coupled lidar-inertial motion-distortion correction in a localization and mapping framework named 2Fast-2Lamaa.
    \item The development of an efficient \ac{gp}-based distance field for large-scale online mapping in the presence of dynamic objects.
    \item \edit{The integration of an online} loop-closure detection and correction mechanism to extend 2Fast-2Lamaa's capabilities beyond odometry and incremental mapping.
    \item An extensive odometry and localization evaluation of 2Fast-2Lamaa with more than \SI{750}{\km} of automotive and handheld data.
    \item An open-source real-time implementation.
\end{itemize}

\section{Related work}

\subsection{Lidar state estimation}

\begin{figure*}
    \centering
    \def\scale{0.98}
    \def\legenddist{0.1cm}
    \def\legendtextsize{\small}
    \def\xoffset{6.3cm}
    \begin{tikzpicture}
        \node[inner sep = 0, outer sep = 0 ](diagram){\includegraphics[clip, width=\scale\linewidth]{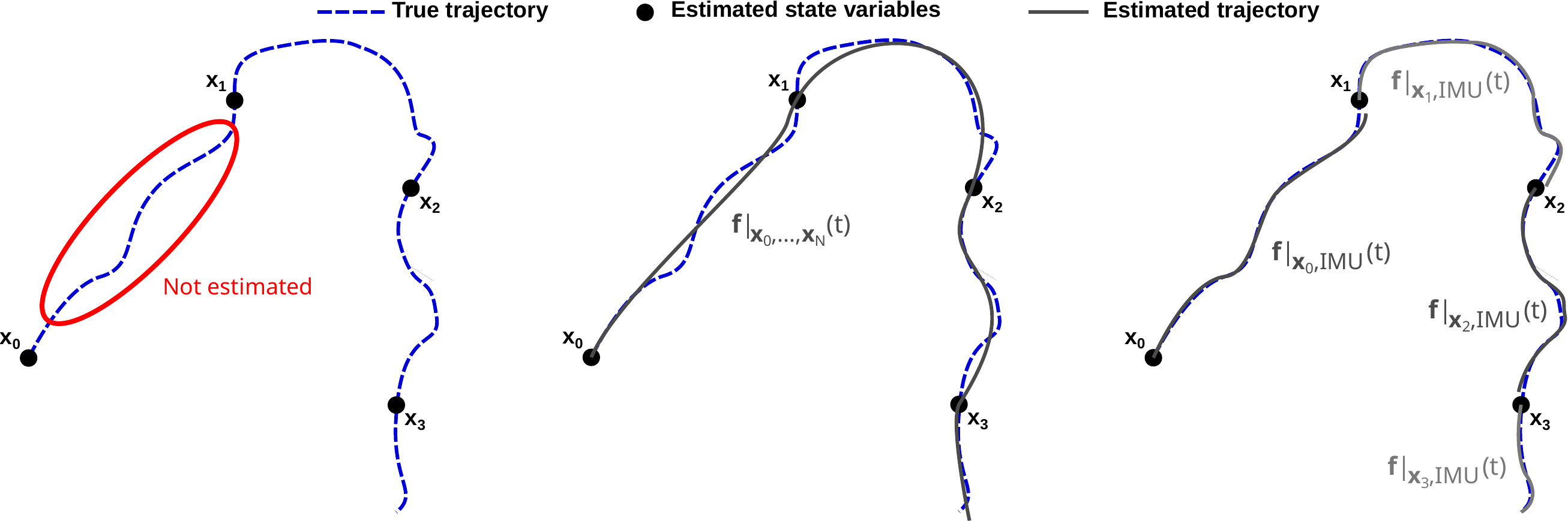}};
        \node[below = \legenddist of diagram, xshift = -\scale*\xoffset] {(a) Discrete-time estimation};
        \node[below = \legenddist of diagram] {(b) Continuous-time estimation};
        \node[below = \legenddist of diagram, xshift = \scale*\xoffset] {(c) Hybrid estimation};
    \end{tikzpicture}
    \caption{Most optimization-based lidar state estimation frameworks can be classified into three categories. (a) represents discrete-time estimation where the system's pose is estimated at a finite set of timestamps. The trajectory between timestamps is not estimated. (b) leverages a function over the whole duration of operations to represent the motion continuously. Such a method generally relies on some motion model and is parameterized by a set of supporting points. (c) shows a hybrid approach that locally characterizes the trajectory continuously based on inertial data, but uses discrete state variables at the global scale. This approach does not require an explicit motion model, but the global trajectory shows discontinuities when switching to a new timestamp. 2Fast-2Lamaa is built on the latter paradigm.}
    \label{fig:state_estimation}
\end{figure*}

There are many different approaches to state estimation in the robotics literature.
In this section, we mainly consider optimization-based lidar(-inertial) frameworks.
The majority of such methods can be classified into three distinct state estimation categories illustrated in Fig.~\ref{fig:state_estimation}: (a) discrete, (b) continuous, and (c) hybrid.
As mentioned in the introduction, motion distortion is addressed differently depending on the state representation.
The first one (a) dissociates the act of undistortion from state estimation.
Thus, motion correction is an open-loop process that preprocesses the scans based on previous estimates, and state estimation is discretely solved at the starting timestamp of each scan.
With (b) and (c), state estimation and motion distortion are coupled in a single problem that estimates the continuous trajectory of the sensor (at least locally).
Fully continuous approaches (b) generally assume a certain motion model to represent the sensor's movement with a continuous function.
\ac{imu} measurements can be used as residuals or control inputs of the system's trajectory.
Hybrid approaches (c) attempt to leverage the best of both (a) and (b) by locally parameterizing the trajectory based on inertial information, while keeping discrete variables at the global scale for simplicity and computational efficiency.
The rest of this subsection provides a brief literature review of the three types of methods

Considered as state-of-the-art odometry frameworks, Fast-LIO2 \citep{xu2022fastlio2} and DLIO \citep{chen2023dlio} are examples of one-off undistortion processes based on discrete-state estimation.
The former uses an iterated Kalman filter, with the core registration step consisting of optimizing point-to-plane residuals between the open-loop-undistorted scans and a global map built incrementally.
DLIO introduces a more precise way to undistort the point clouds using a constant-jerk motion model for continuous integration of the \ac{imu} measurements.
After motion-distortion correction, DLIO performs a variant of Generalized-\ac{icp} \citep{Segal2009} and uses a ``hierarchical geometric observer" \citep{lopez2023contracting} to estimate the system's pose efficiently.
Note that DLIO and Fast-LIO2 only optimize for the last sensor pose.
Other works leverage a similar undistortion process but optimize over a window of poses in a factor-graph formulation \citep{Ye2019,shan2020liosam}.
Note that the aforementioned frameworks require sufficient geometric features in the system's surroundings.
To address geometrically degenerated environments, COIN-LIO \citep{pfreundschuh2024coinlio} proposes to leverage intensity information from the lidar data to extract distinctive features even on flat surfaces.
More discrete-state lidar-inertial pipelines are discussed in a recent survey paper \citep{lee2024lidar}.

As shown in another recent survey paper \citep{talbot2025continuous}, there is a wide variety of continuous-time state formulations for robotics.
Some of the early lidar(-inertial) work built upon continuous-time representations that make strong assumptions about the motion.
For example, Zebedee \citep{Bosse2012} uses piece-wise linear functions to model the system's trajectory.
This corresponds to a strict assumption of constant velocity between control points.
Using inertial residuals, it demonstrated the ability to map a 3D environment using a randomly oscillating 2D lidar.
Later, LOAM \citep{Zhang2014b} used the same constant velocity assumption with both actuated 2D and 3D lidars.
This assumption is still used in recent frameworks like CT-LIO \citep{dellenbach2021cticp} or MOLA \citep{blanco2025mola}.
A key contribution of LOAM was the introduction of planar and edge features extracted from the raw lidar data for efficient registration.
This concept has been reused in many works, including LeGo-LOAM \citep{shan2018lego} and the present work (using a different extraction method).

Throughout the years, other continuous representations have been used to model more complex lidar trajectories.
A recurrent formulation is based on splines \citep{Droeschel2018,cao2025resple}.
The benefit of splines with respect to discrete state variables has been demonstrated in the context of vision-based \ac{slam} by \cite{cioffi2022continuous} at the cost of a higher computational cost.
Another branch of continuous-time estimation is based on a particular class of sparse \acp{gp} \citep{barfoot2024state}.
This approach elegantly models the trajectory based on a probabilistic motion model.
For both lidars and radars, continuous \ac{gp} states with a white-noise-on-acceleration motion prior have demonstrated high levels of accuracy and real-time computation \citep{burnett2024continuous}.
Note that different motion priors are possible, as demonstrated by \cite{Tang2019} with a white-noise-on-jerk prior.
And inertial data can also be used as control inputs \citep{burnett2025imu,lilge2025incorporating}.

The hybrid approach shown in Fig.~\ref{fig:state_estimation}(c) relies on \ac{imu} measurements to locally represent the system's trajectory.
An issue with inertial-based motion prediction/characterization is the strong dependence on initial conditions and biases when integrating and double-integrating the gyroscope and accelerometer measurements.
To prevent the reintegration of the \ac{imu} data every time the initial conditions are updated during the state optimization process, \cite{Lupton2012} introduced the concept of preintegration.
It corresponds to the creation of pseudo measurements that combine the information from multiple \ac{imu} measurements without the need to know the initial conditions.
Numerous works have improved on the original preintegration method \citep{Forster2017,Eckenhoff2019,Yang2020}.
The inertial chapter of the \ac{slam} Handbook provides an overview and comparison of several of these works \citep{slamhandbookch11}.
Note that, unlike the original preintegration use case, the hybrid state estimation approach does not attempt to combine \ac{imu} measurements into a small number of pseudo-measurements but rather requires an `upsampling' of the inertial information for it to be available at any timestamp.
This approach was originally introduced to address the issue of lidar-\ac{imu} extrinsic calibration by first upsampling the raw inertial readings before performing preintegration at a higher frequency \citep{LeGentil2018}.
Later approaches elegantly addressed preintegration under the scope of continuous state representation \citep{LeGentil2020a,legentil2023latent}, enabling full-batch lidar-inertial localization and mapping \citep{LeGentil2021}.
As this hybrid approach is well-suited to asynchronous inertial-aided estimation, other works with event cameras \citep{LeGentil2020idol,li2024asynchronous} and radars \citep{hatleskog2025imu} have adopted preintegration as their state representation.
2Fast-2Lamaa also uses continuous preintegration to characterize the system's motion during each lidar scan.
The corresponding discrete state is the gravity vector orientation, the linear velocity at the beginning of the scan, and the \ac{imu} biases, resulting solely in 11-\ac{dof} to optimize.

\subsection{Map representation}

Traditionally, robotic maps consist of geometric landmarks jointly estimated with the system’s pose \citep{dissanayake2001solution}.
Over the past decades, we have seen a constant increase in sensor bandwidth and computational power.
Naturally, the robotics community has harnessed these new hardware capabilities with algorithms that build and use denser and denser maps of the environment.
Nowadays, the most common representation for lidar state estimation consists of voxel maps where each cell stores the centroid and normal vector of all the points that occurred in the cell.
\edit{Fast-LIO2 \citep{xu2022fastlio2} is an example of such methods, leveraging point-to-plane registration constraints.
Other works, such as KISS-ICP \citep{vizzo2023kiss} and KISS-SLAM \citep{guadagnino2025kissslam}, chose to use point-to-point residuals, but suffer from significantly lesser performance on benchmarks such as the Newer College dataset \citep{ramezani2020newer}.}
Other frameworks keep more information in each cell, for example, the local distribution of the points with normal distributions \citep{biber2003ndt,magnusson2009three}.
Another way to model the environment is to use sets of surface primitives.
Surfels might be the simplest of these primitives and have been used in numerous lidar frameworks \citep{Bosse2012,Droeschel2018,Park2018}.
Some recent works deal with more complex primitives by directly building and using a mesh of the environment \citep{lin2023immesh,ruan2023slamesh}.

The aforementioned representations (point clouds, surfels, meshes, etc.) store information about the environment's surface.
While memory-efficient, they contain limited information about the rest of the space explored by the system.
Volumetric approaches, on the other hand, store information over the whole space and can track occupancy, \ac{tsdf}, etc.
These are less explored for lidar-based estimation, but are crucial for autonomous navigation (e.g., to plan safe trajectories through the environment).
Voxblox \citep{oleynikova2017voxblox} and FIESTA \citep{han2019fiesta} are prime examples of such mapping techniques.
Leveraging advances in computer graphics, many mapping frameworks have been built atop the OpenVDB structure \citep{museth2013openvdb,museth2013vdb}.
Some fuse information from multiple scans at the \ac{tsdf} level \citep{vizzo2022vdbfusion}, where others use the occupancy \citep{zhu2021vdbedt}, or the Euclidean distance field \citep{wu2024vdbgp}.
\edit{Focusing on state estimation, VoxGraph \citep{reijgwart2020voxgraph} extends VoxBlox \citep{oleynikova2017voxblox} and achieves real-time performance. 
It leverages SDF submaps to compute submap-to-submap pose constraints, which are then integrated into a pose-graph optimization framework.
More specific to lidar systems}, D-LIO \citep{cotoelena2025dlio} uses a fast truncated distance field stored in a multi-level hashmap to perform scan-to-map registration.
While the authors claim scalability to large-scale environments, their experiments on the VBR dataset \citep{brizi2024vbr} show a large RAM requirement that cannot be fulfilled without removing information from memory.
Another interesting work is shown by \cite{boche2025okvis2x} with OKVIS2-X, where vision-based estimation can be complemented with lidar-to-occupancy-map factors \citep{boche2024tightly} using the Supereight2 data structure \citep{funk2021multi}.
\edit{Focusing on exploration, \cite{schnid2021unified} also leverage volumetric mapping with a submap-based strategy to account for large state estimate drift.}

With the democratization of GPU computing, many robotics works have explored the use of neural representations and Gaussian splatting for map representation.
Point-SLAM \citep{sandstrom2023point} and Gs-icp SLAM \citep{ha2024rgbd} are examples of both approaches for \ac{slam} with RGB-D sensors.
DeepSDF \citep{park2019deepsdf} proposes an auto-decoder architecture to learn and infer the signed distance function at object-scale.
Later, iSDF \citep{ortiz2022isdf} demonstrates the ability to model and learn such a field online for room-scale environments.
While less popular, neural representations can be used for lidar-based state estimation.
\edit{LocNDF \citep{wiesmann2023locndf} addresses the localization problem with a neural field close to a \ac{tsdf}, but cannot be run at sensor framerate, even in small-scale environments, while} Nerf-LOAM \citep{deng2023nerfloam} performs odometry and mapping in large-scale environments.
More recently, PIN-SLAM \citep{pan2024pinslam} showcases top performance on the VBR dataset \citep{brizi2024vbr} via \ac{tsdf} modelling with real-time operations.
PINGS \citep{pan2025pings} builds upon PIN-SLAM by coupling Gaussian splats with neural distance fields for improved rendering capacities using both lidar and vision-based sensing.

2Fast-2Lamaa follows a line of work that mixes both surface-based and volumetric information.
Similar to the former, it only stores information on the surface of elements in the environment, but allows for the query of an Euclidean distance field approximation over the whole space without large memory requirements.
This trend originated from the goal of surface reconstruction with \ac{gp} implicit surfaces \citep{Williams2006}, which models a signed distance field close to the objects' surface.
Later, \cite{Wu2021} shows the ability to approximate the distance to the closest surface anywhere in space by applying a non-linear operation over a \ac{gp}-inferred field.
It is demonstrated with offline experiments that such \ac{gp}-based approaches enable lidar state estimation and planning \citep{wu2023log}.
\cite{legentil2024accurate} significantly improves the distance approximation while alleviating the original tradeoff between accuracy and surface interpolation.
It represents the foundation of the fast distance field derived in the present work, \edit{which does not consider a single and inefficient \ac{gp}, but breaks down the distance field modelling into many small local \acp{gp}.
Consequently, our new field formulation enables registration in kilometre-long maps in real-time.
In contrast, the original work by \cite{legentil2024accurate} can barely run object-level odometry at \SI{2}{\hertz}.}

\subsection{Dynamic object rejection}

A common assumption of many robotic state estimation algorithms is to consider the system’s environment to be static.
This assumption is very rarely verified in real-world applications.
While robust techniques like RANSAC or m-estimators enable a robot to deal with a certain level of dynamicity in the scene (considering dynamic object points as outliers), there has been a growing interest in performing dynamic object detection to create maps that only contain static elements \citep{yoon2019mapless,schmid2023dynablox,falque2023dynamic,jia2024beautymap,duberg2024dufomap,wu2024observation}.
However, most methods that focus on state estimation in dynamic environments first classify the points before performing standard scan registration \citep{pfreundschuh2021dynamic}, and optionally tracking objects \citep{jia2024trlo}.
Other methods simply detect dynamic objects after lidar data registration \citep{lichtenfeld2024efficient,legentil2024undistortion}, relying on robust estimators to `ignore' dynamic objects during the scan-alignment process.
An exception to these two-step approaches is BTSA  \citep{chen2025breaking}, which introduces a dynamic-aware \ac{icp} algorithm that couples the problems of dynamic point detection and state estimation in a single process.
\edit{Another method, HiMo \citep{zhang2025himo}, does not address the problem of ego-motion distortion, but tackles the moving object point cloud undistortion through scene flow estimation. 
We refer the reader to the relevant chapter of the \ac{slam} Handbook \citep{slamhandbookch15} for a deeper dive into state estimation in dynamic and deformable environments.}

As the problem of dynamic object detection is not the core focus of 2Fast-2Lamaa, the proposed optimization-based motion-distortion correction step relies on robust loss functions to consider dynamic objects as outlier information.
However, after scan-to-map registration, dynamic points can be removed from the map in an online or offline manner through some sort of ray tracing, similarly to the work from \cite{pomerleau2014long}.
In our ablation study, we evaluate the impact of the presence of dynamic points in the map on localization.

\section{Framework overview}

Let us consider a 6-DoF \ac{imu} (3-axis gyroscope and 3-axis accelerometer) and a 3D lidar rigidly mounted together.
The lidar collects 3D points denoted as $\lidarpoint{\lidar}{j}$. 
The gyroscope measures the \ac{imu} angular velocity $\gyr$, and the accelerometer the proper acceleration $\properacc{}$.
The homogeneous transformation $\Tc$ represents the extrinsic calibration between the two sensors.
We aim to estimate the pose of the \ac{imu} $\trans{\world}{\imu_\time{}}$  at time $\time{}$ with respect to a world frame $\refframe{\world}{}$, with or without a prior map of the environment.
The timestamp of the start of a lidar scan is denoted $\frametime{i}$, and $\refframe{\imu_\time{}}{}$ refers to the \ac{imu} frame at time $\time{}$.
As shown in Fig.~\ref{figure:overview}, the proposed framework consists of two main components: one module for tightly-coupled lidar-inertial motion-distortion correction and a second one for scan-to-map localization.

To undistort lidar scans, 2Fast-2Lamaa first extracts lidar features from the raw scans and performs continuous preintegration with the \ac{imu} data.
Using continuous \ac{imu} preintegration, the system's trajectory is parameterized by the gravity vector $\gravity_\frametime{i}$ in $\refframe{\imu_\frametime{i}}{}$, the velocity $\vel{\imu_\frametime{i}}{\frametime{i}}$ at time $\frametime{i}$ in $\refframe{\imu_\frametime{i}}{}$, and both the accelerometer and gyroscope biases $\biasacc{i}$ and $\biasgyr{i}$ (considered constant for the duration of the scan).
After associating features together, the motion during the lidar scans is estimated by minimizing lidar feature distances (point-to-plane and point-to-line, depending on the feature's nature) in a non-linear least-squares formulation.
The output of this module consists of motion-corrected lidar point clouds $\pointcloud{i}{}$ and associated scan-to-scan incremental motion $\transtilde{\imu_\frametime{i}}{\imu_\frametime{i+1}}$.

For trajectory estimation, the localization module integrates the aforementioned incremental motion estimates and refines the global pose $\transtilde{\world}{\imu_\frametime{i}}$ of $\pointcloud{i}{}$ with scan-to-map registration.
The proposed framework can build the map $\map$ incrementally (mapping mode) or directly leverage an existing map of the environment (localization mode).
Using \ac{gp} regression, the map provides an approximation of the scene's Euclidean distance field.
With the ability to efficiently query the distance $\dist(\lidarpoint{}{})$ to the closest surface for any $\lidarpoint{}{}\in\reals^3$, the registration is the direct minimization of $\dist$ for a subset of points from $\pointcloud{i}{}$.
Once an undistorted scan is registered to the map, it can be used to update the underlying data structure: extending the area covered by $\map$ or removing elements that are no longer present in the environment (such as dynamic objects or structural changes).
Note that for repeating trajectories, both localization and mapping can be performed in a topometric manner by considering a chain of connected submaps instead of a unique global map.
When mapping the environment with submaps, an optional \edit{online} loop-closure detection and correction mechanism, presented in Section~\ref{sec:loop}, can be used for globally consistent trajectory estimation.

\begin{figure*}
    \centering
    \def\hdist{4em}
\def\vdistlong{2.5em}
\def\vdist{1.5em}
\def\blockheight{3.0em}
\def\blockwidth{5.0em}
\def\innerpad{1.0em}
\def\textsize{\scriptsize}
\begin{tikzpicture}[auto]
    \tikzstyle{input} = [draw, fill=white, rectangle, minimum height = 2.5em, text width = 3.5em,  minimum width = 3.8em, inner sep=0, outer sep=0, align = center, node distance = 5em, draw=red, execute at begin node=\setlength{\baselineskip}{8pt}]
    \tikzstyle{block} = [draw, fill=white, rectangle, minimum height = \blockheight, text width = \blockwidth,  minimum width = \blockwidth, align = center, inner sep=0, outer sep=0, node distance = 11em, execute at begin node=\setlength{\baselineskip}{8pt}] 
    \tikzstyle{wideblock} = [block, text width=((2*\blockwidth)+\hdist), minimum width=((2*\blockwidth)+\hdist)]
    \tikzstyle{midblock} = [block, text width=(\blockwidth+(0.25*\hdist)), minimum width=(\blockwidth+(0.25*\hdist))]
    \tikzstyle{output} = [draw=none, fill=white, text=NavyBlue, rectangle, minimum height = 3em, text width = \blockwidth,  minimum width = \blockwidth, align = center, node distance = 11em, execute at begin node=\setlength{\baselineskip}{8pt}] 
    \tikzstyle{bigbox} = [draw, dashed, ultra thick, fill=white, rectangle, text width = 0.3\columnwidth,  minimum width = 0.3\columnwidth, minimum height = ((2*\blockheight)+(2*\innerpad)+\vdist), align = center, node distance = 11em, inner sep=0, outer sep=0, execute at begin node=\setlength{\baselineskip}{8pt}]

    \tikzstyle{varrowleft} = [text width = 5em, align=right, left, text width=((0.5*\blockwidth)+\innerpad), execute at begin node=\setlength{\baselineskip}{7pt}] 
    \tikzstyle{varrowright} = [text width = 5em, align=left, right, text width=((0.5*\blockwidth)+\innerpad), execute at begin node=\setlength{\baselineskip}{7pt}] 
    \tikzstyle{harrowabove} = [text width = \hdist, align=center, above, execute at begin node=\setlength{\baselineskip}{7pt}] 
    \tikzstyle{harrowbelow} = [text width = \hdist, align=center, below, execute at begin node=\setlength{\baselineskip}{7pt}]

    \node [bigbox, text width = ((2*\blockwidth)+\hdist+(2*\innerpad)),  minimum width = ((2*\blockwidth)+\hdist+(2*\innerpad))] (odometry) {};
    \node [block, below right= 1.41*\innerpad of odometry.north west] (feature) {\textsize \textbf{Feature point extraction}};
    \node [block, above right=1.41*\innerpad of odometry.south west] (preint) {\textsize \textbf{Continuous preintegration}};
    \node [block, right= \hdist of feature] (asso) {\textsize \textbf{Data association}};
    \node [block, right=\hdist of preint] (opti) {\textsize \textbf{11-DoF optimization}};

    \node [input, left=\hdist of feature] (lidar) {\textsize \textbf{3D lidar}};
    \node [input, left=\hdist of preint] (imu) {\textsize \textbf{6-DoF IMU}};

    \node [bigbox, right=\hdist of odometry, text width = ((2*\blockwidth)+\hdist+(2*\innerpad)),  minimum width = ((2*\blockwidth)+\hdist+(2*\innerpad))] (mapping) {};
    \node[wideblock, above right=1.41*\innerpad of mapping.south west] (register) {\textsize \textbf{Localization}\\Scan-to-map registration};
    \node[block, below right=1.41*\innerpad of mapping.north west] (carving) {\textsize \textbf{Map free space carving}};
    \node[block, right=(\hdist) of carving] (map) {\textsize \textbf{Continuous map}\\(or submaps)};

    \draw[->] (imu) -- node[harrowabove, xshift=-0.5*\innerpad]{\textsize Acc./gyr. data} (preint);
    \draw[->] (lidar) -- node[harrowabove, xshift=-0.5*\innerpad]{\textsize Scans} (feature);
    \draw[->] (feature) -- node[varrowleft]{\textsize Feature times} (preint);
    \draw[->] (feature) -- node[harrowabove]{\textsize Features $\planar_i, \edge_i$} (asso);
    \draw[->] ([xshift=0.5*\hdist]preint.east) |- ([yshift=-1.0em]asso.west);
    \draw[->] (preint) -- node[harrowbelow]{\textsize Preint. meas. \\$\drot{}{},\dpos{}{}$} (opti);
    \draw[->] (asso) -- node[varrowright]{\textsize Feature asso.} (opti);

    \draw[->] ([xshift=1.0em]carving.south |- register.north) -- node[varrowleft]{\textsize Registered point cloud} ([xshift=1.0em]carving.south);
    \draw[->] ([xshift=1.0em,yshift=-(0.5*\vdist)]carving.south)-| ([xshift=-1.5em]map.south);
    \draw[->] ([xshift=-1.1em]map.south) -- node[varrowright, text width = ((0.5*\blockwidth)+\innerpad+1.0em)]{\textsize Distance function $\dist(\lidarpoint{}{})$} ([xshift=-1.1em]map.south |- register.north);
    \draw[->] ([yshift=-0.2em]map.west) -- node[harrowbelow]{\textsize Voxel map} ([yshift=-0.2em]carving.east);
    \draw[->] ([yshift=0.2em]carving.east) -- node[harrowabove]{\textsize Map update} ([yshift=0.2em]map.west);
    \draw[->] (opti) -- node[harrowabove]{\textsize Pose increment $\transtilde{\imu_\frametime{i}}{\imu_\frametime{i\plus1}}$} node[harrowbelow, swap]{\textsize Undistorted scan $\pointcloud{i}{}$} (register);

    \node[output, right=(\innerpad+1.0em) of map] (outmap) {\textsize \textbf{Map $\map$ for reuse}};
    \node[output, right=(\innerpad+1.0em) of register] (outpose) {\textsize \textbf{Pose estimate $\transtilde{\world}{\imu_\frametime{i}}$}};
    \draw[->] (map) -- (outmap);
    \draw[->] (register) -- (outpose);
    \node[below=0.3em of mapping.south, inner sep=0em] {\small \textbf{Localization and mapping}};
    \node[below=0.3em of odometry.south, inner sep=0em] {\small \textbf{Undistortion/scan-to-scan odometry}};
    
\end{tikzpicture}
    \caption{2Fast-2Lamaa consists of two functional blocks. The first one is an optimization-based motion correction module that undistorts lidar scans using continuous IMU preintegration to characterize the system motion with only 11-DoFs. Once corrected, the scans are used for scan-to-map registration to estimate the global pose of the system. }
    \label{figure:overview}
\end{figure*}

\section{Undistortion-based odometry}

The undistortion module in 2Fast-2Lamaa is based on previous work for map-less and initialization-free dynamic object detection \citep{legentil2024undistortion}.
The main differences are a novel approach for feature extraction and a change in the definition of the temporal window used for motion estimation.
2Fast-2Lamaa offers a faster front-end and reduces the time delay in the state computation by estimating the motion for the last incoming lidar scan, not waiting for subsequent scans as originally done.

\subsection{IMU preintegrated continuous state}

To undistort the incoming lidar data, let us consider the lidar points and \ac{imu} readings that have been collected over a short period of time.
For convenience and implementation efficiency, this temporal window is chosen to span over two lidar scans (each defined as a full revolution for spinning sensors), with the previous scan from $\frametime{i\minus1}$ to $\frametime{i}$ and the current scan from $\frametime{i}$ to $\frametime{i\plus1}$.
As formulated for SE(3) by \cite{Forster2017}, the rotation, velocity and translation preintegrated measurements from time $\frametime{i\minus1}$, $\drot{\frametime{i}}{\time{}}$ and $\dpos{\frametime{i}}{\time{}}$, correspond to
\begin{align}
    \begin{aligned}
    \drot{\frametime{i\minus1}}{\time{}} & = \prod_{\frametime{i\minus1}}^{\time{}} \left(\Exp(\gyr(\mathfrak{\time{}}) - \biasgyr{}(\mathfrak{\time{})})\right)^{d\mathfrak{\time{}}},
    \\
    \dvel{\frametime{i\minus1}}{\time{}} & = \int_{\frametime{i\minus1}}^{\time{}} \drot{\frametime{i\minus1}}{s}(\properacc{}(\mathfrak{\time{}}) - \biasacc{}(\mathfrak{\time{}}))  ds
    \\
    \dpos{\frametime{i\minus1}}{\time{}} & = \int_{\frametime{i\minus1}}^{\time{}}\int_{\frametime{i\minus1}}^{\mathfrak{\time{}}} \drot{\frametime{i\minus1}}{s}(\properacc{}(\mathfrak{\time{}}) - \biasacc{}(\mathfrak{\time{}}))  dsd\mathfrak{\time{}}
    \label{eq:preintegration},
    \end{aligned}
\end{align}
with $\biasgyr{}$ and $\biasacc{}$ slow-varying additive biases.
Using the \acp{lpm} \citep{legentil2023latent} to compute~\eqref{eq:preintegration} based on piece-wise linear continuous representation of the inertial data, the \ac{imu} pose and velocity through time are defined as
\begin{align}
    \begin{aligned}
    \trans{\imu_\frametime{i\minus1}}{\imu_\time{}} &= \begin{bmatrix}
        \rot{\imu_\frametime{i\minus1}}{\time{}}&\pos{\imu_\frametime{i\minus1}}{\time{}} \\ \mathbf{0}&1
    \end{bmatrix}\quad \text{with} \quad \rot{\imu_\frametime{i\minus1}}{\time{}} = \drot{\frametime{i\minus1}}{\time{}},
    \\
    \vel{\imu_\frametime{i\minus1}}{\imu_\time{}} &= \vel{\imu_\frametime{i\minus1}}{\imu_\frametime{i\minus1}} + (\time{} - \frametime{i\minus1})\gravity_{\frametime{i\minus1}} + \dvel{\frametime{i\minus1}}{\time{}}.
    \\
    \pos{\imu_\frametime{i\minus1}}{\imu_\time{}} &= (\time{} - \frametime{i\minus1})\vel{\imu_\frametime{i\minus1}}{\imu_\frametime{i\minus1}} + \frac{(\time{} - \frametime{i\minus1})^2}{2}\gravity_{\frametime{i\minus1}} + \dpos{\frametime{i\minus1}}{\time{}}.
    \label{eq:preint_trans}
    \end{aligned}
\end{align}
Note that the preintegrated measurements $\drot{\frametime{i\minus1}}{\time{}}$ and $\dpos{\frametime{i\minus1}}{\time{}}$ are computed using a prior knowledge of the biases $\biasaccbar{i\minus1}$ and $\biasgyrbar{i\minus1}$.
Similar to the original on-manifold preintegration work \citep{Forster2017}, the preintegrated measurements are corrected through a first-order Taylor expansion to account for the unknown nature of the biases:
\begin{align}
    \drot{\frametime{i\minus1}}{\time{}} &\approx \drotbar{\frametime{i\minus1}}{\time{}}\Exp\left(\frac{\partial\drot{\frametime{i\minus1}}{\time{}}}{\partial\biasgyr{i\minus1}}\Delta\biasgyr{i\minus1}\right),
    \nonumber
    \\
    \dvel{\frametime{i\minus1}}{\time{}} &\approx \dvelbar{\frametime{i\minus1}}{\time{}} + \frac{\partial\dvel{\frametime{i\minus1}}{\time{}}}{\partial\biasgyr{i\minus1}}\Delta\biasgyr{i\minus1} + \frac{\partial\dvel{\frametime{i\minus1}}{\time{}}}{\partial\biasacc{i\minus1}}\Delta\biasacc{i\minus1},
    \\
    \dpos{\frametime{i\minus1}}{\time{}} &\approx \dposbar{\frametime{i\minus1}}{\time{}} + \frac{\partial\dpos{\frametime{i\minus1}}{\time{}}}{\partial\biasgyr{i\minus1}}\Delta\biasgyr{i\minus1} + \frac{\partial\dpos{\frametime{i\minus1}}{\time{}}}{\partial\biasacc{i\minus1}}\Delta\biasacc{i\minus1},
    \nonumber
\end{align}
with $\Delta\biasacc{i\minus1} = \biasacc{i\minus1} - \biasaccbar{i\minus1}$, and $\Delta\biasgyr{i\minus1} = \biasgyr{i\minus1} - \biasgyrbar{i\minus1}$.
Thus, the system's trajectory during the temporal window is fully characterized by $\state = \{\gravity_{\frametime{i\minus1}}$, $\vel{\imu_\frametime{i\minus1}}{\imu_\frametime{i\minus1}}$, $\biasacc{i\minus1}$, $\biasgyr{i\minus1} \}$.
Accordingly, a lidar point $\lidarpoint{\lidar}{}$ collected at time $\pointtime{}$ can be projected to the $\refframe{\imu_\frametime{i\minus1}}{}$ frame as 
\begin{equation}
\begin{bmatrix}
    \lidarpoint{\imu_\frametime{i\minus1}}{} \\ 1
\end{bmatrix} = \trans{\imu_\frametime{i\minus1}}{\imu_\pointtime{}} \Tc \begin{bmatrix}
    \lidarpoint{\lidar}{} \\ 1
\end{bmatrix}.
\label{eq:projection}
\end{equation}

\subsection{Lidar features and data association}

\begin{figure}
    \centering
    \includegraphics[width=0.99\columnwidth]{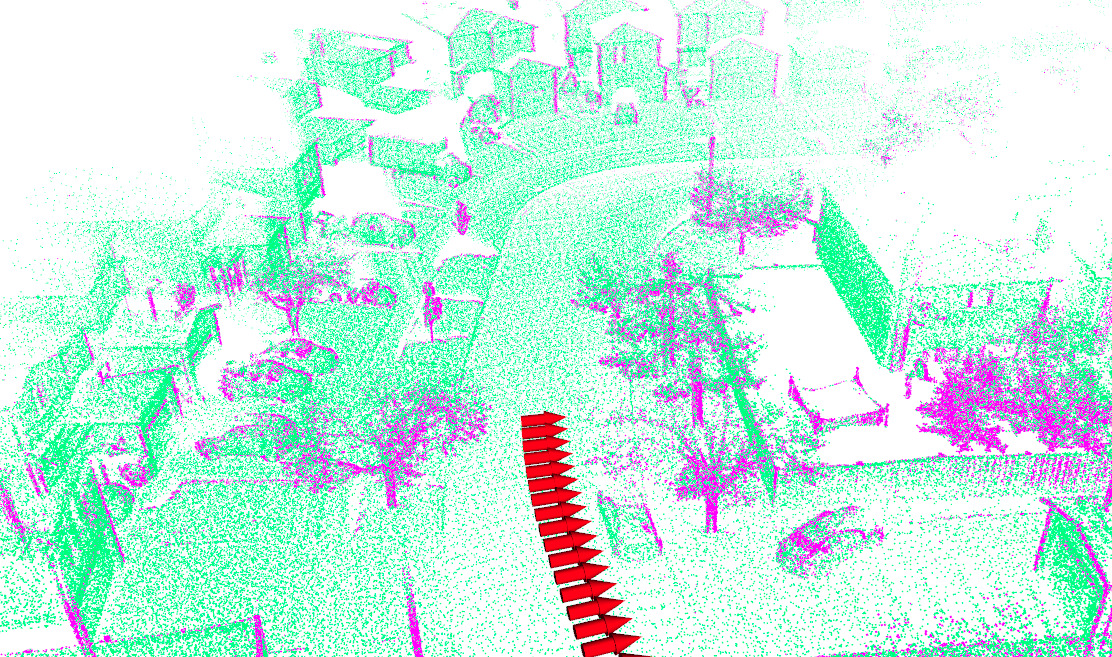}
    \caption{Accumulated lidar features (planar in green, edge in magenta) over \SI{3}{\s} of data in a \emph{Suburbs} sequence from the Boreas-RT dataset.}
    \label{fig:features}
\end{figure}

Using the raw lidar point clouds, features are extracted independently in the two scans that constitute the temporal window $[\frametime{i\minus1},\frametime{i\plus1}]$.
\edit{Similarly to the work by~\cite{Zhang2014b}}, features can be \emph{planar} or \emph{edge} points.
\edit{However, the feature point selection does not follow any type of curvature/roughness score.}
The planar points $\planar_{i\minus1}$ and $\planar_{i}$ are obtained by voxel-subsampling the corresponding scans and retaining only one point per voxel (\edit{random selection among the points in each voxel}). 
For the edge points, the proposed method considers each lidar channel/ring independently and detects jumps in range values between consecutively collected points.
The latter form the sets of edges $\edge_{i\minus1}$ and $\edge_{i}$.
\edit{Note that the sets $\planar_{i}$ and $\edge_{i}$ are mutually exclusive.}
Fig.~\ref{fig:features} shows an example of features obtained over \SI{3}{\s} of data in a suburban environment.

Then, the features from the two scans are associated together, between $\planar_{i\minus1}$ and $\planar_{i}$ or between $\edge_{i\minus1}$ and $\edge_{i}$, with a k-closest-neighbour search that enforces some conditions that depend on the feature type.
For planar points, each point in $\planar_{i}$ is associated with three features from $\planar_{i\minus1}$, enforcing a minimum spread in terms of distance and angle, to avoid colinearity \edit{(a minimum of \SI{0.05}{\m} and \SI{75}{\degree} in our implementation)}. 
Similarly, each edge feature in $\edge_{i}$ is matched with two sufficiently spaced points from $\edge_{i\minus1}$ \edit{(typically \SI{0.05}{\m})}.
For each point, the association begins with a radius search that provides the set of points $\neighbours$.
Starting with the closest point in $\neighbours$, the process iteratively tests the type-specific conditions for each point in $\neighbours$ by order of distance, until the required number of valid points is reached (successful association) or when there are no more points to test in $\neighbours$ (failed association). 

\subsection{Motion correction}

The motion is estimated via a non-linear least-squares optimization of point-to-plane and point-to-line distances
\begin{equation}
    \tilde{\state} = \underset{\state}{\operatorname{argmin}} \sum_{\lidarpoint{\lidar}{j}\in\planar_{i}} \residualplanar{j}^2 + \sum_{\lidarpoint{\lidar}{j}\in\edge_{i}} \residualedge{j}^2,
    \label{eq:odom_optimization}
\end{equation}
where $\residualplanar{j}$ is the point-to-plane distance between a planar feature $\lidarpoint{\lidar}{j}$ and the associated trio $\lidarpoint{\lidar}{k}$, $\lidarpoint{\lidar}{l}$, and $\lidarpoint{\lidar}{m}$, and $\residualedge{j}$ is the point-to-line distance between an edge feature $\lidarpoint{\lidar}{j}$ and the associated pair $\lidarpoint{\lidar}{k}$ and $\lidarpoint{\lidar}{l}$.
Using \eqref{eq:preint_trans} and \eqref{eq:projection} for point projection, the residuals are computed as 
\begin{equation}
    \residualplanar{j} = \textstyle \frac{(\lidarpoint{\imu_\frametime{i\minus1}}{j} - \lidarpoint{\imu_\frametime{i\minus1}}{k})^\top \left((\lidarpoint{\imu_\frametime{i\minus1}}{k} - \lidarpoint{\imu_\frametime{i\minus1}}{l}) \times (\lidarpoint{\imu_\frametime{i\minus1}}{k} - \lidarpoint{\imu_\frametime{i\minus1}}{m}) \right)}{\Vert(\lidarpoint{\imu_\frametime{i\minus1}}{k} - \lidarpoint{\imu_\frametime{i\minus1}}{l}) \times (\lidarpoint{\imu_\frametime{i\minus1}}{k} - \lidarpoint{\imu_\frametime{i\minus1}}{m})\Vert},
\end{equation}
and
\begin{equation}
    \residualedge{j} = \textstyle \frac{\Vert (\lidarpoint{\imu_\frametime{i\minus1}}{j} - \lidarpoint{\imu_\frametime{i\minus1}}{k}) \times (\lidarpoint{\imu_\frametime{i\minus1}}{j} - \lidarpoint{\imu_\frametime{i\minus1}}{l}) \Vert}{\Vert\lidarpoint{\imu_\frametime{i\minus1}}{k} - \lidarpoint{\imu_\frametime{i\minus1}}{l}\Vert}.
\end{equation}
As the magnitude of the gravity vector $\gravity_\frametime{i\minus1}$ is known, the optimization \eqref{eq:odom_optimization} leverages the 2-sphere manifold, resulting in a total of eleven \acp{dof}.
It is solved using the Levenberg-Marquardt algorithm.
After convergence, the estimated state $\tilde{\state}$ and \eqref{eq:preint_trans} allow the computation of $\transtilde{\imu_\frametime{i}}{\pointtime{}}$ to correct the current scan's motion distortion and provide the scan-to-scan pose increment $\transtilde{\imu_\frametime{i}}{\imu_\frametime{i\plus1}}$.
The undistorted point cloud $\pointcloud{i}{}$ and $\transtilde{\imu_\frametime{i}}{\imu_\frametime{i\plus1}}$ are passed to the localization module (next section) for global pose estimation.
The velocity and gravity estimates projected to $\frametime{i}$, as well as an average of the past bias estimates, are used as the state's initial guess for the next temporal window.

The undistortion process presented in this section is performed for every incoming scan and estimates the continuous trajectory between $\frametime{i\minus1}$ and $\frametime{i\plus1}$.
It is done without fixing the motion between $\frametime{i\minus1}$ and $\frametime{i}$ in any way.
Computing the motion for the duration of two scans at every scan creates an overlap in the consecutive temporal windows.
That is why, at each step, the module solely outputs the transformation between $\frametime{i}$ and $\frametime{i\plus1}$, ignoring the estimate between $\frametime{i\minus1}$ and $\frametime{i}$.
It is important to note that the trajectory over a small temporal window (typically \SI{200}{\ms}) is often close to translation-only movement\edit{, or can be close to rotation-only in some scenarios}.
These motions are not informative enough to make the accelerometer observable, as there is an ambiguity with gravity, as detailed by \cite{Tereshkov2015}.
To address this issue, a weak zero-mean prior residual \edit{$\residualbias{} = c\biasacc{i\minus1}$, with $c$ a constant,} is added to prevent unrealistically high accelerometer bias estimates.
\edit{Thus, \eqref{eq:odom_optimization} becomes
\begin{equation}
    \tilde{\state} = \underset{\state}{\operatorname{argmin}} \sum_{\lidarpoint{\lidar}{j}\in\planar_{i}} \residualplanar{j}^2 + \sum_{\lidarpoint{\lidar}{j}\in\edge_{i}} \residualedge{j}^2 + \Vert\residualbias{}\Vert^2.
\end{equation}
}

\section{Localization}
\label{sec:localization}

\subsection{Scan-to-map registration}
Given a motion-corrected point cloud $\pointcloud{i}{}$, 2Fast-2Lamaa estimates the scan-to-map transformation $\trans{\world}{\imu_\frametime{i}}$ using a map $\map$ (given or incrementally built as detailed in Section~\ref{sec:map}), which provides a distance field $\dist(\lidarpoint{}{})$ that can be queried for any point $\lidarpoint{}{}$.
As the undistortion step provides decent pose increment estimates, we do not estimate $\trans{\world}{\imu_\frametime{i}}$ directly, but perform a right-hand perturbation 
\begin{equation}
    \trans{\world}{\imu_\frametime{i}} = \transtilde{\world}{\imu_\frametime{i-1}}\transtilde{\imu_\frametime{i-1}}{\imu_\frametime{i}}\begin{bmatrix}
        \Exp(\Delta\mathbf{r}) & \Delta\mathbf{p} \\ \mathbf{0} & 1
    \end{bmatrix},
    \label{eq:perturbation}
\end{equation}
with $\transtilde{\world}{\imu_\frametime{i-1}}$ the global pose estimate of the previous scan, $\transtilde{\imu_\frametime{i-1}}{\imu_\frametime{i}}$ the pose increment from the undistortion step, and $\Delta \mathbf{r}$ and $\Delta\mathbf{p}$ the rotational and translational perturbation, respectively.
Thus, using~\eqref{eq:perturbation}, the global registration estimates
\begin{equation}
    \tilde{\Delta \mathbf{r}}, \tilde{\Delta \mathbf{p}} = \underset{\Delta \mathbf{r}, \Delta \mathbf{p}}{\operatorname{argmin}} \  \sum_{\lidarpoint{j}{} \in \pointcloud{i}{}} \left(\dist\left(\trans{\world}{\imu_\frametime{i}}\begin{bmatrix}
        \lidarpoint{j}{}\\1
    \end{bmatrix}\right)\right)^2
    \label{eq:registration}
\end{equation}
using the Levenberg-Marquardt algorithm.
The key of 2Fast-2Lamaa is the computation of the distance field $\dist$.
It is detailed in Section~\ref{sec:map}.

\edit{
\subsection{Gravity-alignment residual}
Inspired by~\cite{noh2025garlio}, the scan-to-map registration~\eqref{eq:registration} can be complemented with a gravity-alignment residual $\mathbf{r}_{g_i}$, when used for odometry or mapping, as opposed to pure localization.
First the gravity vector $\gravity_{\world}$ in the original sensor frame at $\frametime{0}$ is estimated alongside the system's linear velocities $\vel{\world}{\imu_\frametime{i}}$ and \ac{imu} biases $\biasacc{}$ and $\biasgyr{}$, using the first $M$ estimated poses $\trans{\world}{\imu_{\frametime{i}}}$, obtained through scan-to-map registration.
$M$ is programmatically chosen to guarantee that enough translation and rotation occurred in the trajectory to make the accelerometer biases observable.
Defining the optimized state $\mathcal{S}_g = \{\gravity_{\world}, \vel{\world}{\imu_\frametime{0}}, \cdots, \vel{\world}{\imu_\frametime{M\minus1}}, \biasacc{}, \biasgyr{}\}$, the corresponding optimization is
\begin{equation}
    \tilde{\mathcal{S}}_g = \underset{\mathcal{S}_g}{\text{argmin}} \sum_{i=0}^{M\minus2} \Vert\residualimu{i}\Vert^2,
\end{equation}
with the residual $\residualimu{i}$ as
\begin{equation}
    \residualimu{i} = \begin{bmatrix}
        \Log\left(\rot{\world}{\imu_\frametime{i}}^\top\rot{\world}{\imu_\frametime{i\plus1}} \drot{\frametime{i}}{\frametime{i\plus1}}^\top\right)
        \\
        \vel{\world}{\imu_\frametime{i\plus1}} - \vel{\world}{\imu_\frametime{i}} - \Delta\frametime{i}\gravity_{\world} - \rot{\world}{\imu_\frametime{i}}\dvel{\frametime{i}}{\frametime{i\plus1}}
        \\
        \pos{\world}{\imu_\frametime{i\plus1}} - \pos{\world}{\imu_\frametime{i}} - \Delta\frametime{i}\vel{\world}{\imu_\frametime{i}} - \frac{\Delta\frametime{i}}{2}\gravity_{\world} - \rot{\world}{\imu_\frametime{i}}\dpos{\frametime{i}}{\frametime{i\plus1}}.
    \end{bmatrix}
\end{equation}

Then, for any incoming undistorted scan and the associated linear body velocity $\vel{\imu_\frametime{i}}{\imu_\frametime{i}}$, a measurement of the gravity $\gravity_{\frametime{i}}$ in the current frame is obtained as
\begin{equation}
    \gravity{\frametime{i}} = \frac{\vel{\imu_\frametime{i}}{\imu_\frametime{i}} - \drot{\frametime{i\minus1}}{\frametime{i}}^\top(\vel{\imu_\frametime{i\minus1}}{\imu_\frametime{i\minus1}}-\dvel{\frametime{i\minus1}}{\frametime{i}})}{\Delta\frametime{i\minus1}}.
\end{equation}
Finally, the gravity residual $\mathbf{r}_{g_i}$ is defined as
\begin{equation}
    \mathbf{r}_{g_i} = \arccos\left(\frac{\gravity_{\world}^\top \rot{\world}{\imu_\frametime{i}} \gravity_{\frametime{i}}}{\Vert\gravity_{\world}\Vert \Vert\gravity_{\frametime{i}}\Vert}\right),
\end{equation}
and its squared norm is added to the right-hand side of~\eqref{eq:registration}.

}

\subsection{Topometric localization}

\begin{figure}
    \centering
    \includegraphics[width=0.99\columnwidth]{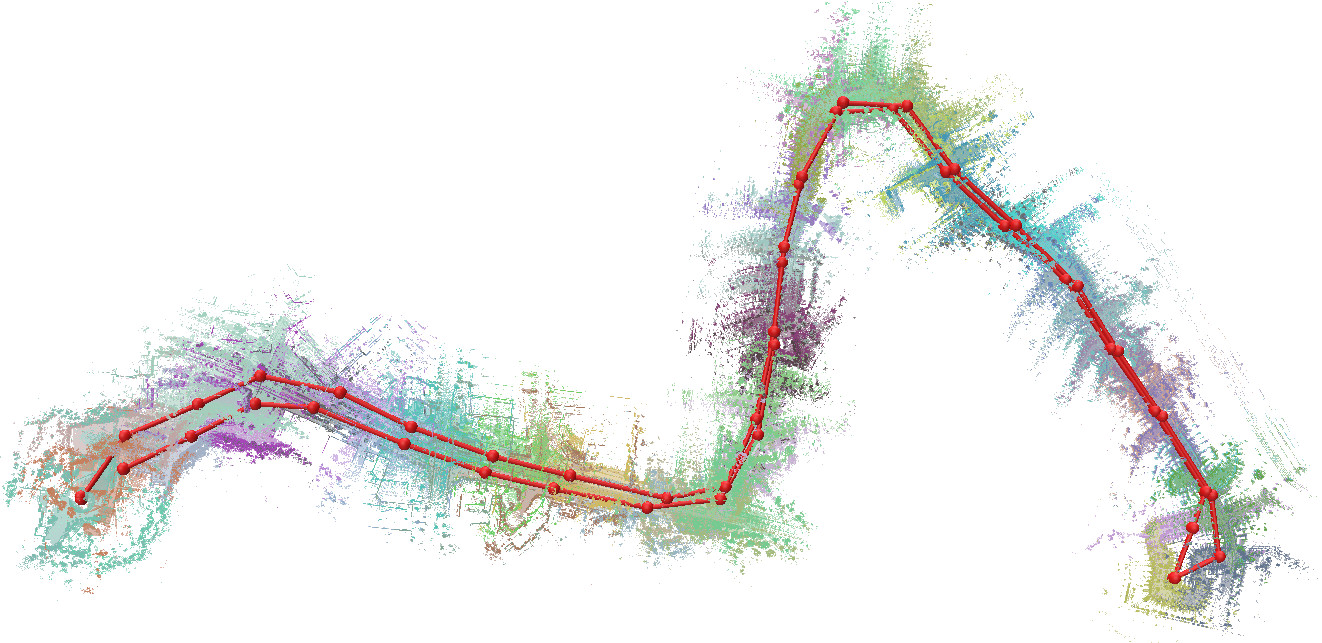}
    \caption{Example of topometric map obtained on a \emph{Suburbs} sequence of the Boreas-RT dataset. It consists of a succession of submaps (shown as point clouds of different colours) and a topological graph (shown in red, with the sphere being the submaps' centroids) that connects the submaps. This topometric map does not require global consistency to enable state-of-the-art localization in repeating trajectories.}
    \label{fig:topometric_map}
\end{figure}

Atop localization in a globally consistent map, 2Fast-2Lamaa enables topometric localization (and mapping) for repeated robot operations along the same path.
This is inspired by the \emph{Teach and Repeat} framework \citep{furgale2010vtr}, which represents the environment with a graph of overlapping geometric submaps topologically linked based on a demonstrated robot trajectory.
The localization process has two `layers'.
At the high level, it consists of moving from submap to submap (node to node) in the topological graph.
At the low level, it performs scan-to-submap geometric registration.
This approach alleviates the need for a globally consistent map while allowing high-precision localization for autonomous navigation.
We have integrated this mapping and localization strategy into the proposed framework.
In its current version, 2Fast-2Lamaa only considers the simplest type of graph: an \emph{undirected chain}.
Concretely, the localization in that chain consists of performing scan-to-submap registration using~\eqref{eq:registration} and the submap associated with the current node in the chain.
Then, if the state estimate reaches the edge of the submap, the next node (or previous one, depending on the direction of motion) becomes the current node.
Fig.~\ref{fig:topometric_map} illustrates a topometric map in the \emph{Suburbs} environment of our self-collected dataset.
While the trajectory used to create the map drifted from the ground-truth, the topometric nature of the localization enables accurate localization for repeating trajectories ($<~\SI{3}{\cm}$ lateral position \ac{rmse}, see Section~\ref{sec:exp_self}).

\section{Mapping}
\label{sec:map}

This section presents the proposed \ac{gp}-based distance field for efficient scan-to-map registration in Section~\ref{sec:localization}.
The map can be provided with a single point cloud of the robot's environment or incrementally built according to the scan-to-map odometry estimates.
The incremental map creation can be done using a single map or a collection of submaps for the topometric localization mode.
The process presented in this section applies to all of these mapping strategies.
The only difference for the topometric approach is that a new submap is created after a user-defined distance travelled since the last submap creation.

\subsection{GP-based distance field}

The proposed mapping process is based on \ac{gp}-based distance fields \citep{legentil2024accurate}.
Using a point cloud as input, it performs standard \ac{gp} regression to obtain a latent field $\occ(\lidarpoint{}{})$ that can be seen as an occupancy field equal to 1 on object surfaces (where the lidar points lie), and that decreases to zero further away from it.
Applying a specific non-linear function over $\occ(\lidarpoint{}{})$ provides an approximation of the Euclidean distance between the query location $\lidarpoint{}{}$ and the closest surface.
Formally, $\occ$ is modelled with a zero-mean \ac{gp}
\begin{equation}
    \occ(\lidarpoint{}{}) \sim \gp\left(0,\kernel{}{\lidarpoint{}{}}{\lidarpoint{}{}'}\right),
\end{equation}
with $\kernel{}{\lidarpoint{}{}}{\lidarpoint{}{}'}$ the covariance kernel function that specifies the covariance between two instances of the field $\occ(\lidarpoint{}{})$ and $\occ(\lidarpoint{}{}')$.
Given a set of $N$ observations/points (equal to one) at locations $\lidarpoints{}{}$, and following standard \ac{gp} regression \citep{Rasmussen2006}, the mean and variance of the latent field can be inferred as
\begin{align}
    \hat{\occ}(\lidarpoint{}{}) &= \kernelvec{}{\lidarpoint{}{}}{\lidarpoints{}{}}\left(\kernelmat{}{\lidarpoints{}{}}{\lidarpoints{}{}} + \sigma\mathbf{I}\right)^{-1} \mathbf{1},
    \label{eq:occ_inference}
    \\
    \text{var}(\hat{\occ}(\lidarpoint{}{})) &= \kernel{}{\lidarpoint{}{}}{\lidarpoint{}{}} - \kernelvec{}{\lidarpoint{}{}}{\lidarpoints{}{}}\left(\kernelmat{}{\lidarpoints{}{}}{\lidarpoints{}{}} + \sigma\mathbf{I}\right)^{-1}\kernelvec{}{\lidarpoint{}{}}{\lidarpoints{}{}}^\top,
    \nonumber
\end{align}
where $\kernelvec{}{\lidarpoint{}{}}{\lidarpoints{}{}}$ is the vector of covariance kernels between the query point $\lidarpoint{}{}$ and the input observation locations $\lidarpoints{}{}$, $\kernelmat{}{}{}$ the kernel-based covariance matrix of $\lidarpoints{}{}$, and $\sigma$ the observation noise.
Following \cite{legentil2024accurate}, the use of a `reverting function' $\revert$ over $\occ$ approximates the Euclidean distance field
\begin{equation}
    \dist(\lidarpoint{}{}) = \revert\left(\occ(\lidarpoint{}{})\right).
    \label{eq:dist}
\end{equation}
The definition of $\revert$ depends on the kernel function $\kernelraw{}$
\begin{equation}
    \revert\left(\kernel{}{\lidarpoint{}{}}{\lidarpoint{}{}'}\right) \triangleq \Vert\lidarpoint{}{} - \lidarpoint{}{}'\Vert.
\end{equation}
With an unscaled, stationary, isotropic kernel (that can be written as a function of the distance $\kernel{}{\lidarpoint{}{}}{\lidarpoint{}{}'} \rightarrow \kernelraw{}(\Vert\lidarpoint{}{} - \lidarpoint{}{}'\Vert)$), the reverting function is the inverse of $\kernelraw{}$.
Note that due to the non-linear nature of $\revert$, the distance field $\dist$ is not a \ac{gp}.
Accordingly, the inferred variance from \eqref{eq:occ_inference} is not really informative about the distance estimate's uncertainty.
While not used in 2Fast-2Lamaa, we propose a novel uncertainty proxy in Appendix~\ref{app:uncertainty}.

\subsection{Efficient large-scale GP distance field}
\label{sec:data_structure}
An issue with naively applying standard \ac{gp} regression is the cubic computational complexity $O(N^3)$ with respect to the number $N$ of points in the map due to the matrix inversion in \eqref{eq:occ_inference}.
In this work, we propose \edit{a novel and} efficient strategy for \ac{gp}-based distance field by computing local \acp{gp} instead of a single global one.
The core principle behind this strategy is the spatially limited impact of each input point due to the \emph{lengthscale} of the kernel.\footnote{Most of the commonly used kernel functions rely on a lengthscale hyperparameter that characterizes the spread of the covariance function. An intuitive way to think about lengthscale is that it somewhat represents the maximum gap in observations that can be interpolated through \ac{gp} regression.}
In other words, performing \ac{gp} regression considering a large-enough local neighbourhood of points will (locally) yield a similar inferred mean when compared with using all the available observations.

The key to our approach is to leverage both a sparse voxelized representation $\voxelmap$ of the dense 3D point cloud of the environment, and a spatial index $\spaceindex$ that allows for efficient point insertion/removal and closest/radius neighbour searches.\footnote{One can think about the spatial index as being a standard KD-Tree \citep{Bentley1975} however our implementation is based on a different data structure (cf. Section~\ref{sec:implementation}).}
The voxelized representation consists of a hashmap that maps 3-integer tuples to \emph{cells}, which store the centroid of all the lidar/map points that occurred in the voxel.
The tuples are the cell's grid index that can be computed directly from the 3D coordinates of a point and the fixed cell size.

Integrating new points in the map simply consists of checking if the corresponding cell index is already present in the hashmap.
If yes, the cell's centroid is updated incrementally with the new point location.
Otherwise, a new cell is created, and its position is added to the spatial index.
Thanks to the hashmap properties, accessing an existing cell or inserting a new one is constant-time $O(1)$ on average.
The insertion of a new cell in a spatial index is slower with an ${O}(\log(N))$ average complexity at best.
Fortunately, thanks to the voxelized nature of the stored information, only a small portion of the incoming lidar points lead to cell creation (for example, a \SI{7.9}{\km}-long suburban trajectory in the Boreas-RT dataset contains over 1.8 trillion lidar points, but the final number of cells in the map is around 10 million).

To query $\dist(\lidarpoint{}{})$ at any location $\lidarpoint{}{}$, the closest cell $\voxel_i$ in $\voxelmap$ is queried using the spatial index $\spaceindex$.
Then, the local neighbourhood around $\voxel_i$ is obtained through a radius search in $\spaceindex$.
This set of points represents the surface around the closest element to $\lidarpoint{}{}$ in the map. 
An estimate of $\dist(\lidarpoint{}{})$ can therefore be obtained using \eqref{eq:occ_inference} and \eqref{eq:dist} with the aforementioned local neighbourhood as input observations.
Considering the voxel nature of $\voxelmap$ and a fixed radius for the local neighbour search, there is a maximum of points $M$ used for \ac{gp} inference.
Knowing that a query of $\spaceindex$ is $O(\log(N))$ on average, the overall distance field inference is $O(\log(N) + M^3)$.
As $M$ is fixed, the global complexity for large-scale mapping only grows logarithmically with the size of the observed environment.

\begin{figure}
    \centering
    \def\imgwidth{0.65}
    \def\legendspace{0.0cm}
    \def\legendstyle{\footnotesize}
    \begin{tikzpicture}
        \node [inner sep = 0em] (pc) {\includegraphics[width=\imgwidth\columnwidth, clip, trim= 0cm 0cm 0cm 0cm]{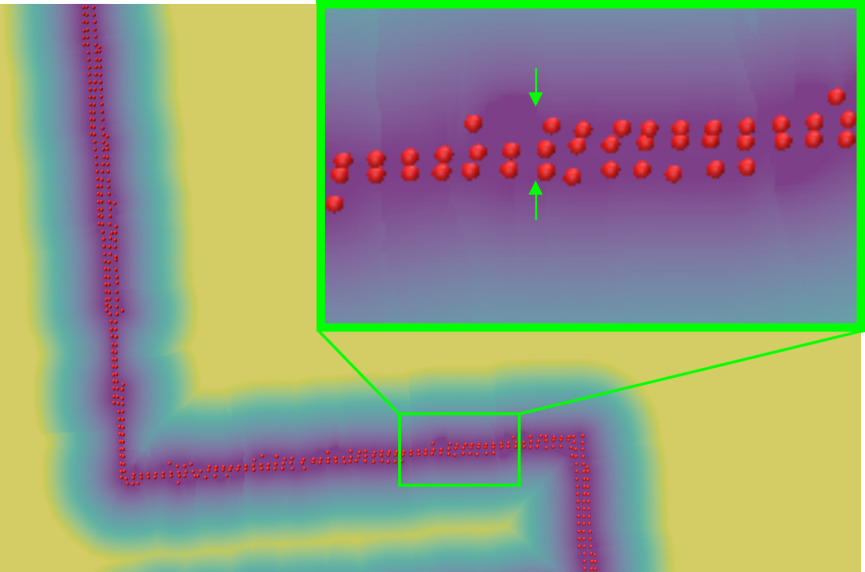}};
        \node [inner sep = 0em, below=0.5cm of pc] (mesh) {\includegraphics[width=\imgwidth\columnwidth, clip, trim= 0cm 0cm 0cm 0cm]{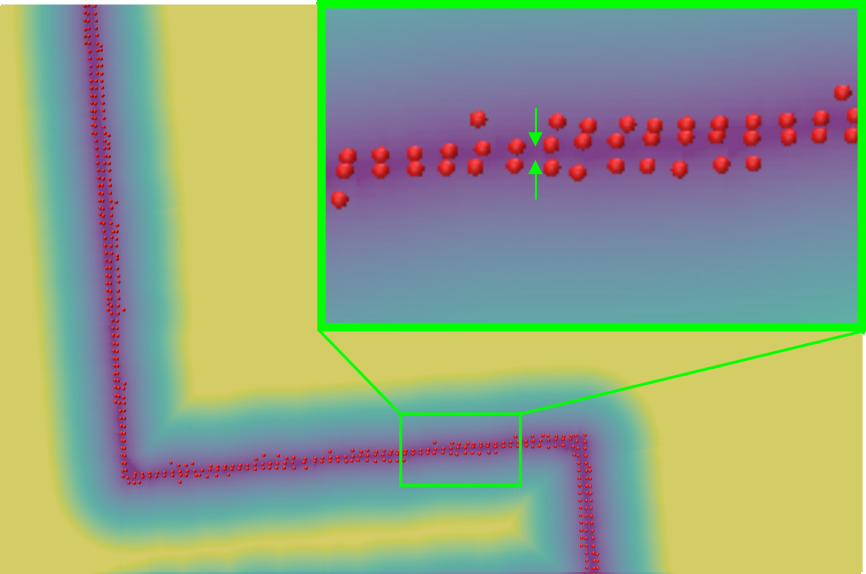}};
        \node [below=\legendspace of pc] {\legendstyle(a) Without weighting};
        \node [below=\legendspace of mesh] {\legendstyle(b) With proposed weighting};
    \end{tikzpicture}
    \caption{Illustration of the impact of the proposed weighting mechanism for our efficient GP-based distance field inference with sparse voxelized observations. The cell centroids are in red, and the colourmap represents the distance field (the colour is purposely saturated at 1m for the sake of visibility. (a) is the inference with equal weight for each voxel observation. (b) is the inference with the proposed weighting based on the number of lidar points that occurred in each cell. One can clearly see the improvement in terms of smoothness and `wall thickness', thus the accuracy of the field.}
    \label{fig:gp_weighting}
\end{figure}

\edit{There are two main caveats to the proposed field.
The first is linked to the use of local \acp{gp}, creating a discrete behaviour when potentially switching from one local \ac{gp} to the next while querying two locations that are side by side.
In other words, the global field is not guaranteed to be continuous over the full space $\reals^3$.
However, the `jumps' are generally small enough, in terms of value and gradient, that this does not prevent the use of the field in gradient-based optimization problems, as illustrated in this work.
The second drawback is that the voxelized representation degrades} the overall probabilistic properties of \ac{gp}-based distance inference.
Simply put, naively feeding the voxels' centroids to the \ac{gp} inference using $\sigma\mathbf{I}$ as the measurement noise gives the same importance to each cell.
However, not all cells provide the same amount of information: a cell in which only one point occurred should not have the same impact as a cell that averaged the position of 100 points.
Fig.~\ref{fig:gp_weighting}(a) shows the corresponding negative impact on the field: non-smooth surface and thicker walls.
To alleviate this issue, we propose to `weight' the cell centroid observations based on the number of points that occurred in each voxel.
Let us denote the per-voxel number of observations as the \emph{counter} $c_i$ of a cell.
Accordingly, in \eqref{eq:occ_inference}, we replace the measurement uncertainty model $\sigma\mathbf{I}$ with $\text{diag}(\mathbf{w})$ where $w_i$ (components of $\mathbf{w}$) are computed as a function of the cell's counter and the maximum cell counter in the neighbourhood.
Note that a lower $w_i$ means that the cell can be trusted more than a cell with a higher $w_i$.
Thus, the function that transforms $c_i$ to $w_i$ has to be decreasing.
We picked a decreasing sigmoid function in the shape of $\frac{1}{1+\exp(-c_i/c_{\max})}$.
Fig.~\ref{fig:gp_weighting}(b) illustrates the improved efficient field inference with the proposed weighting mechanism.
Appendix~\ref{app:field_accuracy} presents a brief quantitative analysis of the proposed efficient \ac{gp}-based distance field with and without cell weighting.

\subsection{Free-space carving}

In self-driving scenarios (among others), it is impossible to guarantee that the environment is static at the time of mapping.
Thus, a mechanism for rejecting dynamic objects from the map can help build cleaner maps for later reuse in a localization-only phase.
2Fast-2Lamaa includes a \emph{free-space carving} step to remove dynamic points previously inserted in the map and account for changes in the environment, such as parked cars that are no longer present.
The principle of the proposed mechanism relies on the comparison between the spherical projection of the current scan and that of the map cells in the vicinity of the current sensor position.
First, the scan is projected onto an image-like data structure by \edit{discretizing the azimuth and elevation of each point using a resolution of around \SI{1}{\degree}.
Each pixel retains the range value of the closest point that is projected into it.}
Then, the map points within a radius around the sensor position are queried and projected in the aforementioned image-like data structure using spherical coordinates.
\edit{If the range of a map point is smaller than the corresponding scan range, using a margin equal to the voxel resolution, it is} removed from the map as it occurred between a surface currently observed and the sensor.
Fig.~\ref{figure:free_space_carving} illustrates this process with dense simulated data for the sake of clarity.
Free-space carving can be performed online before inserting a freshly registered point cloud into the map, and offline by revisiting the whole trajectory and scans.

\begin{figure}
    \centering
    \input{figures/carving_figure}
    \caption{Illustration of the free-space carving process of the proposed mapping framework (using \cite{handa2014benchmark}'s living room simulated environment).}
    \label{figure:free_space_carving}
\end{figure}

\edit{
\section{Online pose-graph optimization}
}
\label{sec:loop}

\begin{figure}
    \centering
    \def\hdist{2em}
\def\vdistlong{2.5em}
\def\vdist{1.5em}
\def\blockheight{3.0em}
\def\blockwidth{20.0em}
\def\pad{1.0em}
\def\textsize{\scriptsize}
\begin{tikzpicture}[auto]
    \tikzstyle{input} = [draw, fill=white, rectangle, minimum height = 2.5em, text width = 12.5em,  minimum width = 12.8em, inner sep=0, outer sep=0, align = center, node distance = 5em, draw=red, execute at begin node=\setlength{\baselineskip}{8pt}\textsize]
    \tikzstyle{block} = [draw, fill=white, rectangle, minimum height = \blockheight, text width = \blockwidth,  minimum width = \blockwidth, align = center, inner sep=0, outer sep=0, node distance = 11em, execute at begin node=\setlength{\baselineskip}{8pt}\textsize] 
    \tikzstyle{output} = [draw=none, fill=white, text=NavyBlue, rectangle, minimum height = 2em, text width = 12.5em,  minimum width = 12.8em, align = center, node distance = 11em, execute at begin node=\setlength{\baselineskip}{8pt}\textsize]

    \tikzstyle{varrowleft} = [text width =0.5*\blockwidth, align=right, left, text width=((0.5*\blockwidth)), execute at begin node=\setlength{\baselineskip}{7pt}\textsize] 
    \tikzstyle{varrowright} = [text width =0.5*\blockwidth, align=left, right, text width=((0.5*\blockwidth)), execute at begin node=\setlength{\baselineskip}{7pt}\textsize] 
    \tikzstyle{harrowabove} = [text width = \hdist, align=center, above, execute at begin node=\setlength{\baselineskip}{7pt}\textsize] 
    \tikzstyle{harrowbelow} = [text width = \hdist, align=center, below, execute at begin node=\setlength{\baselineskip}{7pt}\textsize]

    \node[input] (submaps) {\textbf{2Fast-2Lamaa topometric mode}};

    \node[block, below=\hdist of submaps] (features) {\textbf{SIFT feature extraction}\\- Conversion to image-like gradient maps\\- Per-submap SIFT feature and descriptor computation };
    
    \node[block, below=\hdist of features] (se2) {\textbf{SE(2) registration}\\- Brute-force feature matching\\- RANSAC-based SE(2) alignment (threshold on inliers)};
    
    \node[block, below=\hdist of se2] (coarse) {\textbf{Coarse SE(3) loop-closure}\\- Elevation alignment (threshold on difference)\\- Combining with SE(2) transforms};
    
    \node[block, below=\hdist of coarse] (refinement) {\textbf{SE(3) loop-closure refinement}\\- Distance-field based registration};
    
    \node[block, below=\hdist of refinement] (posegraph) {\textbf{SE(3) pose-graph optimization}};

    \node[output, below=0.5*\hdist of posegraph] (out) {SE(3) globally consistent trajectory};

    \draw[->] (submaps) -- node[varrowright]{3D submaps and odometry poses} (features);
    \draw[->] ([yshift=-0.5*\vdist]submaps.south) -- ([yshift=-0.5*\vdist, xshift=-(0.5*\blockwidth+\pad)]submaps.south) -- ([xshift=-\pad]coarse.west) -- (coarse.west);
    \draw[->] ([xshift=-\pad]coarse.west) -- ([xshift=-\pad]refinement.west) -- (refinement.west);
    \draw[->] ([xshift=-\pad]refinement.west) -- ([xshift=-\pad]posegraph.west) -- (posegraph.west);
    \draw[->] (features) -- node[varrowright]{SIFT 2D features and descriptors} (se2);
    \draw[->] (se2) -- node[varrowright]{SE(2) loop-closure transforms} (coarse);
    \draw[->] (coarse) -- node[varrowright]{Coarse SE(3) loop-closure transforms} (refinement);
    \draw[->] (refinement) -- node[varrowright]{Fine SE(3) loop-closure transforms} (posegraph);
    \draw[->] (posegraph) -- (out);
%
    
\end{tikzpicture}
    \caption{Overview of the proposed loop closure detection and correction based on the output of 2Fast-2Lamaa in topometric mapping mode. First, the 3D submaps are projected into image-like elevation gradient maps (2.5D). Using visual features, the 2.5D submaps are coarsely aligned by SE(2) RANSAC registration and elevation alignment. Thresholds on the number of inliers and the elevation coherence are used to discard non-loop-closure submap pairs. Based on the proposed GP-based distance field, coarse loop-closure transforms are refined before being used as residuals in a pose-graph optimization that leverages the odometry pose estimates between consecutive submaps.}
    \label{fig:offline_posegraph}
\end{figure}

When used in mapping mode with a single global map incrementally built based on odometry estimates, \edit{the proposed scan-to-map strategy cannot guarantee global consistency when the robot's trajectory forms loops due to the inherent drift in any odometry algorithm.}
While using submaps in a topometric approach alleviates this issue for any robotic application that involves repeated motion (teach-and-repeat-like), one might be interested in obtaining a globally consistent trajectory for globally consistent mapping.
Toward this goal, 2Fast-2Lamaa also integrates an optional\footnote{In our experiments, this feature is only used for evaluation on the VBR-SLAM benchmark.} \edit{online} loop-closure detection and correction mechanism based on feature descriptor matching and 
pose-graph optimization, similar to the work by~\cite{giubilato2022gpgm}.

\begin{figure}
    \centering
    \def\legenddist{0.1cm}
    \def\vdist{0.0cm}
    \def\imgwidth{0.99\columnwidth}
    \def\hdist{0.02\columnwidth}
    \def\legendtextsize{\small}
    \begin{tikzpicture}
        \tikzstyle{legend} = [align = center, inner sep=0, outer sep=0, node distance = 0em, execute at begin node=\setlength{\baselineskip}{8pt}\small] 
        \tikzstyle{image} = [align = center, inner sep=0, outer sep=0, node distance = 0em] 

        \node[image] (features) {\includegraphics[clip, width=\imgwidth]{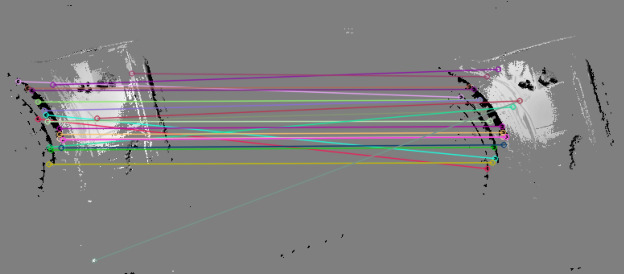}};
        \node[image, below=\vdist of features] (overlay) {\includegraphics[clip, width=\imgwidth]{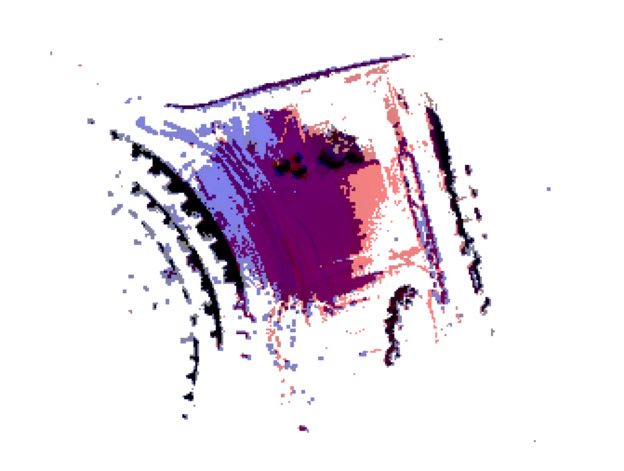}};

        \node[legend, below=\legenddist of features] {\legendtextsize (a) SIFT features matched between submaps};
        \node[legend, below=\legenddist of overlay] {\legendtextsize (b) Overlay after coarse SE(2) registration};
        
    \end{tikzpicture}
    \caption{Illustration of the loop-closure detection and SE(2) registration used in our \edit{online} pose-graph optimization of 2Fast-2Lamaa's submaps. The proposed pipeline relies on the extraction, matching, and registration of visual features in image-like elevation representations of each submap.}
    \label{fig:loop_match}
\end{figure}

Fig.~\ref{fig:offline_posegraph} shows the proposed pipeline.
First, 2Fast-2Lamaa's topometric-mapping mode is leveraged to obtain submaps and submap-to-submap transformation estimates $\transtilde{}{}{s_i}{s_{i\plus1}}$.
The voxelized representation of each \edit{gravity-aligned} submap is projected in individual image-like data structures which correspond to the \edit{\emph{elevation gradient} of the environment.
Unlike the work by~\cite{giubilato2022gpgm}, no \ac{gp} is used to generate the image-like data.
2Fast-2Lamaa uses a simple top-down orthographic projection.
The pixel size is one and a half times the voxel resolution.}
Then SIFT features \citep{lowe1999object} are extracted and matched between the current and past submaps as illustrated in Fig.~\ref{fig:loop_match}.
Loop-closure candidates are obtained by matching and registering features between submaps.
To avoid considering wrong associations and limit the computational burden, the brute-force feature matching is only performed between submaps \edit{for which pose-to-pose distances are under a threshold computed based on the typical odometry drift in percentage (user-defined) and} the distance travelled between pairs.
Given the SIFT associations of a submap pair, an SE(2) transformation between the two submaps is estimated with RANSAC \citep{fischler1981ransac} and elevation correction is performed so that the mean elevation is the same in both submaps.
Combining these together, a rough loop-closure SE(3) transformation is obtained.
Using the proposed continuous distance field~\eqref{eq:registration}, the SE(3) loop-closure constraint is refined \edit{individually between four scans of the oldest of the two matched submaps and the distance field of the current submap}.
Finally, the odometry relative pose estimates and the loop closure transformations are used in a pose-graph optimization.
This non-linear least-squares problem is solved with the Levenberg-Marquardt algorithm.

\section{Implementation}
\label{sec:implementation}

In this section, we present and discuss some specific details of our ROS2 C++ implementation.

\subsection{Low-level data structures}

The hashmaps in the work use the \texttt{ankerl::unordered\_dense::map} implementation \citep{ankerl2022hasmapgit}.
It has been benchmarked against numerous C++ hashmap implementations and displayed the best overall performance \citep{ankerl2022hasmapbenchmark}.
The cells are stored via pointers in the hashmap to ensure the compactness of the hashmap storage.

The constraints on the choice of the spatial index $\spaceindex$ are as follows:
\begin{itemize}
    \item Fast and scalable insertion of new elements.
    \item Possibility to remove points without the need to recreate the index.
    \item Allowing for N-closest neighbour and radius search.
\end{itemize}
The ikd-Trees \citep{cai2021ikd}, PH-Trees \citep{zaschke2014phtree}, \edit{and i-Octrees \citep{zhu2024ioctree}} match the aforementioned requirements.
Appendix~\ref{app:phtree} shows a toy-example evaluation. 
We observed a significant advantage for the \edit{i-Octree}.
For efficiency, the spatial indexing structure is not updated with the changing centroid of the voxels, \edit{but the hashmap is.}
The spatial index relies on the coordinates of the first point that occurred in each voxel.
Accordingly, the closest neighbour or radius searches performed with the `out-of-date' spatial index are only approximations.
We found this not to be a problem empirically, \edit{as the \acp{gp} are computed with the up-to-date voxel centroids of the local neighbourhood that is more than twice as large as the voxel resolution.
For some specific scenarios, such as queries in the middle of a corridor, the query for the closest voxel in the spatial index, whether it is out-of-date or up-to-date, can erroneously trigger \ac{gp} computations using points from the furthest of the two sides.
A simple workaround is to perform $K$ closest neighbour searches with associated local \ac{gp}-based distance field computation for each query.}
The final distance would be the smallest of the $K$ resulting \ac{gp}-inferred distances.
\edit{Note that this corridor-like scenario is unlikely when performing scan-to-map registration thanks to the accurate initial guess from 2Fast-2Lamaa's undistortion module.
Thus, our implementation uses $K=1$ for registration.
For other queries, the default is $K=2$.}

\subsection{Non-linear optimizations}

All the optimizations in this work (lidar-inertial undistortion and scan-to-map registration) are based on Ceres, an open-source non-linear least-squares solver \citep{ceres}.
Both \eqref{eq:odom_optimization} and \eqref{eq:registration} leverage Cauchy loss functions to attenuate the impact of outliers.
The analytical Jacobians of the residuals are provided to the solver.
To lower the computational cost of the overall pipeline, we adopted a basic key-framing strategy for the scan-to-map registration, performing \eqref{eq:registration} only when the sensor has moved sufficiently or after a fixed time period.

\edit{
\subsection{GP kernel and distance field}

While it was shown that the square-exponential provides the best accuracy for distance field estimation \citep{legentil2024accurate}, 2Fast-2Lamaa's implementation uses the rational-quadratic kernel $(1-\frac{\Vert\mathbf{x}-\mathbf{x}'\Vert^2}{2\alpha l^2})^{\minus\alpha}$ with $\alpha=2$ for efficiency, as it avoids the use of operations such as exponential or logarithm, which are computationally costly.
The lengthscale $l$ is defined as twice the voxel resolution.
To compute the local \ac{gp} around a voxel, the neighbourhood search uses a radius of two and a half times the voxel resolution.

\subsection{Point-cloud filtering}

Lidar data is generally not affected by different levels of luminosity in the environment.
However, it can be degraded by floating particles in the air, such as dust or precipitation.
To increase robustness in more challenging scenarios, 2Fast-2Lamaa offers the possibility to filter incoming point clouds based on the sensor-provided per-point intensity.
While not thoroughly analyzed in this paper, we have observed a significant improvement of 2Fast-2Lamaa's odometry and localization performance on the \emph{Farm} sequences of the Boreas-RT dataset \citep{lisus2026boreasrt}, which were collected on an extremely dusty road.
Additionally, a `density check' is always performed on each undistorted point cloud in case the intensity filtering does not sufficiently remove amorphous point clusters.
Using the user-provided map resolution, the scan is discretized into voxels.
Points that fall into voxels that have more than twelve neighbours are ignored for registration.

\subsection{Submapping strategy}

When used with submaps, as opposed to building a single global map, 2Fast-2Lamaa uses a very simple submapping strategy: a scan is integrated in a submap if the sensor has travelled less than a user-defined distance $\gamma$ since the submap creation.
Note that the submaps overlap by a ratio $\beta$ (equal to $0.2$ in our experiments).
Concretely, it means that a new submap is created when the distance travelled goes above $(1-\beta)\gamma$.
During the overlap, each scan is added to both submaps.
When performing topometric localization, the change from one submap to the next is done when the current pose estimate is closer to the poses in the second half of the overlap than to the first half.
}

\section{Experiments}

\begin{table*}
    \centering
    \caption{List and description of the different parameters that differ between the different datasets}
    \label{tab:params}
    \setlength{\tabcolsep}{2pt}
    \begin{tabularx}{\linewidth}{lYYYYYY}
        \toprule
         & \textbf{Voxel size} & \textbf{Global/submaps} & \textbf{Keyframing} & \textbf{Num pts registration} & \textbf{Planar only} & \textbf{Free-space carving}
        \\
        \midrule
        \textbf{Description}
        &  \scriptsize Size of the voxels for lidar feature extraction and map
        & \scriptsize Selection between the use of a single global map or submaps of given length $\gamma$ 
        & \scriptsize Travelled distance, orientation, and time threshold for keyframing scan-to-map registration
        & \scriptsize Maximum number of points for scan-to-map registration
        & \scriptsize Using only planar features
        & \scriptsize Radius for free-space carving
        \\
        \midrule
        
        \textbf{Boreas / Boreas-RT} & \scriptsize\SI{0.30}{\m} & \scriptsize Submaps \SI{300}{\m} & \scriptsize\SI{10}{\m}/\SI{15}{\degree}/\SI{0.5}{\s} & \scriptsize 8000 & \scriptsize No & \scriptsize\SI{50}{\m}
        \\
        \textbf{VBR} & \scriptsize\SI{0.30}{\m} & \scriptsize Submaps \SI{50}{\m} + online pose-graph & \scriptsize No & \scriptsize2000 & \scriptsize No & \scriptsize\SI{50}{\m}
        \\
        \textbf{Newer College} & \scriptsize\SI{0.15}{\m} & \scriptsize Global map & \scriptsize No & \scriptsize 2000 & \scriptsize Yes & \scriptsize No
        \\
        \bottomrule
    \end{tabularx}
\end{table*}

We validate the proposed framework for both localization and odometry using various public datasets as follows: Section~\ref{sec:exp_boreas} uses the Boreas dataset; Section~\ref{sec:exp_self} is based on \edit{part of the Boreas-RT dataset}; and Section~\ref{sec:exp_newer} the Newer College dataset.
\edit{We also leverage the VBR dataset for \ac{slam} evaluation in Section~\ref{sec:exp_vbr}.}
On top of this extensive benchmarking, we conduct an ablation using part of the Boreas-RT dataset in Section~\ref{sec:exp_ablation}.
Finally, Section~\ref{sec:exp_computation} provides the reader with information about the computational requirements of 2Fast-2Lamaa.

\edit{
\subsection{Parameters}
Aside from hardware-related variables such as extrinsic calibrations or maximum lidar range, 2Fast-2Lamaa can be tailored to specific applications or environments through a set of user-defined parameters.
Table~\ref{tab:params} provides an overview of the parameters that differ from dataset to dataset.
Notably, the voxel size is smaller for the Newer College dataset, and a single global map is used as the scale of the environment is smaller than in other datasets, and the trajectory does not form large loops with loss of covisibility.
As the VBR dataset contains some indoor and handheld sequences, we use a shorter submap length when compared to the Boreas datasets to increase the potential for submap-to-submap loop closures.
Not that the VBR dataset evaluation is the only set of experiments that leverages the proposed loop-closure mechanism.
As the Boreas and Boreas-RT datasets' \ac{imu} are of good quality, the optimization-based undistortion provides good odometry information. 2Fast-2Lamaa thus leverages a keyframing strategy to limit the number of scan-to-map registrations, allowing for the use of a higher number of points for registration.
Finally, free-space carving is disabled in the Newer College dataset evaluation as very few dynamic objects are present in the scene.
Please note that 2Fast-2Lamaa exposes many more parameters for the sake of generality, but only the ones mentioned here have been changed in our evaluation.  
}
Performance improvements could be attained by tuning parameters for each specific sequence type.

\subsection{Boreas dataset}
\label{sec:exp_boreas}

\subsubsection{Dataset description:}

\begin{figure}
    \centering
    \def\legenddist{0.1cm}
    \def\vdist{0.0cm}
    \def\imgwidth{0.49\columnwidth}
    \def\hdist{0.02\columnwidth}
    \def\legendtextsize{\small}
    \begin{tikzpicture}
        \tikzstyle{legend} = [align = center, inner sep=0, outer sep=0, node distance = 0em, execute at begin node=\setlength{\baselineskip}{8pt}\small] 
        \tikzstyle{image} = [align = center, inner sep=0, outer sep=0, node distance = 0em] 

        \node[image] (clearA) {\includegraphics[clip, width=\imgwidth]{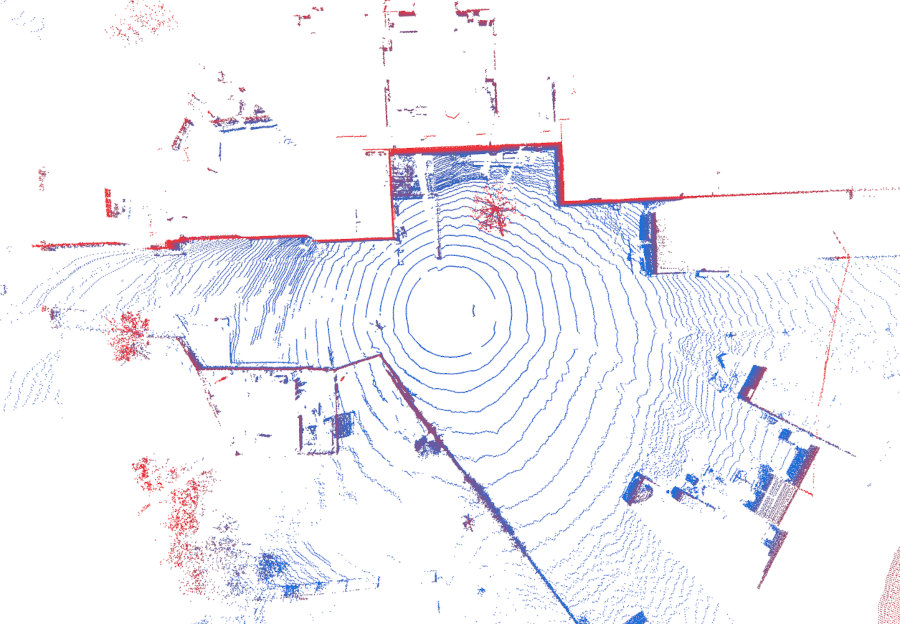}};
        \node[image, below=\vdist of clearA] (clearB) {\includegraphics[clip, width=\imgwidth]{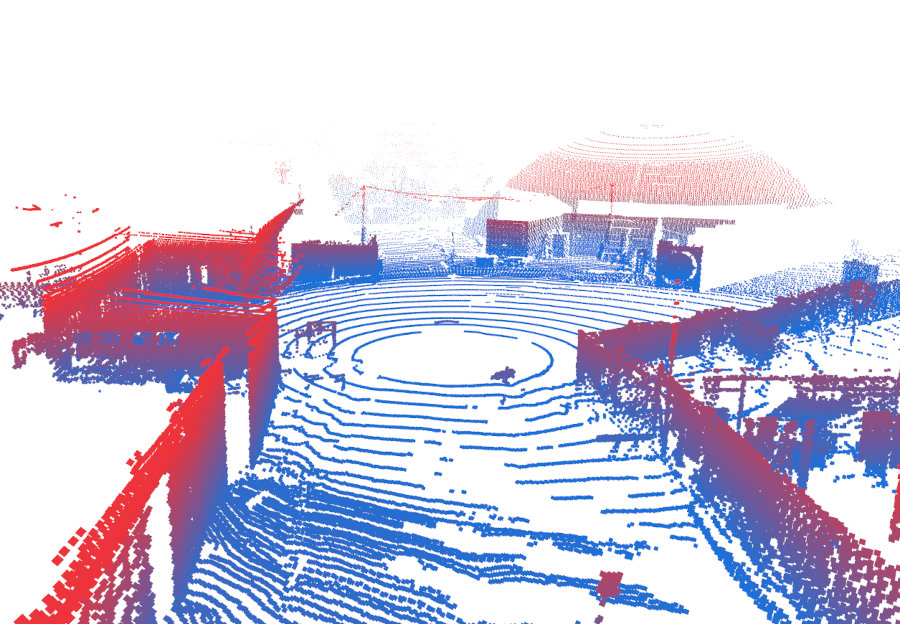}};
        
        \node[image, right=\hdist of clearA] (snowA) {\includegraphics[clip, width=\imgwidth]{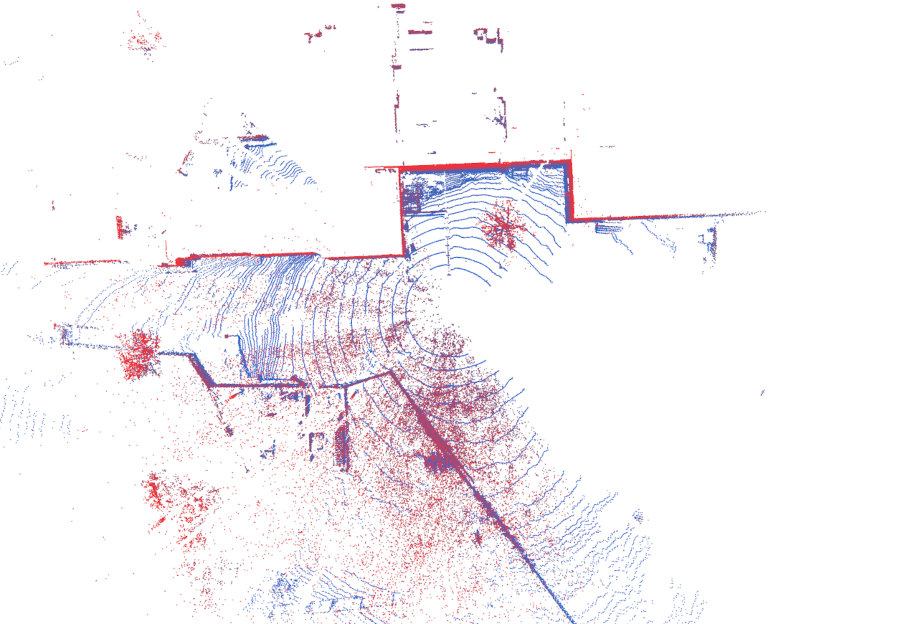}};
        \node[image, below=\vdist of snowA] (snowB) {\includegraphics[clip, width=\imgwidth]{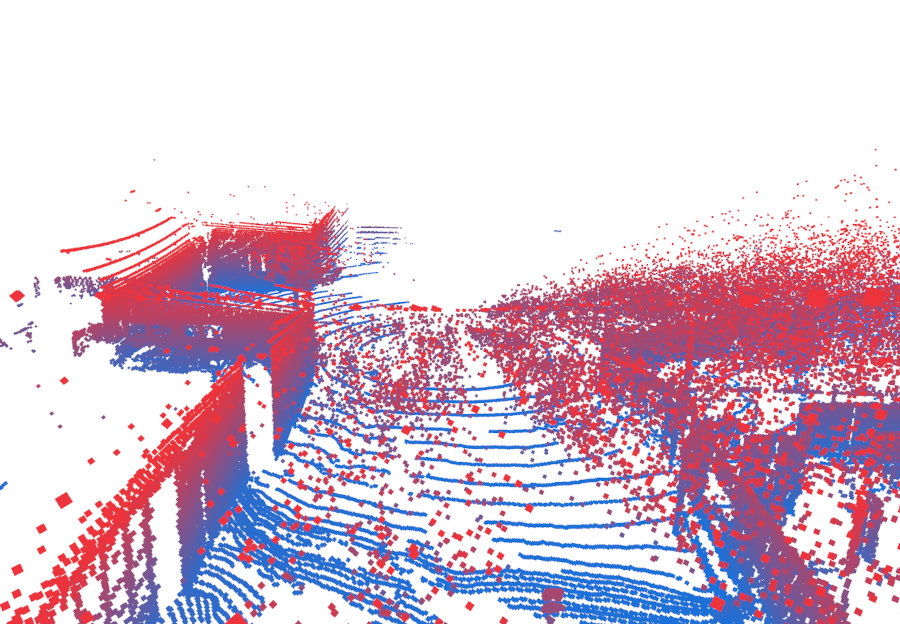}};

        \node[legend, below=\legenddist of clearA] {\legendtextsize (a) Clear weather (view A)};
        
        \node[legend, below=\legenddist of snowA] {\legendtextsize (b) Snowstorm (view A)};
        
        \node[legend, below=\legenddist of clearB] {\legendtextsize (c) Clear weather (view B)};
        
        \node[legend, below=\legenddist of snowB] {\legendtextsize (d) Snowstorm (view B)};
    \end{tikzpicture}
    \caption{Impact of a snowstorm on the lidar data in the Boreas dataset (sequences \texttt{2020-12-04-14-00} and \texttt{2021-01-26-10-59}). Ice has formed on the front-facing part of the lidar, completely blocking part of the sensor's FoV. Additionally, falling snow creates many points floating in the air (z-colored), making state estimation challenging.}
    \label{fig:boreas_snow}
\end{figure}

The Boreas dataset \citep{burnett2023boreas} consists of 44 sequences collected along a single \SI{8}{\km} suburban route with an automotive sensing platform over the course of a year.
The particularity of this dataset is the wide range of weather conditions in between different sequences, from bright sunny days to heavy rainfall and snowstorms.
The vehicle is equipped with a Velodyne Alpha Prime lidar, a Navtech RAS6 imaging radar, a FLIR Blackfly S camera, and an Applanix RTK-GNSS-IMU ground-truthing solution.
Fig.~\ref{fig:boreas_snow} illustrates the impact of extreme weather conditions on lidar data.
The Boreas dataset withholds the ground-truth trajectory for 13 of the 44 sequences, corresponding to more than \SI{100}{\km} of data collection.
These are used to benchmark various state estimation methods with a public leaderboard.
While not optimal due to data contamination between the ground-truth solution and the input of the various algorithms, the Applanix \ac{imu} is used in conjunction with the lidar in the rest of that section, as it is the only inertial sensor present aboard the vehicle.

\subsubsection{Baselines and metrics:}

For odometry, we report the leaderboard's KITTI odometry metric.
Succinctly, this metric aligns trajectory chunks (every 10 lidar frames) of length [$\SI{100}{\m},\ \SI{200}{\m}, \dots,\ \SI{800}{\m}$] based on the first pose of each chunk, computes the error in position and orientation at the end of each segment, and reports the average translation and rotation error relative to the distance travelled.
By submitting our results to the leaderboard, we benchmark our method against LTR \citep{burnett2022arewe}, STEAM-LIO \citep{burnett2024continuous}, and OG \citep{legentil2025dowe}.
Both LTR and STEAM-LIO are based on continuous-time \ac{icp} for lidar state estimation.
The continuous trajectory formulation leverages efficient \acp{gp} with noise-on-acceleration motion priors.
A major difference is the integration of inertial measurements in STEAM-LIO.
The OG baseline does not use the lidar but solely the 3D gyroscope and wheel encoder.
It represents the simplest and most efficient method for automotive odometry.
Nonetheless, OG provides odometry accuracy on par with state-of-the-art exteroceptive frameworks. 

To assess the localization ability of 2Fast-2Lamaa, we leverage the localization part of the leaderboard.
It consists of creating a map of the route given one sequence, and localizing within that map using 10 other sequences.
The metric is the \ac{rmse} of the relative pose between the trajectory of the mapping sequence and the localization ones.
Note that only LTR has been submitted to the Boreas benchmark (more baselines are used with a subset of the Boreas-RT dataset in Section~\ref{sec:exp_self}).
LTR's strategy for localization is performing scan-to-local-map registration while accounting for past estimates and odometry in the form of a prior on the state.

\subsubsection{Odometry benchmark:}

\begin{table}
    \centering
    \caption{Average relative pose accuracy of the proposed method and several baselines on the Boreas dataset leaderboard \citep{burnett2023boreas}.}
    \setlength{\tabcolsep}{2pt}
    \begin{tabularx}{\linewidth}{Lcc}
        \toprule
        \textbf{Method} &  \textbf{Trans.} \scriptsize[\%] & \textbf{Rot.} \scriptsize[$\si{\degree}/100\,\si{\m}$]
        \\
        \midrule
        \textbf{OG} \scriptsize\citep{legentil2025dowe} & 0.54 & \textbf{0.07}
        \\
        \textbf{LTR} \scriptsize\citep{burnett2022arewe} & 0.54 & 0.16
        \\
        \textbf{STEAM-LIO} \scriptsize\citep{burnett2024continuous} & 0.45 & 0.15
        \\
        \textbf{2Fast-2Lamaa} \scriptsize(ours) & \textbf{0.27} & 0.10 
        \\
        \bottomrule
    \end{tabularx}
    \label{tab:odom_boreas}
\end{table}

Table~\ref{tab:odom_boreas} reports the odometry accuracy of 2Fast-2Lamaa and the different baselines.
Except for OG, which displays a very small rotational error, 2Fast-2Lamaa is the most accurate method, significantly outperforming the other frameworks.
In this experiment, 2Fast-2Lamaa uses submaps (topometric mapping) as it provides slightly better odometry results than the global map version \edit{on the Boreas-RT dataset}.
Section~\ref{sec:exp_ablation} provides more details about this difference.
As shown in Table~\ref{tab:boreas_detailed}, 2Fast-2Lamaa's accuracy is consistent throughout the different sequences, except for \texttt{2021-01-26-10-59}, which has been collected in a snowstorm and ice accumulated on the front-facing part of the lidar, thus blocking a significant portion of the sensor's field of view (cf. Fig.~\ref{fig:boreas_snow}).
As all the methods provide decent odometry estimates, it is difficult to draw any strong conclusion on why our method performs better.
Note that STEAM-LIO and LTR elegantly perform motion-distortion correction as part of their continuous-time state estimation without `fixing' the undistortion in an open-loop fashion based on previous state estimates.
Thus, the difference in the undistortion strategy is unlikely to explain the difference in the global performance.
We believe that our continuous map representation makes a difference, as LTR and STEAM-LIO rely on discrete downsampled point clouds with normals to represent the environment.
The impact of the continuous map is investigated in Section~\ref{sec:exp_ablation}.

\begin{table}
    \centering
    \caption{Per-sequence odometry error (KITTI metric) of 2Fast-2Lamaa on the Boreas dataset leaderboard.}
    \label{tab:boreas_detailed}
    \setlength{\tabcolsep}{2pt}
    \begin{tabularx}{\linewidth}{lYYY}
        \toprule
        \multicolumn{1}{c}{\textbf{\shortstack{Sequence\\ID}}} & \multicolumn{1}{c}{\textbf{\shortstack{Weather\\conditions}}} & \multicolumn{1}{c}{\shortstack{\textbf{Trans.}\\\textbf{err.} \scriptsize[\%]}} & \multicolumn{1}{c}{\shortstack{\textbf{Rot. err.} \\ \scriptsize[$\si{\degree}/100\,\si{\m}$]}}
        \\
        \midrule
        2020-12-04-14-00 & \FilledCloud\Snow & 0.20 & 0.07
        \\ 
        2021-01-26-10-59 & \FilledSnowCloud\Snow\Snow & 0.36 & 0.14
        \\ 
        2021-02-09-12-55 & \FilledSunCloud\Snow & 0.21 & 0.08
        \\ 
        2021-03-09-14-23 & \Sun & 0.26 & 0.10
        \\ 
        2021-04-22-15-00 & \FilledSnowCloud & 0.22 & 0.08
        \\ 
        2021-06-29-18-53 & \FilledRainCloud & 0.28 & 0.10
        \\ 
        2021-06-29-20-43 & \Cloud\HalfSun & 0.37 & 0.12
        \\ 
        2021-09-08-21-00 & \NoSun & 0.31 & 0.11
        \\ 
        2021-09-09-15-28 & \SunCloud & 0.32 & 0.12
        \\ 
        2021-10-05-15-35 & \FilledCloud & 0.30 & 0.10
        \\ 
        2021-10-26-12-35 & \FilledRainCloud & 0.26 & 0.09
        \\ 
        2021-11-06-18-55 & \NoSun & 0.25 & 0.09
        \\ 
        2021-11-28-09-18 & \FilledSnowCloud\Snow\Snow & 0.21 & 0.08
        \\
        \bottomrule
        \multicolumn{4}{c}{\FilledCloud: Overcast, \Snow: Snow coverage, \Snow\Snow: High snow coverage}
        \\
        \multicolumn{4}{c}{ \FilledSnowCloud: Snowing, \Sun: Sun, \FilledRainCloud: Rain, \HalfSun: Dusk, \NoSun: Night}
    \end{tabularx}
\end{table}

\subsubsection{Localization benchmark:}

For consistency with the odometry benchmark and to match the topometric nature of LTR's maps, we used 2Fast-2Lamaa with submaps (as opposed to a single global map).
2Fast-2Lamaa's submaps correspond to \SI{300}{\m} of trajectory while LTR uses a threshold of \SI{30}{\m} to create a new submap.
Table~\ref{tab:localization_boreas} provides the leaderboard results.
LTR provides slightly better performance over most of the axes.
We believe that the state prior in LTR's formulation helps stabilize the pose estimate between consecutive scans, thus providing more accurate localization abilities.
This is supported by the higher roll and pitch rotational error of 2Fast-2Lamaa: our pure scan-to-map registration does not enforce any motion model nor leverage information from previous state estimates.
Analyzing the per-sequence results for 2Fast-2Lamaa, there is no strong correlation between both the lateral and longitudinal error and the weather conditions.
However, sequences with snow display a notably higher error along the vertical axis.
This is expected as the mapping sequence \texttt{boreas-2020-11-26-13-58} is free from snow.
Accordingly, the snow cover on the ground creates a vertical offset in the estimated poses.
Overall, despite slightly worse performance compared to LTR, both methods offer high levels of accuracy with under \SI{5}{\cm} lateral \ac{rmse}, which is less than the width of any road lane marking.

\begin{table*}
    \centering
    \caption{RMSE localization error for 2Fast-2Lamaa and LTR on the Boreas dataset leaderboard.}
    \setlength{\tabcolsep}{2pt}
    \begin{tabularx}{\linewidth}{lYYYYYY}
        \toprule
        \textbf{Method} & \textbf{Long.} \scriptsize[m] & \textbf{Lat.} \scriptsize[m] & \textbf{Vert.} \scriptsize[m] & \textbf{Roll} \scriptsize[$^\circ$] & \textbf{Pitch} \scriptsize[$^\circ$] & \textbf{Yaw} \scriptsize[$^\circ$] 
        \\
        \midrule
        \textbf{LTR} \scriptsize\citep{burnett2022arewe} & \textbf{0.039} & \textbf{0.031} & \textbf{0.055} & \textbf{0.043} & \textbf{0.025} & 0.040
        \\
        \textbf{2Fast-2Lamaa} \scriptsize topometric (ours) & 0.048 & 0.046 & 0.061 & 0.080 & 0.037 & \textbf{0.039}
        \\
        \midrule
        $\downarrow$ \textbf{2Fast-2Lamaa} per sequence
        \\
        \midrule
        2020-12-04-14-00 \quad\qquad$\:\:$\FilledCloud\Snow     &0.068&0.076&0.108&0.071&0.037&0.039 \\ 
        2021-01-26-10-59 \quad\qquad\FilledSnowCloud\Snow\Snow  &0.040&0.030&0.132&0.103&0.093&0.038 \\ 
        2021-02-09-12-55 \quad\qquad$\:$\FilledSunCloud\Snow    &0.029&0.031&0.037&0.077&0.034&0.039 \\ 
        2021-03-09-14-23 \quad\qquad$\quad$\Sun                 &0.037&0.029&0.045&0.079&0.034&0.031 \\ 
        2021-06-29-18-53 \quad\qquad$\quad$\FilledRainCloud     &0.079&0.058&0.058&0.097&0.045&0.051 \\ 
        2021-09-08-21-00 \quad\qquad$\quad$\NoSun               &0.078&0.062&0.065&0.103&0.042&0.049 \\ 
        2021-10-05-15-35 \quad\qquad$\quad$\FilledCloud         &0.048&0.050&0.047&0.086&0.039&0.035 \\ 
        2021-10-26-12-35 \quad\qquad$\quad$\FilledRainCloud     &0.057&0.059&0.047&0.087&0.043&0.043 \\ 
        2021-11-06-18-55 \quad\qquad$\quad$\NoSun               &0.040&0.044&0.057&0.074&0.036&0.036 \\ 
        2021-11-28-09-18 \quad\qquad\FilledSnowCloud\Snow\Snow  &0.043&0.033&0.074&0.075&0.038&0.032 \\
        \bottomrule
        \multicolumn{7}{c}{\FilledCloud: Overcast, \Snow: Snow coverage, \Snow\Snow: High snow coverage, \FilledSnowCloud: Snowing, \Sun: Sun, \FilledRainCloud: Rain, \HalfSun: Dusk, \NoSun: Night}
    \end{tabularx}
    \label{tab:localization_boreas}
\end{table*}

\subsection{Boreas-RT}
\label{sec:exp_self}

\subsubsection{Dataset description:}

\edit{To evaluate the proposed framework, we use data from the Boreas-RT dataset~\citep{lisus2026boreasrt}.
It uses the same sensing platform as the Boreas dataset with} the main hardware differences being a different radar firmware and an additional 6-\ac{dof} Silicon Sensing DMU41 \ac{imu} to prevent data contamination between ground-truth and raw sensor data.
\edit{In this section, we leverage 16 sequences in four different environments, totalling \SI{120}{\km} of data for benchmarking against various baselines.}
The different routes are \emph{Suburbs}, \emph{Regional}, \emph{Tunnel}, and \emph{Skyway}, listed in increasing level of difficulty.
\edit{We also provide 2Fast-2Lamaa's results on all of the 60 sequences of Boreas-RT ($\approx$\SI{650}{\km}) in Appendix~\ref{app:boreasrt}.}

\begin{figure*}
    \centering
    \def\legenddist{0.1cm}
    \def\vdist{0.6cm}
    \def\imgwidth{0.49\columnwidth}
    \def\hdist{0.02\columnwidth}
    \def\legendtextsize{\small}
    \begin{tikzpicture}
        \tikzstyle{legend} = [align = center, inner sep=0, outer sep=0, node distance = 0em, execute at begin node=\setlength{\baselineskip}{8pt}\small] 
        \tikzstyle{image} = [align = center, inner sep=0, outer sep=0, node distance = 0em] 

        \node[image] (subcam) {\includegraphics[clip, width=\imgwidth]{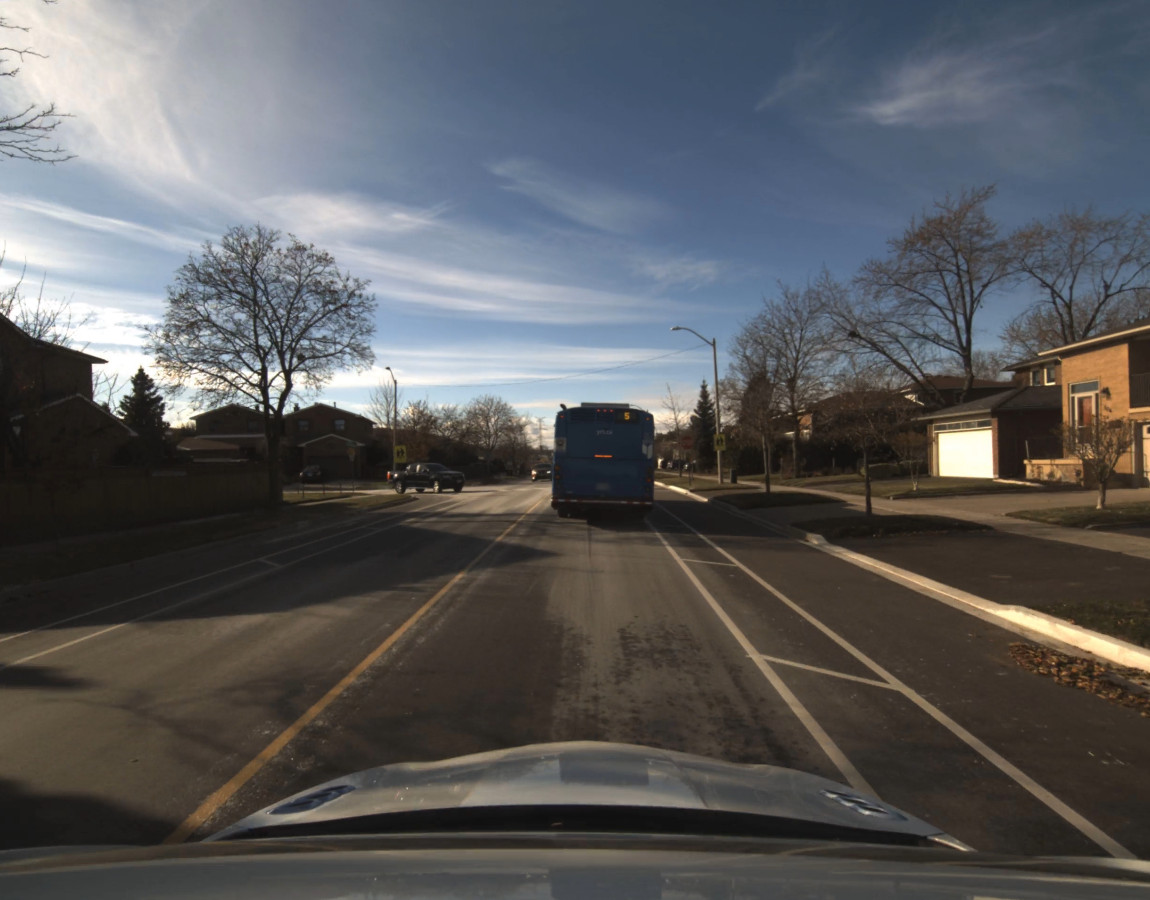}};
        \node[image, below=\vdist of subcam] (submap) {\includegraphics[clip, width=\imgwidth]{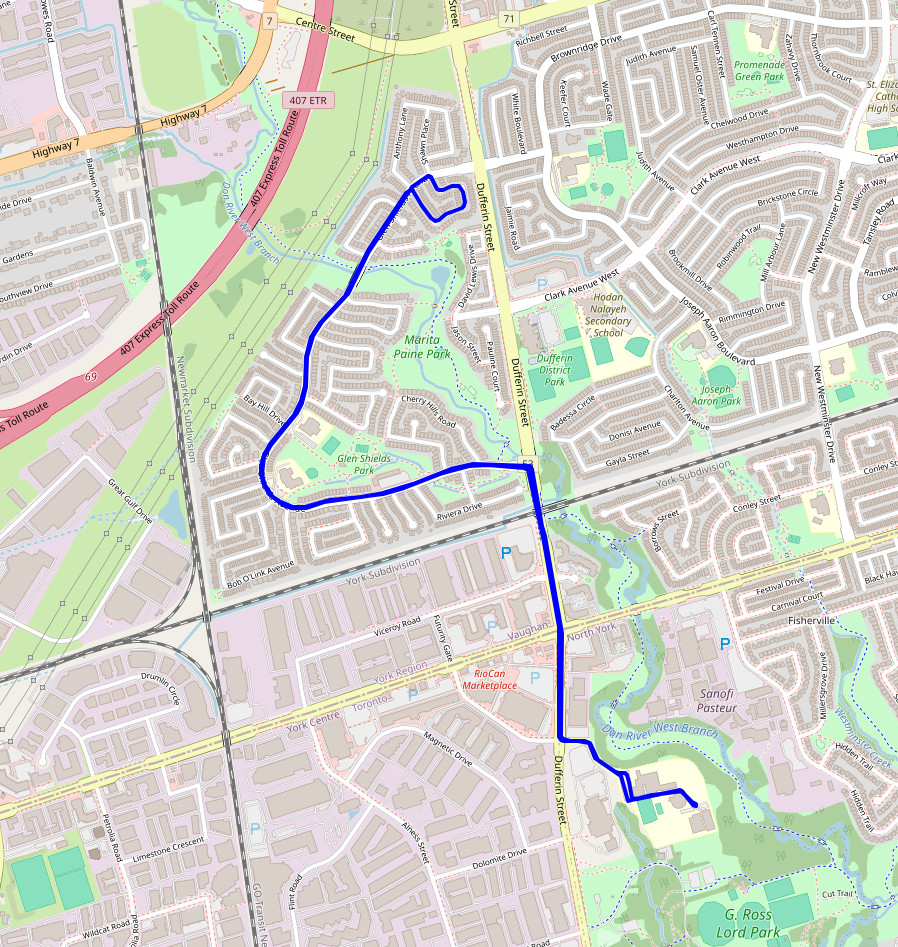}};
        
        \node[image, right=\hdist of subcam] (higcam) {\includegraphics[clip, width=\imgwidth]{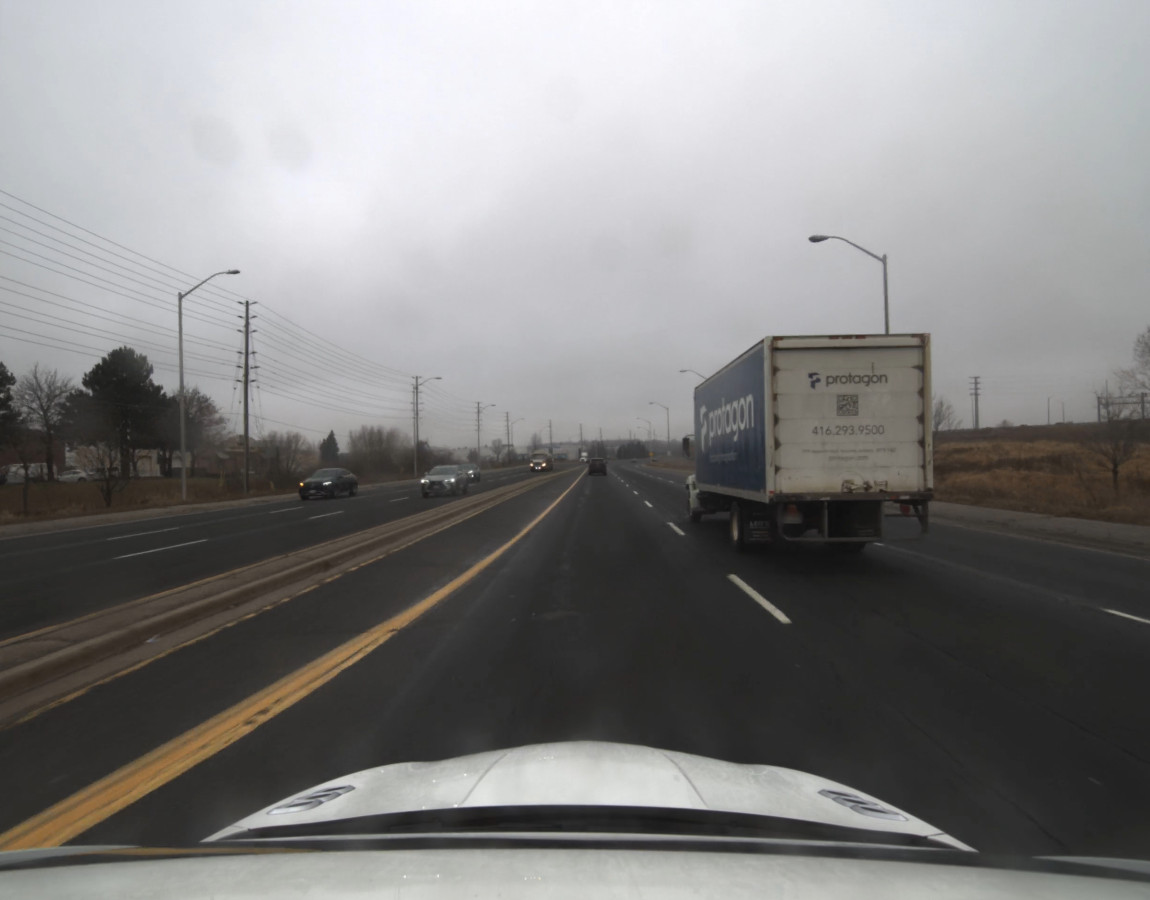}};
        \node[image, below=\vdist of higcam] (higmap) {\includegraphics[clip, width=\imgwidth]{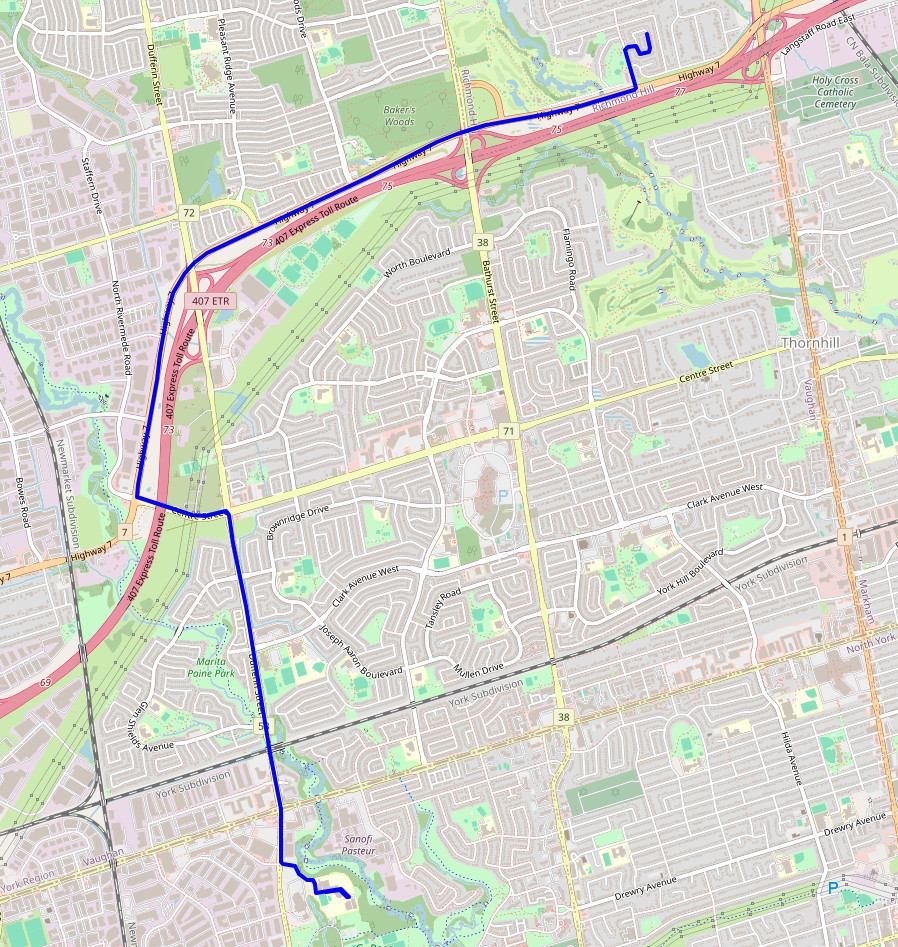}};
        
        \node[image, right=\hdist of higcam] (tuncam) {\includegraphics[clip, width=\imgwidth]{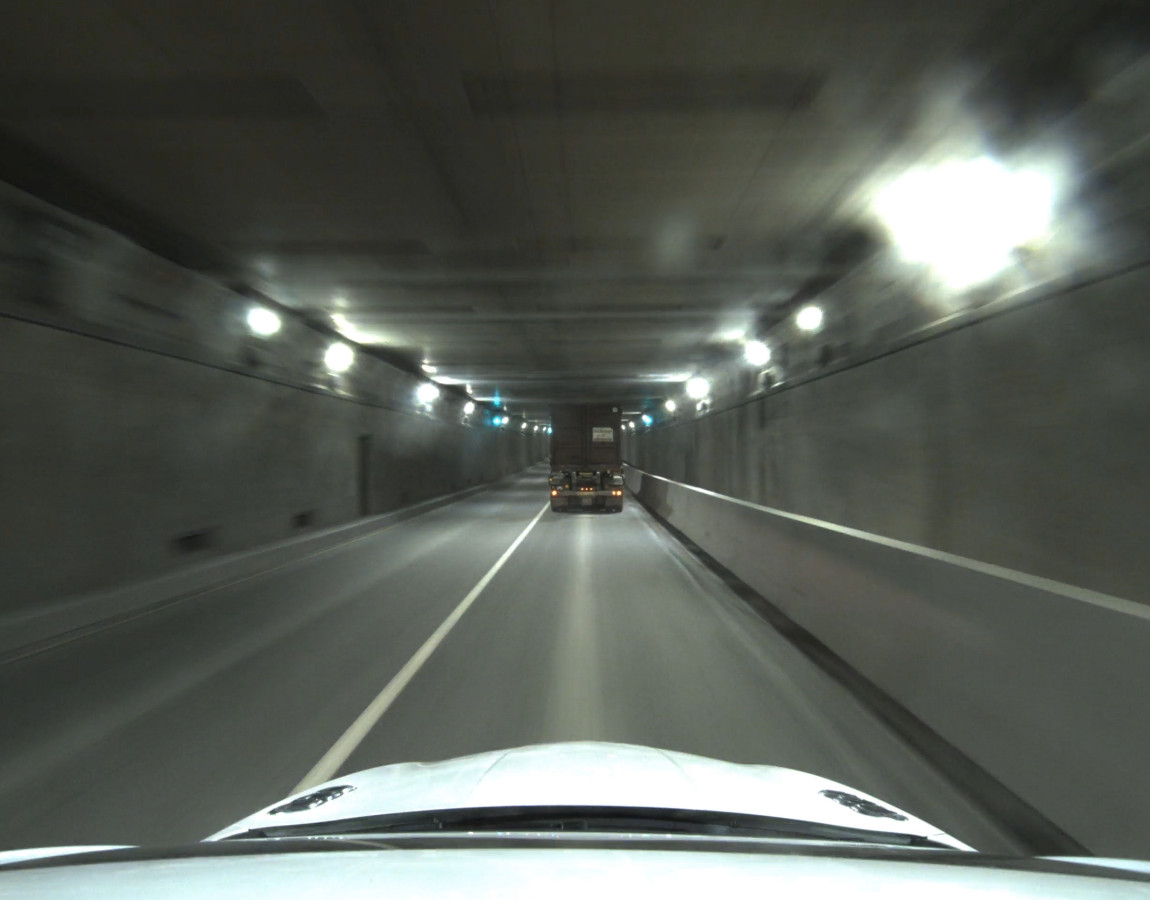}};
        \node[image, below=\vdist of tuncam] (tunmap) {\includegraphics[clip, width=\imgwidth]{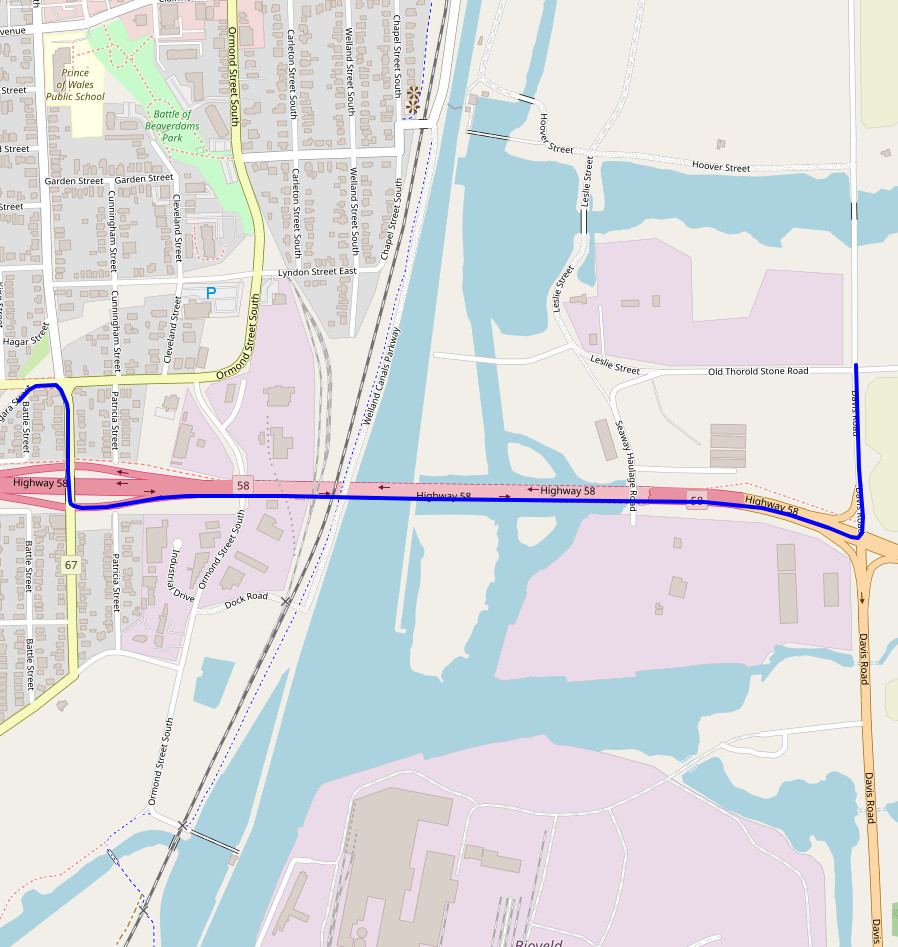}};
        
        \node[image, right=\hdist of tuncam] (skycam) {\includegraphics[clip, width=\imgwidth]{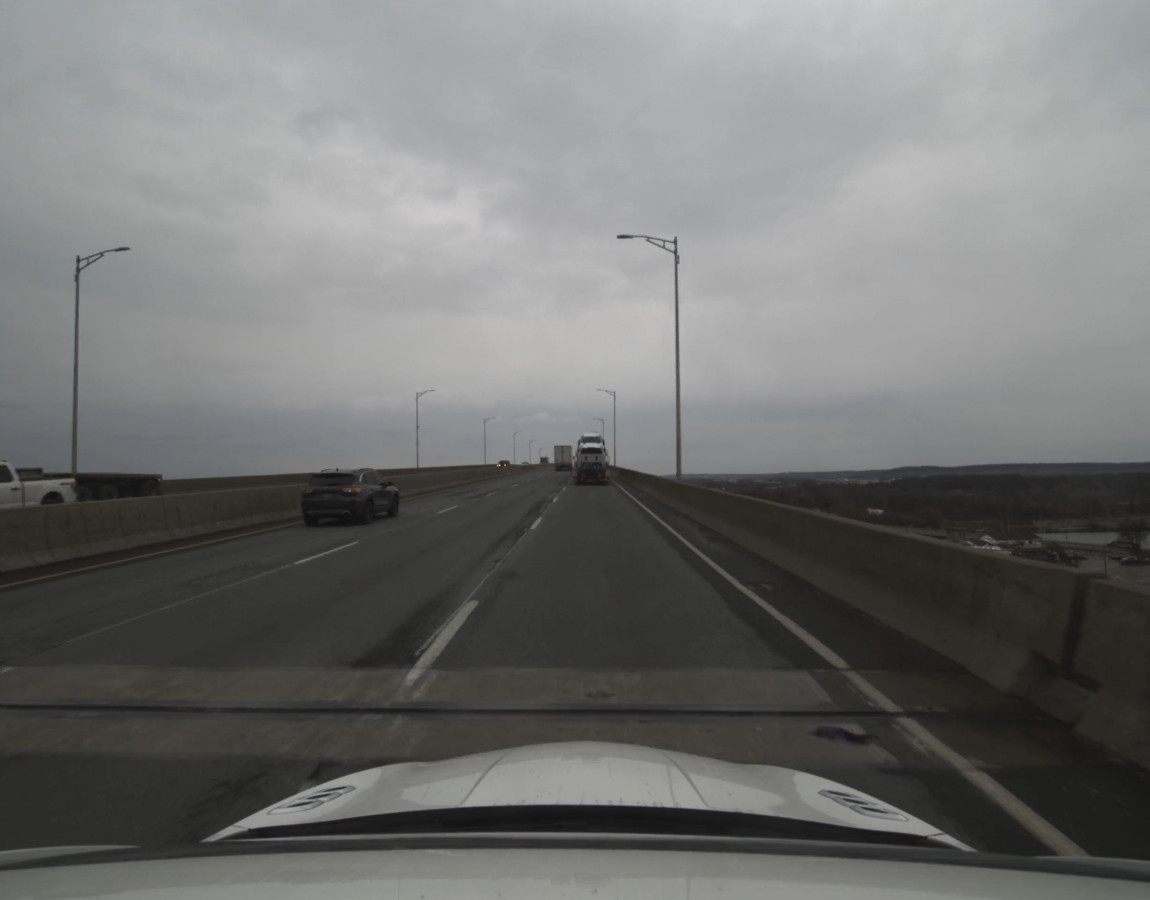}};
        \node[image, below=\vdist of skycam] (skymap) {\includegraphics[clip, width=\imgwidth]{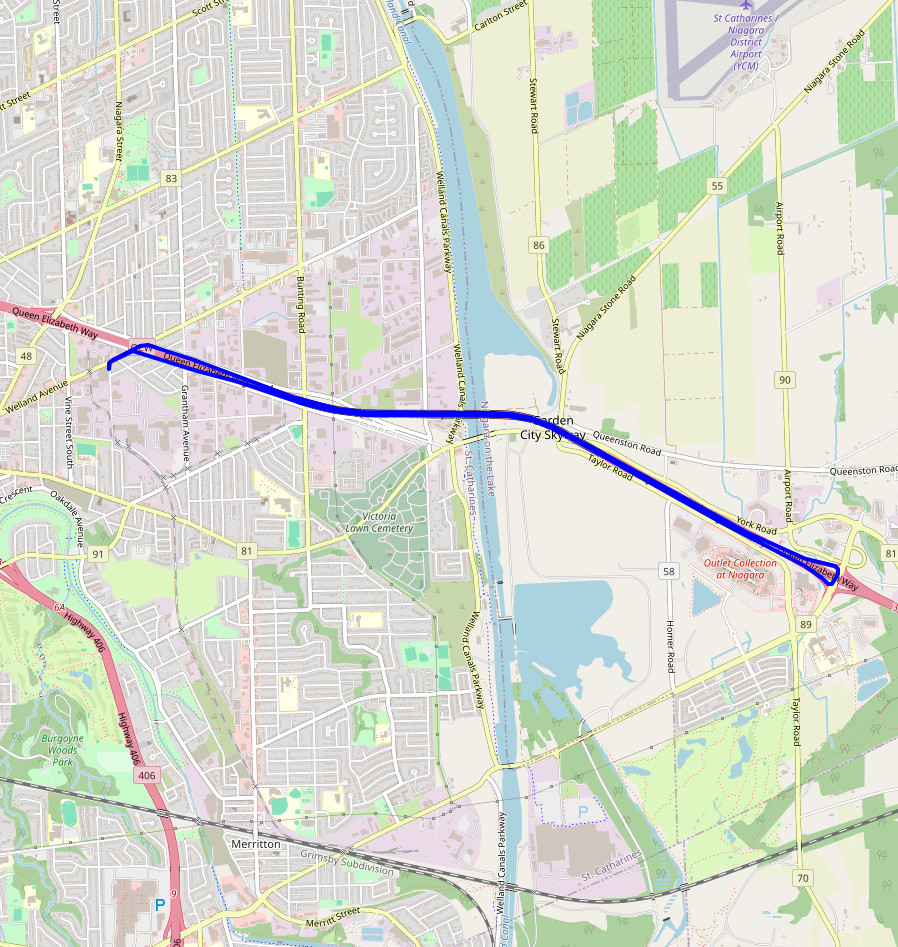}};

        \node[legend, below=\legenddist of subcam] {\legendtextsize (a) Suburbs};
        \node[legend, below=\legenddist of submap] {\legendtextsize (e) Suburbs trajectory};
        
        \node[legend, below=\legenddist of higcam] {\legendtextsize (b) Regional};
        \node[legend, below=\legenddist of higmap] {\legendtextsize (f) Regional trajectory};
        
        \node[legend, below=\legenddist of tuncam] {\legendtextsize (c) Tunnel};
        \node[legend, below=\legenddist of tunmap] {\legendtextsize (g) Tunnel trajectory};
        
        \node[legend, below=\legenddist of skycam] {\legendtextsize (d) Skyway};
        \node[legend, below=\legenddist of skymap] {\legendtextsize (h) Skyway trajectory};
    \end{tikzpicture}
    \caption{Images from the onboard camera of the Boreas-RT dataset and trajectory overlays (blue lines) on OpenStreetMap.}
    \label{fig:self_data_sample}
\end{figure*}

\begin{figure}
    \centering
    \includegraphics[width=0.99\columnwidth]{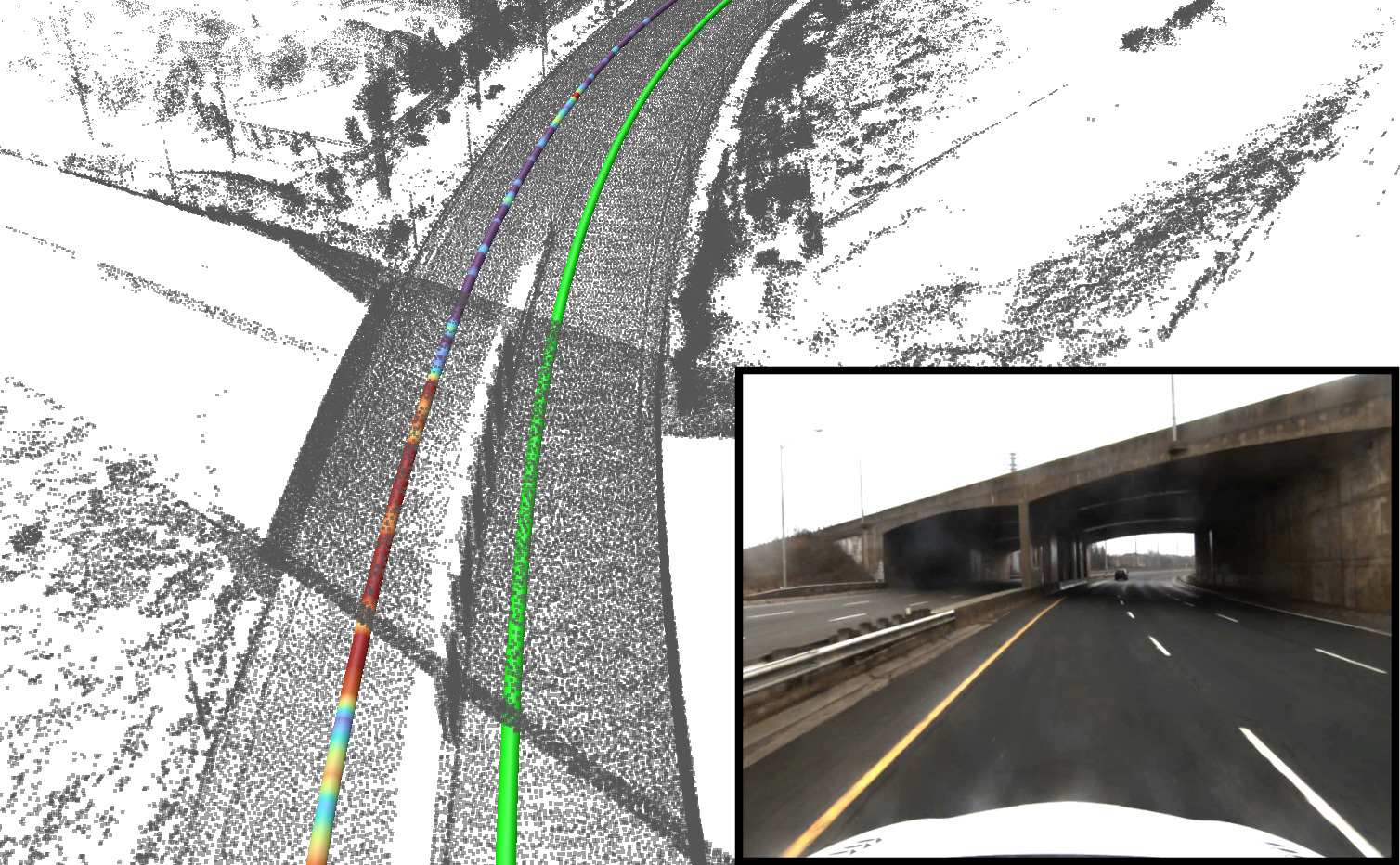}
    \caption{The four \emph{Regional} sequences have been collected in both directions (two each way). When mapping with one sequence, part of the environment is only visible from the opposite traffic lane, thus leaving gaps in the map (as illustrated here under a bridge, with the map in gray, and the mapping trajectory in green). This makes localization challenging: the second line represents the localization trajectory of the second sequence and is colored with the localization error (purple low and red high).}
    \label{fig:highway}
\end{figure}

\emph{Suburbs} sequences follow the same route as in the Boreas dataset.
The road traffic varies between sequences, but a lot of geometric features are present in the surroundings.
The \emph{Regional} data is collected at a higher speed with fewer geometric features to constrain the vehicle's pose.
The \emph{Tunnel} route goes through a kilometre-long tunnel with very few geometric features to leverage.
Finally, the \emph{Skyway} sequences go over a bridge/skyway with only a few lampposts as static features.
A lot of the non-ground lidar points correspond to moving vehicles.
Fig.~\ref{fig:self_data_sample} provides images from the onboard camera as well as the OpenStreetMap overlay of the trajectories.
Both the \emph{Suburbs} and \emph{Skyway} routes form loops, whereas \emph{Regional} and \emph{Tunnel} sequences do not revisit previously driven areas.
The \emph{Regional} data present an extra challenge for localization: the route is driven twice in each direction, thus, when mapping with one sequence and localizing with the other three, two of the localization runs experience quite different viewpoints than the one of the mapping step (cf. Fig.~\ref{fig:highway}).
Similarly, the tunnel sequences are collected while driving in both directions.
However, the tunnel consists of two separate tubes (one for each traffic direction) without co-visibility.
Thus, our \emph{Tunnel} localization experiments are run for each driving direction independently (two times: one sequence for mapping, one sequence for localization), and the quantitative results are the average of both.

\subsubsection{Baselines and metrics:}

The metrics used for odometry and localization are the same as for the Boreas experiments in Section~\ref{sec:exp_boreas}.
Regarding the odometry baselines, we benchmark 2Fast-2Lamaa against MOLA \citep{blanco2025mola}, Fast-LIO2 \citep{xu2022fastlio2}, and LTR with gyroscope \citep{burnett2022arewe}.
Each of these method address motion distortion in a different manner: MOLA assumes strict constant velocity during each lidar scan (optimizing for the velocity at each \ac{icp} step), Fast-LIO2 undistorts the incoming data based on the previous estimate and the open-loop propagation of the \ac{imu} readings, and LTR performs fully continuous-time estimation with \acp{gp}.
Note that the LTR version here is different from the one present on the Boreas leaderboard.
The main difference is the integration of angular velocity readings from the gyroscope as extra factors in the continuous-time optimization.
For evaluating 2Fast-2Lamaa's localization abilities, we use LTR and Fast-LIO-localization \citep{tsoi2020fastliolocalization} as baselines.
The latter is an open-source project that performs localization in maps built with Fast-LIO2.

\edit{Sensor-specific parameters are updated for each baseline.
Our experiments with MOLA use the default algorithmic parameters from the public implementation.
Fast-LIO2 does not expose algorithmic parameters to the user.
For LTR, the parameters have been tuned for the best average performance.
Fast-LIO localization has been tuned to limit the number of critical failures.}

\subsubsection{Odometry benchmark:}

\begin{table*}
    \centering
    \caption{Average relative pose accuracy of the proposed method and several baselines on a subset of the Boreas-RT dataset.}
    \setlength{\tabcolsep}{2pt}
    \begin{tabularx}{\linewidth}{lYYYY}
        \toprule
        \multirow{2}{*}{\textbf{Sequence type} \scriptsize(length, avg. / max. vel.)} & \textbf{MOLA} & \textbf{Fast-LIO2} & \textbf{LTR} w/ gyro & \textbf{2Fast-2Lamaa} 
        \\
         & \scriptsize\citep{blanco2025mola} & \scriptsize\citep{xu2022fastlio2} & \scriptsize\citep{burnett2022arewe} & \scriptsize topometric (ours)
        \\
        \midrule
        \textbf{Suburbs} \scriptsize($4\times7.9\, \si{\km}$, $8.1$ / $18.6\,\si{\m/\s}$) & 0.74 / 0.27 & 0.74 / 0.20 & 0.29 / 0.09 & \textbf{0.22} / \textbf{0.08}
        \\ 
        \textbf{Regional} \scriptsize($4\times9.3\, \si{\km}$, $11.1$ / $27.0\,\si{\m/\s}$) & $\:\:$0.81 / 0.23$^\dagger$ & $\:\:$0.88 / 0.19$^*$ & 0.35 / 0.09 & \textbf{0.20} / \textbf{0.08}
        \\
        \textbf{Tunnel} \scriptsize($4\times 1.9\, \si{\km}$, $8.9$ / $28.2\,\si{\m/\s}$) & $\:\:$1.72 / 0.30 $^\dagger$ & 0.83 / 0.17 & 2.49 / \textbf{0.09} & \textbf{0.34} / 0.11
        \\
        \textbf{Skyway} \scriptsize($4\times 11.1\, \si{\km}$, $18.2$ / $31.7\,\si{\m/\s}$) & $\:\:$X$^\dagger$ & 0.98 / 0.20 & 0.59 / \textbf{0.08} & \textbf{0.30} / 0.09
        \\
        \bottomrule
        \multicolumn{5}{l}{\scriptsize KITTI odometry metric reported as \textit{XX / YY} with \textit{XX} [\%] and \textit{YY} [$\si{\degree}/100\,\si{\m}$] the translation and orientation errors, respectively.}
        \\
        \multicolumn{5}{l}{\scriptsize $^\dagger$ Mola displayed high drift (error $> \SI{10}{\percent}$) on three sequences from Regional, one from Tunnel, and all from Skyway. Average computed by omitting these failed sequences.}
        \\
        \multicolumn{5}{l}{\scriptsize $^*$ Fast-LIO2 failed on one of the Regional sequences (trajectory divergence). Average computed by omitting that sequence.}
    \end{tabularx}
    \label{tab:odom_ours}
\end{table*}

The odometry errors obtained with our method and the baselines are reported in Table~\ref{tab:odom_ours}.
Both MOLA and Fast-LIO2 experienced failures in at least one sequence (large drift above \SI{10}{\percent}, or complete state divergence), with MOLA failing in half of the total sequences.
For fairness, we only display the average across the successful runs.
When both methods succeed in estimating the vehicle's trajectory, they reach a similar level of accuracy.
LTR performs well when the environment is rich in geometric features (\emph{Suburbs} and \emph{Regional}), but its odometry error significantly increases for more challenging routes.
2Fast-2Lamaa is the only method that provides consistent results across all environments without failures.
Its translation accuracy outperforms all the baselines.
The \emph{Suburbs} and \emph{Regional} results of both LTR and 2Fast-2Lamaa highlight the accuracy gain of elegantly modelling the continuous motion of the sensors, thus addressing the problem of motion-distortion correction in a principled way. 
We believe that the increased difference in accuracy for more challenging sequences can be explained by the use of our continuous dense distance fields that better represent the environment.

\subsubsection{Localization benchmark:}

\begin{figure}
    \centering
    \includegraphics[width=0.99\columnwidth]{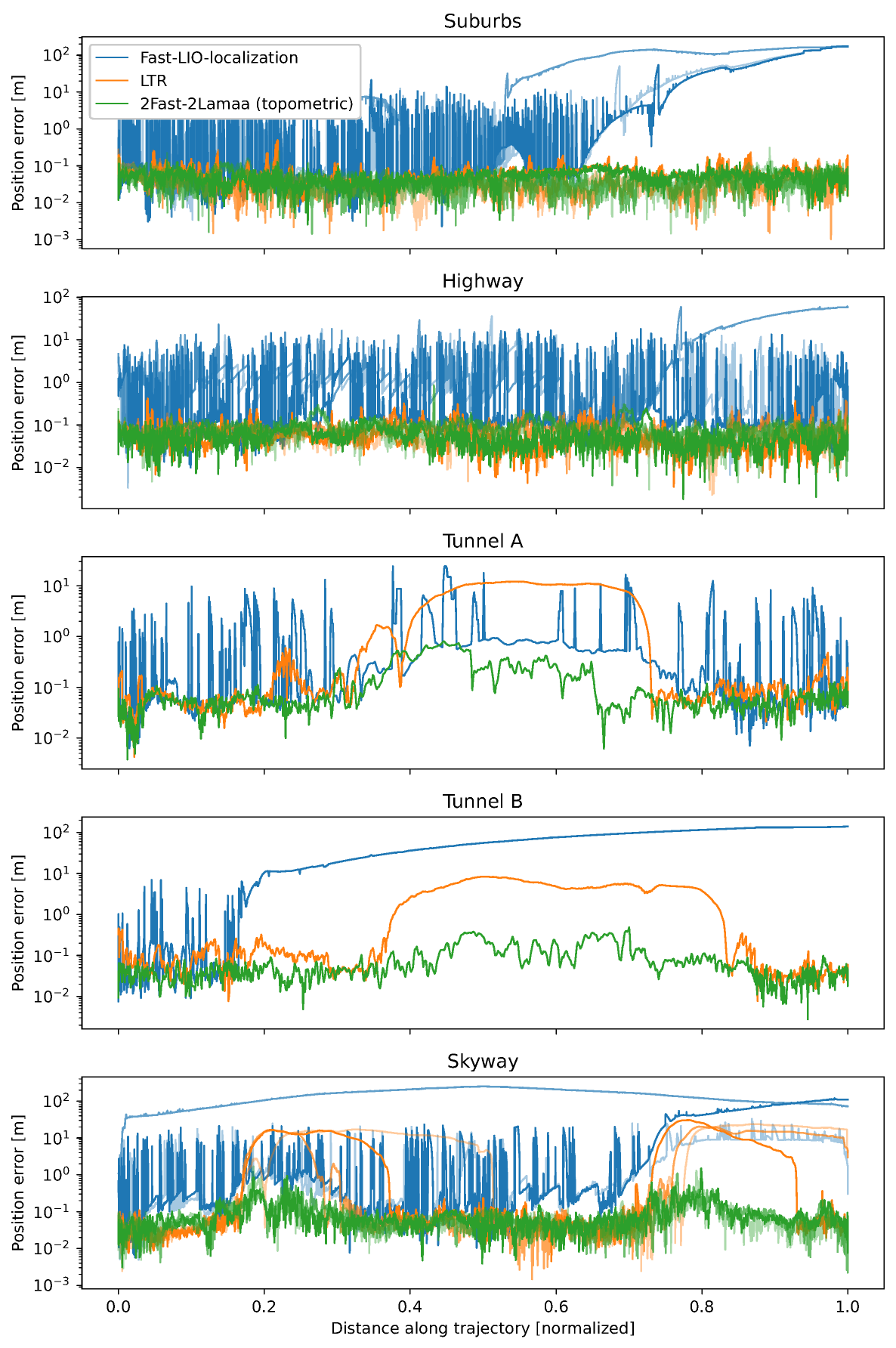}
    \caption{Localization error (position [m] with logarithmic scale) against travelled distance of 2Fast-2Lamaa and the different baselines on the Boreas-RT dataset. The different sequences are shown with different shades of colour.}
    \label{fig:localization_self}
\end{figure}

Table~\ref{tab:localization_self} shows the results of our evaluation.
On the \emph{Suburbs} sequences, 2Fast-2Lamaa and LTR perform similarly with a slight advantage for LTR, as in the Boreas dataset.
However, for all the other routes, 2Fast-2Lamaa is the best performing method with the smallest translational errors.
When considering the more challenging environments (\emph{Tunnel} and \emph{Skyway}), the difference between our method and the baselines is larger.
2Fast-2Lamaa's error stays contained with \acp{rmse} under \SI{0.20}{\m}, while LTR and Fast-LIO-localization display errors of a couple of meters.
Note that overall Fast-LIO-localization performs poorly and fails on many sequences (\ac{rmse} $> \SI{10}{\m}$).
Fig.~\ref{fig:localization_self} shows the localization (position) for all methods and sequences as a function of the distance travelled.
We believe that the Fast-LIO-localization implementation does not fuse the odometry information and the regular localization steps appropriately, and potential dropouts exacerbate the issue due to high computational load.
This is supported by the noisy nature of Fast-LIO-localization's plots, which oscillate between centimetre- and meter-level errors.
Eventually, most sequences end by drifting away from the original trajectory.

We also want to remind the reader that the \emph{Regional} sequences are run in both directions for localization, while the mapping is done only in one, as previously illustrated in Fig.~\ref{fig:highway}.
When considering the sequence where the car is travelling in the same direction as the mapping run, the \acp{rmse} are: Long. = \SI{0.030}{m}, Lat. = \SI{0.025}{\m}, Vert. = \SI{0.029}{\m}, Roll = \SI{0.029}{\degree}, Pitch = \SI{0.022}{\degree}, and Yaw = \SI{0.028}{\degree}.
These values are similar to those obtained in the easiest environment (\emph{Suburbs}), demonstrating robustness to high-velocity trajectories.

\begin{table*}
    \centering
    \caption{RMSE localization error for 2Fast-2Lamaa and different baselines on a subset of the Boreas-RT dataset.}
    \setlength{\tabcolsep}{2pt}
    \begin{tabularx}{\linewidth}{lYYYYYY}
        \toprule
        \textbf{Seq. type \& Method} & \textbf{Long.} \scriptsize[m] & \textbf{Lat.} \scriptsize[m] & \textbf{Vert.} \scriptsize[m] & \textbf{Roll} \scriptsize[$^\circ$] & \textbf{Pitch} \scriptsize[$^\circ$] & \textbf{Yaw} \scriptsize[$^\circ$] 
        \\
        \midrule
        \textbf{Suburbs} &
        \\\quad\textbf{Fast-LIO-localization} \scriptsize\citep{tsoi2020fastliolocalization}\normalsize$^\dagger$  & X & X & X & X & X & X
        \\\quad\textbf{LTR} \scriptsize w/ gyro \citep{burnett2022arewe}                                            & 0.032 & \textbf{0.021} & \textbf{0.032} & \textbf{0.013} & \textbf{0.011} & \textbf{0.021}
        \\\quad\textbf{2Fast-2Lamaa} \scriptsize topometric (ours)                                                   & \textbf{0.024} & \textbf{0.021} & 0.034 & 0.036 & 0.025 & 0.026
        \\
        \midrule
        \textbf{Regional}
        \\\quad\textbf{Fast-LIO-localization} \scriptsize\citep{tsoi2020fastliolocalization}\normalsize$^\dagger$  & 2.031 & 0.510 & 0.603 & 0.310 & 0.660 & 1.183
        \\\quad\textbf{LTR} \scriptsize w/ gyro \citep{burnett2022arewe}                                            & 0.056 & 0.055 & 0.061 & \textbf{0.037} & 0.039 & 0.035
        \\\quad\textbf{2Fast-2Lamaa} \scriptsize topometric (ours)                                                   & \textbf{0.042} & \textbf{0.035} & \textbf{0.029} & 0.038 & \textbf{0.036} & \textbf{0.033}
        \\
        \midrule
        \textbf{Tunnel} &
        \\\quad\textbf{Fast-LIO-localization} \scriptsize\citep{tsoi2020fastliolocalization}\normalsize$^\dagger$  & 2.212 & 0.100 & 0.048 & 0.205 & 0.192 & 0.715
        \\\quad\textbf{LTR} \scriptsize w/ gyro \citep{burnett2022arewe}                                            & 2.960 & 0.111 & 0.044 & \textbf{0.063} & 0.061 & 0.071
        \\\quad\textbf{2Fast-2Lamaa} \scriptsize topometric (ours)                                                   & \textbf{0.178} & \textbf{0.024} & \textbf{0.026} & 0.087 & \textbf{0.020} & \textbf{0.026}
        \\
        \midrule
        \textbf{Skyway} &
        \\\quad\textbf{Fast-LIO-localization} \scriptsize\citep{tsoi2020fastliolocalization}\normalsize$^\dagger$  & 6.105 & 1.734 & 0.888 & 0.482 & 0.538 & 0.968
        \\\quad\textbf{LTR} \scriptsize w/ gyro \citep{burnett2022arewe}\normalsize$^*$                             & 7.502 & 0.203 & 0.085 & 0.054 & 0.042 & 0.175
        \\\quad\textbf{2Fast-2Lamaa} \scriptsize topometric (ours)                                                   & \textbf{0.059} & \textbf{0.026} & \textbf{0.035} & \textbf{0.050} & \textbf{0.030} & \textbf{0.028}
        \\
        \bottomrule
    \multicolumn{7}{l}{\scriptsize $^\dagger$ Fast-LIO-localization succeeded (RMSE $<$ \SI{10}{\m}) only on 2 out of 3 Regional sequences, 1 out of 2 Tunnel, and 1 out of 3 Skyway.}
    \\
    \multicolumn{7}{l}{\scriptsize $^*$ Only 1 out of 3 Skyway sequences ran to completion for LTR.}
    \end{tabularx}
    \label{tab:localization_self}
\end{table*}

\subsection{Newer College dataset}
\label{sec:exp_newer}

\subsubsection{Dataset description:}

To demonstrate the versatility of 2Fast-2Lamaa, we benchmark it using a dataset collected with a handheld device, the Newer College Dataset \citep{ramezani2020newer}.
It consists of 5 sequences collected while walking around the New College at the University of Oxford (UK) with a sensor suite equipped with an Intel RealSense D435i stereo camera and an Ouster OS1-64 lidar.
Both sensors possess an embedded \ac{imu}.
In our experiments, we only use the lidar and its internal \ac{imu}.
The dataset also provides a detailed map of the environment as a coloured point cloud acquired with a Leica BLK360 survey lidar.
The ground-truth trajectory of the sensors is obtained by registering the individual lidar scans to the map with \ac{icp}.

\subsubsection{Baselines and metrics:}

For odometry, we adopt the standard \ac{ate} to benchmark against various lidar-inertial frameworks.
The \ac{ate} is defined as the \ac{rmse} between the positions of the estimated trajectory and the ones of the ground-truth after alignment using the Umeyama algorithm \citep{umeyama1991least}.
Similarly to the evaluation with the Boreas-RT dataset, we use Fast-LIO2 \citep{xu2022fastlio2} and STEAM-LIO \citep{burnett2024continuous} as baselines.
We also include DLIO \citep{chen2023dlio} as it is one of the best-performing methods on the Newer College dataset.

It is important to note that the ground-truth provided in the Newer College dataset is not as accurate as that of the other datasets in our experiments.
In their paper, \cite{ramezani2020newer} showed that when the system is static at the start of the dataset, the ground-truth position jitters in a sphere of around \SI{15}{\cm} in diameter.
Additionally, we empirically found a non-negligible rotational error (up to \SI{2}{\degree}) that hinders the use of the Boreas localization metric, thus preventing the evaluation of localization from one sequence to the other.\footnote{As a relative pose metric, the Boreas localization error is sensitive to ground-truth orientation accuracy. For example, with a localization run \SI{10}{\m} away from the mapping run, a ground-truth error of \SI{1}{\degree} leads to a position error of \SI{17}{\cm} due to the lever arm effect.}
Accordingly, we only perform localization against the provided map with 2Fast-2Lamaa as a demonstration of its ability to perform localization within an existing \edit{global} map.
We compute the localization \ac{rmse} directly between the provided ground-truth positions and the estimated ones as a sanity check.

\subsubsection{Odometry benchmark:}

Table~\ref{tab:newer_college} shows the \ac{ate} obtained for each of the sequences.
Note that the baseline results are the ones reported by \cite{burnett2024continuous} for STEAM-LIO, and by \cite{chen2023dlio} for Fast-LIO2 and DLIO.
Overall, all the methods provide a similar level of accuracy, with a slight advantage for 2Fast-2Lamaa for three out of the five sequences.
This benchmark demonstrates that our method competes with state-of-the-art frameworks even when considering handheld sensor suites.
It also shows that our method can be used with a low-cost \ac{imu}, such as the one embedded in the Ouster lidar. 

\begin{table*}
    \centering
    \caption{ATE and RMSE localization error [\SI{}{\m}] for 2Fast-2Lamaa and various baselines on the Newer College dataset sequences.}
    \setlength{\tabcolsep}{2pt}
    \begin{tabularx}{\linewidth}{lYYYYY}
        \toprule
        \textbf{Metric \& Method} & \shortstack[c]{\textbf{01\_short}\\\scriptsize(\SI{1609}{\m}, \SI{1530}{\s})} & \shortstack[c]{\textbf{02\_long}\\\scriptsize(\SI{3063}{\m}, \SI{2180}{\s}) }& \shortstack[c]{\textbf{05\_quad}\\\scriptsize(\SI{479}{\m}, \SI{398}{\s}) } & \shortstack[c]{\textbf{06\_spinning}\\\scriptsize(\SI{97}{\m}, \SI{120}{\s})} & \shortstack[c]{\textbf{07\_parkland}\\\scriptsize(\SI{696}{\m}, \SI{500}{\s})}
        \\
        \midrule
        \textbf{ATE}
        \\
        \quad\textbf{Fast-LIO2} \scriptsize\citep{xu2022fastlio2} & 0.378 & 0.332 & 0.088 & 0.078 & 0.148
        \\
        \quad\textbf{DLIO} \scriptsize\citep{chen2023dlio} & 0.360 & 0.327 & 0.084 & \textbf{0.061} & 0.120
        \\
        \quad\textbf{STEAM-LIO} \scriptsize\citep{burnett2024continuous} & 0.304 & 0.337 & 0.109 & 0.082 & 0.144
        \\
        \quad\textbf{2Fast-2Lamaa} \scriptsize global map (ours) & \textbf{0.262} & \textbf{0.266} & \textbf{0.078} & 0.085 & \textbf{0.115}
        \\
        \midrule
        \textbf{Localization RMSE}
        \\
        \quad\textbf{2Fast-2Lamaa} \scriptsize global map (ours) & 0.128 & 0.154 & 0.089 & 0.091 & 0.155
        \\
        \bottomrule
    \end{tabularx}
    \label{tab:newer_college}
\end{table*}

\subsubsection{Localization benchmark:}

The localization \ac{rmse} of 2Fast-2Lamaa is reported in the last line of Table~\ref{tab:newer_college}.
As mentioned previously, the ground-truth accuracy level does not allow for a thorough analysis of the proposed pipeline's localization performance.
Accordingly, it is difficult to draw any conclusion from these figures. 
However, these results demonstrate the soundness of the proposed approach and its ability to perform localization in a map generated with another sensor and mapping algorithm.

\subsection{VBR-SLAM dataset}
\label{sec:exp_vbr}

\subsubsection{Dataset description:}
To further demonstrate the versatility of 2Fast-2Lamaa and the proposed \edit{online} loop closure detection and correction mechanism, we benchmark our method on the VBR-SLAM dataset \citep{brizi2024vbr}.
It consists of 16 sequences collected with a lidar, stereo vision, and an RTK-GPS-IMU solution in 6 different environments.
Half of the sequences use an automotive platform with an Ouster OS0-64 lidar at \SI{20}{\hertz} (environments \emph{Campus} and \emph{Ciampino}).
The other half is collected using a handheld sensor suite comprising an Ouster OS1-128 lidar, which collects scans at \SI{10}{Hz} (environments \emph{Colosseo}, \emph{Pincio}, \emph{Spagna}, and \emph{DIAG}).
In both scenarios, 2Fast-2Lamaa uses the lidar's embedded \ac{imu}.
Out of the 16 sequences, only 8 have publicly available ground-truth.
The ground-truth of the other sequences is held out for use in the dataset's public leaderboard.
In this section, we only leverage sequences with provided ground-truth.\footnote{To avoid potential multiple submissions to the public leaderboard, due to software changes/improvements during the review process, we decided to submit our results to the leaderboard only the final version of the current paper.}

\begin{table*}
    \centering
    \caption{ATE obtained using the VBR-SLAM dataset.} 
    \setlength{\tabcolsep}{3pt}
    \begin{tabularx}{\linewidth}{lYYYYYYYYY}
        \toprule
        \textbf{Method} &  \textbf{Colosseo} [m] & \textbf{Campus 0} [m] & \textbf{Campus 1} [m] & \textbf{Pincio} [m] & \textbf{Spagna} [m] & \textbf{DIAG} [m] & \textbf{Ciamp. 0} [m] & \textbf{Ciamp. 1} [m] & \textbf{Overall score}
        \\
        \midrule
        & \multicolumn{8}{c}{Training sequences}
        \\
        \midrule
        \textbf{PIN-SLAM} \scriptsize\citep{pan2024pinslam} & \textbf{0.506} & \textbf{0.555} & \textbf{0.392} & 0.647 & 0.480 & 0.362 & 1.366 & 1.253 & 74.44
        \\
        \textbf{SMLE} \scriptsize\citep{bhandari2024minimal}& 1.411 & 1.196 & 0.994 & 0.793 & 0.696 & 0.457 & 1.495 & 1.308 & 71.65
        \\
        \textbf{KISS-ICP} \scriptsize\citep{vizzo2023kiss} & 1.517 & 1.026 & 1.039 & 0.785 & 1.048 & 1.397 & 2.808 & 2.153 & 68.23
        \\
        \textbf{2Fast-2Lamaa} \scriptsize pose-graph (ours)& 0.634 & 0.900 & 0.584 & \textbf{0.624} & \textbf{0.442} & \textbf{0.359} & \textbf{1.141} & \textbf{0.831} & \textbf{74.49}
        \\
        \midrule
        & \multicolumn{8}{c}{Testing sequences}
        \\
        \midrule
        \textbf{PIN-SLAM} \scriptsize\citep{pan2024pinslam} & 0.422 & 0.570 & \textbf{0.864} & 1.063 & 0.373 & \textbf{0.394} & 1.790 & 1.521 & 73.00
        \\
        \textbf{SMLE} \scriptsize\citep{bhandari2024minimal}& 0.454 & 0.897 & 1.022 & 1.071 & \textbf{0.247} & 0.456 & 1.350 & \textbf{1.396} & 73.11
        \\
        \textbf{KISS-ICP} \scriptsize\citep{vizzo2023kiss} & 0.525 & 1.019 & 1.289 & 0.864 & 0.271 & 0.463 & 1.700 & 2.566 & 71.30
        \\
        \textbf{2Fast-2Lamaa} \scriptsize pose-graph (ours)& \textbf{0.402} & \textbf{0.484} & 0.874 & \textbf{0.756} & 0.350 & 0.405 & \textbf{1.154} & 1.558 & \textbf{74.02}
        \\
        \bottomrule
    \end{tabularx}
    \label{tab:vbr}
\end{table*}

\subsubsection{Baselines and metrics:}
As the trajectories form large loops, we use 2Fast-2Lamaa with the \edit{online} loop-closure detection and correction mechanism from Section~\ref{sec:loop}.
We use PIN-SLAM \citep{pan2024pinslam}, SMLE \citep{bhandari2024minimal}, and KISS-ICP \citep{vizzo2023kiss} as lidar-based baselines.
PIN-SLAM is especially designed for state estimation with loop closures.
\edit{As done in the VBR-SLAM public benchmark, we report both the \ac{rmse} \ac{ate}, and the \ac{rpe} as computed with the VBR-provided script.
The \ac{rpe} corresponds to the KITTI odometry metric, but computed with trajectory chunks of different sizes proportional to the whole trajectory length.}
The VBR-SLAM dataset also provides an overall score that aggregates the results throughout all the sequences to rank the different methods (the higher the better).

\subsubsection{SLAM benchmark:}

Table~\ref{tab:vbr} shows the results obtained with 2Fast-2Lamaa and the different baselines \edit{on both the training and testing sequences}.
While the ranking of the methods varies from one sequence to another, on average, our method displays the best performance, as shown by the overall score.
\edit{Note that the second-best method, PIN-SLAM, requires a high-grade GPU to perform close to real-time operations.}
As discussed later in Section~\ref{sec:exp_computation}, 2Fast-2Lamaa only uses a fraction of a consumer-grade laptop CPU to achieve better performance.
Our future work includes the formulation of a full-batch optimization built on point-to-surface residuals, with the proposed distance field, for better global map accuracy.

\edit{
\subsubsection{Odometry benchmark:}

For completeness, we also provide the \ac{rpe} metrics as reported on the VBR-SLAM leaderboard in Table~\ref{tab:vbr_odom}.
SMLE displays the best performance on this benchmark.
2Fast-2Lamaa's results are respectively second and last on the training and testing sequences.
It is important to note that the ground-truth of the VBR-SLAM dataset is obtained as a full-batch optimization of the lidar data using the RTK-GNSS measurements as priors.
This approach does not provide an independent ground-truth and highly depends on the registration algorithm used.
While sufficient for global trajectory consistency analysis (via the \ac{ate}), we believe it creates biases in the \ac{rpe} evaluation.
This latter metric is highly sensitive to the rotational accuracy of the ground-truth due to the lever effect of orientation inaccuracies when using long trajectory chunks as done in this leaderboard.
For example, a rotation error of \SI{0.1}{\degree} corresponds to a translation error of around \SI{0.17}{\m} after \SI{100}{\m} travelled, which represents a minimum relative error of \SI{0.17}{\percent}.
To the best of our knowledge, at the time of writing, the VBR-SLAM ground-truth is computed without accounting for motion distortion.
Accordingly, the rotational accuracy of the ground-truth is probably suboptimal, especially on handheld sequences where the sensors' orientation can change a lot during each lidar scan ($\gg \SI{0.1}{\degree}$).
This correlates with the fact that the best performing method, SMLE, does not perform motion distortion correction, and that 2Fast-2Lamaa's performance is significantly lower on the handheld sequences as opposed to the automotive ones.

\begin{table*}
    \centering
    \caption{RPE obtained using the VBR-SLAM dataset.} 
    \setlength{\tabcolsep}{3pt}
    \begin{tabularx}{\linewidth}{lYYYYYYYYY}
        \toprule
        \textbf{Method} &  \textbf{Colosseo} [\%] & \textbf{Campus 0} [\%] & \textbf{Campus 1} [\%] & \textbf{Pincio} [\%] & \textbf{Spagna} [\%] & \textbf{DIAG} [\%] & \textbf{Ciamp. 0} [\%] & \textbf{Ciamp. 1} [\%] & \textbf{Overall score}
        \\
        \midrule
        & \multicolumn{8}{c}{Training sequences}
        \\
        \midrule
        \textbf{PIN-SLAM} \scriptsize\citep{pan2024pinslam} & 0.490 & \textbf{0.243} & \textbf{0.155} & \textbf{0.454} & 11.61 & 0.468 & 0.738 & 0.648 & 66.80
        \\
        \textbf{SMLE} \scriptsize\citep{bhandari2024minimal}& \textbf{0.431} & 0.613 & 0.312 & 0.461 & \textbf{0.310} & \textbf{0.457} & \textbf{0.424} & \textbf{0.293} & \textbf{76.70}
        \\
        \textbf{KISS-ICP} \scriptsize\citep{vizzo2023kiss} & 0.453 & 0.515 & 0.300 & 0.486 & 0.399 & 1.791 & 0.650 & 0.395 & 75.01
        \\
        \textbf{2Fast-2Lamaa} \scriptsize pose-graph (ours)& 0.905 & 0.599 & 0.390 & 0.821 & 0.581 & 0.718 & 0.525 & 0.372 & 75.09
        \\
        \midrule
        & \multicolumn{8}{c}{Testing sequences}
        \\
        \midrule
        \textbf{PIN-SLAM} \scriptsize\citep{pan2024pinslam} & \textbf{0.496} & \textbf{0.228} & \textbf{0.526} & 1.082 & 0.456 & 0.550 & 1.018 & 0.769 & 74.88
        \\
        \textbf{SMLE} \scriptsize\citep{bhandari2024minimal}& \textbf{0.393} & 0.251 & 0.639 & 0.557 & \textbf{0.323} & \textbf{0.515} & \textbf{0.612} & \textbf{0.282} & \textbf{76.43}
        \\
        \textbf{KISS-ICP} \scriptsize\citep{vizzo2023kiss} & 0.453 & 0.250 & 0.655 & \textbf{0.488} & 0.410 & 0.776 & 0.761 & 0.475 & 75.73
        \\
        \textbf{2Fast-2Lamaa} \scriptsize pose-graph (ours)& 0.969 & 0.356 & 0.703 & 0.932 & 0.701 & 1.025 & 0.653 & 0.629 & 74.03
        \\
        \bottomrule
    \end{tabularx}
    \label{tab:vbr_odom}
\end{table*}

}

\subsection{Ablation study}
\label{sec:exp_ablation}

\begin{table}
    \centering
    \caption{Ablation study of 2Fast-2Lamaa based on the Boreas-RT dataset and using the average translational KITTI odometry metric [\%].}
    \setlength{\tabcolsep}{2pt}
    \begin{tabularx}{\linewidth}{lYYYY}
        \toprule
        \textbf{Variant} &  \textbf{Suburbs} & \textbf{Regional} & \textbf{Tunnel} & \textbf{Skyway}
        \\
        \midrule
        \textbf{Baseline} & \textbf{0.22} & \textbf{0.20} & 0.34 & 0.30
        \\
        \textbf{No edge} & \textbf{0.22} & \textbf{0.20} & 0.46 & \textbf{0.26}
        \\
        \textbf{No field} & 0.29 & 0.24 & 0.35 & 0.31
        \\
        \textbf{Global map} & 0.27 & \textbf{0.20} & 0.32 & 0.32
        \\
        \textbf{No map} & 0.55 & 0.66 & 0.56 & 0.85
        \\
        \textbf{No carving} & 0.31 & \textbf{0.20} & 0.33 & 0.31$\dagger$
        \\
        \textbf{No acc.} & 0.23 & 0.21 & 0.33 & 0.30
        \\
        \textbf{No IMU} & 0.25 & 0.25 & 0.36 & 0.33
        \\
        \textbf{No keyframe} & 0.25 & 0.26 & \textbf{0.31} & 0.36
        \\
        \bottomrule
        \multicolumn{5}{l}{$\dagger$ 2Fast-2Lamaa failed on one sequence.}
        \\
        \multicolumn{5}{l}{Average computed with successful runs.}
    \end{tabularx}
    \label{tab:ablation_odom}
\end{table}

\begin{table}
    \centering
    \caption{Ablation study of 2Fast-2Lamaa based on the Boreas-RT dataset and using the RMSE position error [m].}
    \setlength{\tabcolsep}{2pt}
    \begin{tabularx}{\linewidth}{lYYY}
        \toprule
        \textbf{Seq. type \& Variant} & \textbf{Long.} & \textbf{Lat.} & \textbf{Vert.}
        \\
        \midrule
        \textbf{Suburbs} &
        \\
        \quad\textbf{Baseline} & \textbf{0.023} & \textbf{0.021} & \textbf{0.034}
        \\
        \quad\textbf{Global map} & 0.024 & 0.022 & \textbf{0.034}
        \\
        \quad\textbf{Online-carved map} & 0.024 & 0.022 & \textbf{0.034}
        \\
        \quad\textbf{Uncarved map} & 0.024 & 0.022 & \textbf{0.034}
        \\
        \midrule
        \textbf{Regional} &
        \\
        \quad\textbf{Baseline} & \textbf{0.042} & 0.035 & 0.029
        \\
        \quad\textbf{Global map} & 0.044 & \textbf{0.033} & \textbf{0.028}
        \\
        \quad\textbf{Online-carved map} & \textbf{0.042} & 0.034 & 0.030
        \\
        \quad\textbf{Uncarved map} & \textbf{0.042} & \textbf{0.033} & 0.029
        \\
        \midrule
        \textbf{Tunnel} &
        \\
        \quad\textbf{Baseline} & 0.178 & \textbf{0.024} & 0.026
        \\
        \quad\textbf{Global map} & 0.216 & 0.025 & \textbf{0.025}
        \\
        \quad\textbf{Online-carved map} & 0.174 & 0.025 & 0.026
        \\
        \quad\textbf{Uncarved map} & \textbf{0.147} & 0.025 & \textbf{0.025}
        \\
        \midrule
        \textbf{Skyway} &
        \\
        \quad\textbf{Baseline} & 0.059 & \textbf{0.026} & \textbf{0.035}
        \\
        \quad\textbf{Global map} & 0.061 &  0.034 & 0.060
        \\
        \quad\textbf{Online-carved map} & 0.057 & \textbf{0.026} & 0.036
        \\
        \quad\textbf{Uncarved map} & \textbf{0.052} & \textbf{0.026} & 0.036
        \\
        \bottomrule
    \end{tabularx}
    \label{tab:ablation_localization}
\end{table}

2Fast-2Lamaa is a fairly modular framework.
Thus, numerous combinations of components/features can be used (global map vs. submaps, using edge features or not, cleaning the map or not, ...) to tailor the performance to a specific environment or application.
In this section, we perform an ablation study using \edit{our 16-sequence subset of Boreas-RT} (presented in Section~\ref{sec:exp_self}).
We refer to the full framework used to generate 2Fast-2Lamaa's results on Boreas and Boreas-RT datasets (Sections~\ref{sec:exp_boreas} and~\ref{sec:exp_self}, respectively) as the `baseline'.
Each variant corresponds to a single difference from the baseline.
All the results are presented in Tables~\ref{tab:ablation_odom} and~\ref{tab:ablation_localization} for the odometry and localization modes, respectively, and are discussed by theme in the following subsections.

\subsubsection{Features:}

2Fast-2Lamaa's undistortion modules extract planar and edge features from the raw data, and these are used for point-to-plane and point-to-line distance residuals in the optimization.
As a reminder, the first type is obtained by simply subsampling the incoming data, while the second corresponds to jumps in range.
As per the discrete nature of lidar data collection, limited by a certain angular resolution, the likelihood of acquiring a point exactly on the edge of an object in the real world is null.
On the other hand, a point that belongs to a plane in the real world is not affected by the sensor resolution: multiple consecutive points are likely to belong to the same planar patch (given that the observed plane is large enough).
Thus, when considering the reality of lidar data collection and the downstream distance-based residuals, edge features create noisier constraints in the optimization process.
However, they provide more constraints due to a single \ac{dof}, where planar residuals have two.
In environments that are rich with structural elements, we expect the use of both feature types to perform similarly or worse than using solely planar points.
Inversely, edge features provide crucial complementary information in geometrically challenging environments such as tunnels, where planes alone do not constrain the longitudinal motion.
This reasoning is empirically verified in Table~\ref{tab:ablation_odom} with the \emph{No edge} variant that performs better than the baseline for \emph{Skyway} and significantly worse on \emph{Tunnel} sequences.
Note that even without the edge features on these sequences, 2Fast-2Lamaa still outperforms the methods benchmarked in Table~\ref{tab:odom_ours}.

\subsubsection{Continuous distance field:}

In this paragraph, we discuss the contribution of the continuous distance field map towards the global performance of the framework.
2Fast-2Lamaa allows for the direct use of point-to-point distances with the centroids of $\voxelmap$ in the scan-to-map registration step, instead of the \ac{gp}-based distance field.
The voxel size is the same as for the baseline underlying voxelized data structure: \SI{0.3}{\m} in all our automotive experiments.
The corresponding \emph{No field} variant in Table~\ref{tab:ablation_odom} shows a drop in performance with the relative translation error increasing by 20 to 30 percent in the more structured environments.
While the difference is limited, using the proposed distance field has a non-negligible positive impact on the overall performance.
Note that the planar motion in the \emph{Skyway} sequences is mostly constrained by lampposts along the road.
We believe that the fairly dense voxel-based data structure provides enough points along the posts, making the point-to-point residuals sufficient in effectively constraining the motion.
Section~\ref{sec:exp_computation} provides information on the difference in computation time between the baseline and the \emph{No field} variant.

\subsubsection{Global map vs. submaps:}

\begin{figure}
    \centering
    \includegraphics[width=0.99\columnwidth, clip, trim=0cm 0.2cm 0cm 0.68cm]{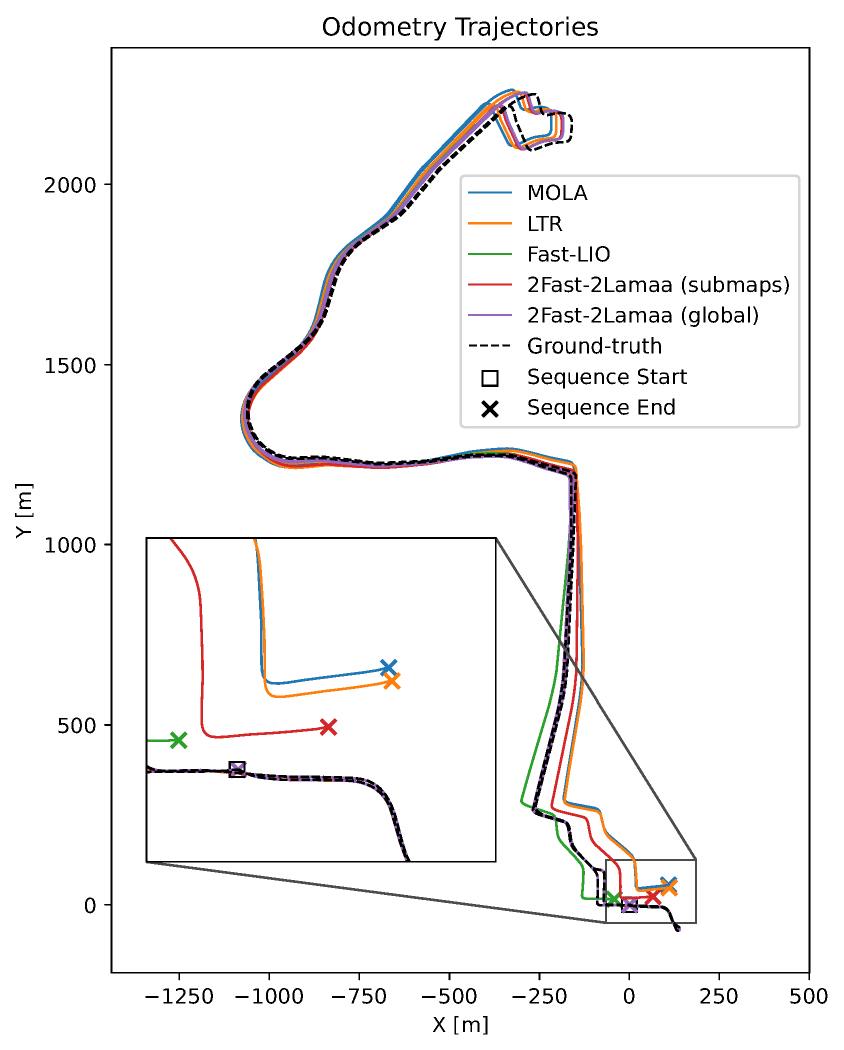}
    \caption{Estimated trajectories with 2Fast-2Lamaa and the different baselines on a \emph{Suburbs} sequence. 2Fast-2Lamaa, with an incrementally built global map, is the only method that produces a trajectory that ends at the same location as the ground-truth. Note that none of the frameworks perform explicit loop-closure correction.}
    \label{fig:submap_vs_global}
\end{figure}

As discussed in Appendix~\ref{app:phtree}, the use of \edit{i-Octree} enables fast distance queries in incrementally built large-scale maps.
If the system's trajectory does not form large loops that accumulate too much drift, using a single map (as opposed to a succession of submaps) enables the estimation of globally consistent trajectories and maps.
If the system's trajectory does not form large loops, \edit{the drift of the estimated state when revisiting previously mapped locations might be small enough to enable the use of a single global map instead of a succession of submaps.}
Fig~\ref{fig:submap_vs_global} illustrates the trajectory obtained when using an incrementally built global map on a \emph{Suburbs} sequence.
Note that 2Fast-2Lamaa is the only method that can provide a consistent trajectory (and map) as the other frameworks end up `forgetting' previously traversed areas.
Looking at the odometry metrics for \emph{Global map} in Table~\ref{tab:ablation_odom}, `single-way' routes (\emph{Regional} and \emph{Tunnel}) are not affected by the change from submap to global map for scan-to-map registration, as expected.
Interestingly, the routes that loop before heading back to the starting position are impacted negatively.
This is due to the high sensitivity of the KITTI odometry metric to the orientation estimates.
When the trajectory revisits a previously mapped area, \edit{due to some drift}, the current state estimate can deviate from the locally consistent odometry to align the current data to the `old part' of the map.
\edit{This creates small `jumps' in the trajectory estimate.}
As the KITTI odometry errors are computed after aligning a single pose per trajectory chunk, this deviation, especially in orientation due to a significant lever arm effect for trajectory chunks from \SI{100}{} to \SI{800}{\m}, locally degrades the metrics.

This last phenomenon does not really impact the localization performance within the global map \edit{in well-structured environments}.
As shown in Table~\ref{tab:ablation_localization}, the results do not significantly change between the baseline and the \emph{Global map} variant except for the \emph{Tunnel} and \emph{Skyway} sequences.
We believe that the inherent drift of 2Fast-2Lamaa when building the global map hinders the sharpness of the few features present in the \emph{Tunnel} and sequences (mostly corresponding to sparse service doors).
Having `blurred' features in the map prevents high accuracy along the longitudinal axis.
\edit{For the \emph{Skyway} sequences, the vertical axis is significantly impacted, while the lateral error is slightly higher.
This is due to the drift of the mapping trajectory creating `double grounds and walls' in areas of the map.}
Note that the estimated pose errors stay similar for the other axes.
While not really demonstrated here, an advantage of building globally consistent maps for later localization is the simplicity of the localization process.
For 2Fast-2Lamaa, it is a simple scan-to-map registration.
There is no need for a complex module to perform topometric graph navigation.

\subsubsection{Map cleaning:}

Another component of our ablation study concerns the removal of dynamic objects from the submaps when performing odometry and localization.
For odometry, the baseline includes an online free-space carving mechanism.
The \emph{No carving} variant does not possess this feature.
The results in Table~\ref{tab:ablation_odom} show no \edit{difference on \emph{Regional} and \emph{Tunnel}, but the \emph{Suburbs} and \emph{Skyway} are negatively impacted, respectively with a higher error and a failure case.}

\begin{figure*}
    \centering
    \def\legenddist{0.1cm}
    \def\vdist{0.6cm}
    \def\imgwidth{0.66\columnwidth}
    \def\hdist{0.02\columnwidth}
    \def\legendtextsize{\small}
    \begin{tikzpicture}
        \tikzstyle{legend} = [align = center, inner sep=0, outer sep=0, node distance = 0em, execute at begin node=\setlength{\baselineskip}{8pt}\small] 
        \tikzstyle{image} = [align = center, inner sep=0, outer sep=0, node distance = 0em] 

        \node[image] (topori) {\includegraphics[clip, width=\imgwidth]{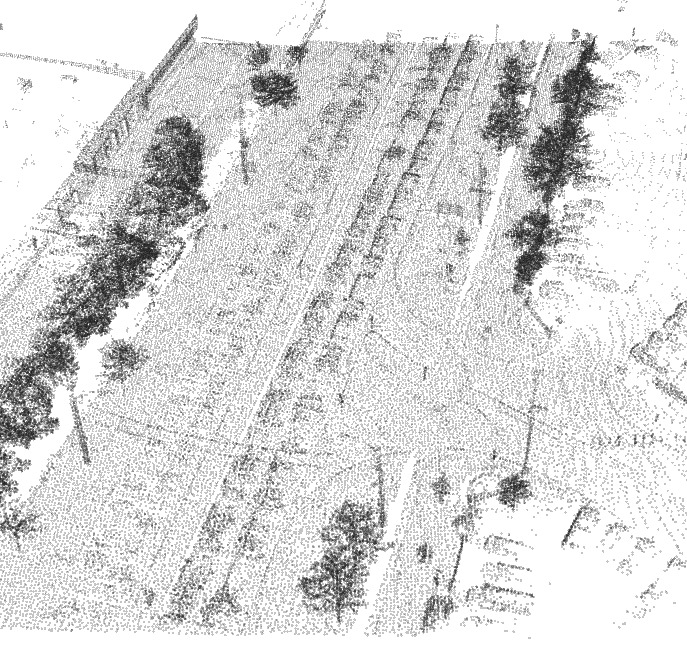}};
        \node[image, below=\vdist of topori] (sidori) {\includegraphics[clip, width=\imgwidth]{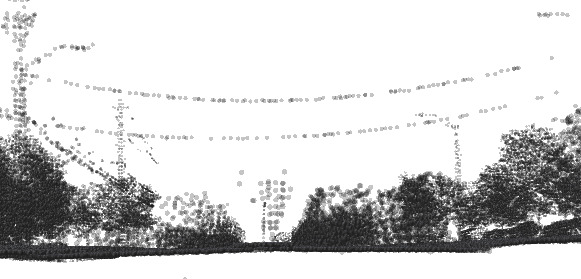}};
        
        \node[image, right=\hdist of topori] (toponline) {\includegraphics[clip, width=\imgwidth]{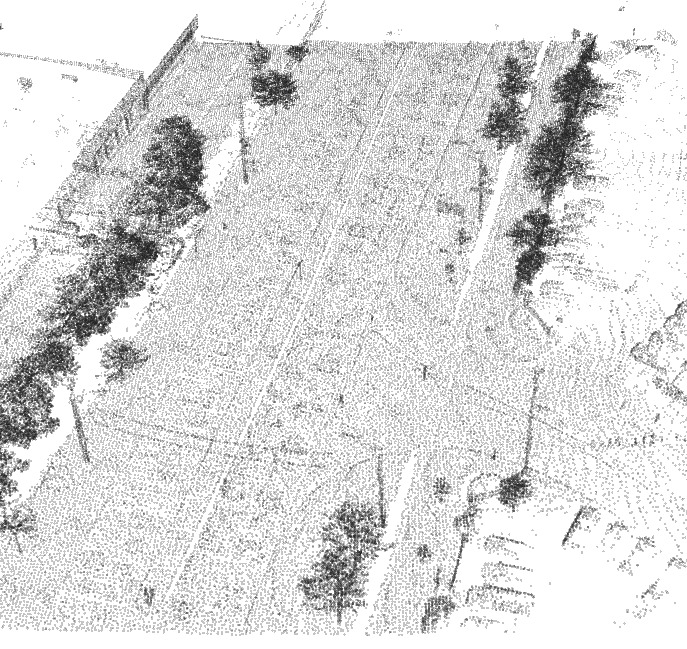}};
        \node[image, below=\vdist of toponline] (sidonline) {\includegraphics[clip, width=\imgwidth]{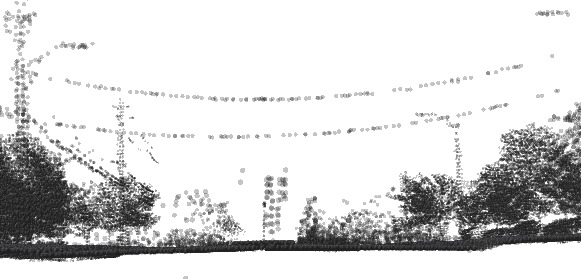}};
        
        \node[image, right=\hdist of toponline] (topoffline) {\includegraphics[clip, width=\imgwidth]{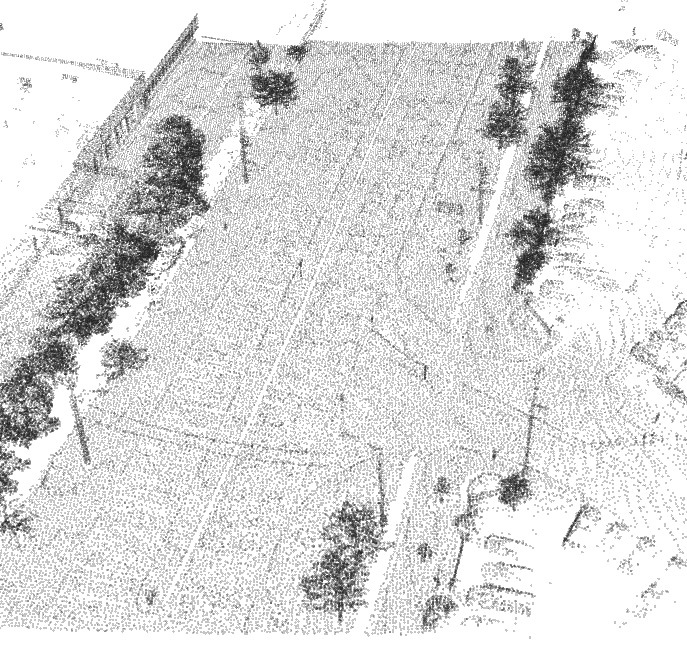}};
        \node[image, below=\vdist of topoffline] (sidoffline) {\includegraphics[clip, width=\imgwidth]{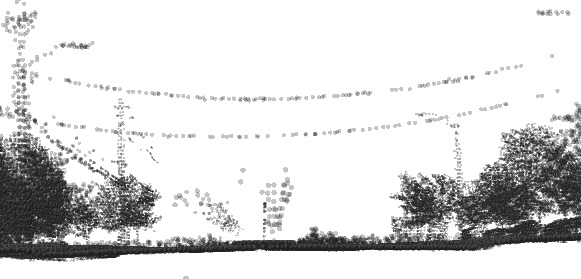}};
        
        \node[legend, below=\legenddist of topori] {\legendtextsize (a) Uncarved topview};
        \node[legend, below=\legenddist of sidori] {\legendtextsize (d) Uncarved sideview};
        
        \node[legend, below=\legenddist of toponline] {\legendtextsize (b) Online-carved topview};
        \node[legend, below=\legenddist of sidonline] {\legendtextsize (e) Online-carved sideview};
        
        \node[legend, below=\legenddist of topoffline] {\legendtextsize (c) Offline-carved topview};
        \node[legend, below=\legenddist of sidoffline] {\legendtextsize (f) Offline-carved sideview};
        
    \end{tikzpicture}
    \caption{Illustration of the dynamic object removal results in a map slice from a \emph{Suburbs} sequence. Most of the dynamic objects are removed by the online free space carving process.}
    \label{fig:exp_carving}
\end{figure*}

As mentioned in the methodology part of this paper, the free-space carving can also be performed offline, given the mapping trajectory estimate, the map(s), and the undistorted scans.
The localization baseline leverages both online and offline free-space carving during the mapping stage.
The \emph{Online-carved map} variant refers to the use solely of the online carving, and the \emph{No carving} one does not perform any dynamic object removal.
Fig.~\ref{fig:exp_carving} illustrates the difference between the uncarved, online-carved, and offline-carved maps.
\edit{Overall, the presence of dynamic points in the map does not seem to have a significant impact on localization on the Boreas-RT dataset.
A deeper analysis of the impact of dynamic points for localization with more datasets and sensors is required before drawing any final conclusion.}

\edit{
\subsubsection{IMU vs constant velocity}

2Fast-2Lamaa's motion distortion correction is based on continuous preintegration to characterize the trajectory of the system.
This continuous state can be replaced with a strict motion model during the interval $[\frametime{i\minus1}, \frametime{i\plus1}]$, similarly to MOLA.
In this setup, we test two combinations of motion model: gyroscope preintegration with constant linear velocity (noted \emph{No acc.}), and both linear and angular constant velocities (noted \emph{No IMU}).
The results in Table~\ref{tab:ablation_odom} show a drop in performance of both variants when compared to the baseline while still staying competitive.
As expected, the performance of \emph{No acc.} is better than that of \emph{No IMU}.
It is important to note that this analysis is based on automotive data.
Thus, the movement of the platform is close to a constant velocity model.
We expect a larger difference in performance with hand-held or drone-mounted sensors.

\subsubsection{Keyframing}

To provide a deeper insight into our framework's parameters, we evaluate the 2Fast-2Lamaa's performance without keyframing and a lower number of points (2000) for scan-to-map registration, matching the parameters used on the VBR and Newer College datasets.
This variant is denoted \emph{No keyframe} in Table~\ref{tab:ablation_odom}.
Our method still outperforms state-of-the-art baselines from Table~\ref{tab:odom_ours} on average.
The total computational load is moderately impacted by this change of parameters, with \SI{9.60}{\percent} total CPU load for the \emph{Baseline} and \SI{15.4}{\percent} for \emph{No keyframe}.
}

\subsection{Computational requirements}
\label{sec:exp_computation}

All the variants of 2Fast-2Lamaa presented in this paper run in real time on a laptop equipped with an Intel i7-1370p CPU and 32GB of RAM.
Table~\ref{tab:computation} shows the CPU load and final RAM usage for a selection of variants from our ablation study.
\edit{All the variants are lightweight and do not use more than \SI{16}{\percent} of the CPU.
Whether it is for odometry or localization, the main difference between using a global map or submaps resides in the larger RAM usage when using a global map.
The similar CPU load in both scenarios highlights the efficiency of the i-Octree used for spatial indexing.
Not using \acp{gp} or free-space carving results in a non-negligible drop in computational load.
An even lighter version could be run by using neither of the two features.
The \emph{No keyframe} variant leads to the highest computational load.
However, this is still moderate and leaves a lot of free CPU resources for other components in a standard robot navigation stack.
In terms of computation time for the \emph{Baseline}, on a typical \emph{Suburbs} sequence, the undistortion/odometry optimization takes \SI{41}{\ms} on average, the scan-to-maps registration (for each keyframe) requires \SI{61}{\ms}, and the subsequent map update lasts \SI{54}{\ms} with free-space carving and \SI{19}{\ms} without.
Note that these numbers can be changed at will by changing the number of threads allocated to each one of the different modules.
}
Note that the baselines (Fast-LIO2, MOLA, and LTR) all required a more powerful computer to run in real time (Intel i9-12900K).

\begin{table}
    \centering
    \caption{Computational load of various configurations of 2Fast-2Lamaa. The ``core usage'' corresponds to the average number of CPU cores used by the specific component.}
    \setlength{\tabcolsep}{2pt}
    \begin{tabularx}{\linewidth}{lYYYYY}
        \toprule
        \textbf{Variant} & \textbf{Undist. core usage} & \textbf{Loc. core usage} & \textbf{Total CPU [\%]} & \textbf{Undist. RAM [GB]} & \textbf{Loc. RAM [GB]}
        \\
        \midrule
        \textbf{Odometry}
        \\
        \quad\textbf{Baseline} & 1.19 & 0.73 & 9.60 & \textbf{0.17} & 0.33
        \\
        \quad\textbf{Global map} & 1.14 & 0.88 & 10.4 & \textbf{0.17} & 1.57
        \\
        \quad\textbf{No field} & 1.09 & \textbf{0.36} & \textbf{7.24} & \textbf{0.17} & \textbf{0.31}
        \\
        \quad\textbf{No carving} & \textbf{1.04} & 0.59 & 8.12 & 0.19 & 0.33
        \\
        \quad\textbf{No keyframe} & 1.42 & 1.66 & 15.4 & 0.18 & 0.50
        \\
        \midrule
        \textbf{Localization}
        \\
        \quad\textbf{Baseline} & 1.06 & \textbf{0.71} & \textbf{8.86} & 0.17 & \textbf{0.32}
        \\
        \quad\textbf{Global map} & \textbf{1.05} & 0.78 & 9.16 & \textbf{0.16} & 2.24
        \\
        \bottomrule
    \end{tabularx}
    \label{tab:computation}
\end{table}

\section{Conclusion}

In this paper, we present 2Fast-2Lamaa, a lidar-inertial mapping and localization framework that consists of two main components: an optimization-based motion-distortion correction module and a scan-to-map localization algorithm that can build maps incrementally.
Given sufficient geometric features in the lidar's surroundings, the undistortion step estimates the continuous trajectory of the sensor through point-to-line and point-to-plane distance minimization between consecutive scans.
The local trajectory is characterized by the inertial data directly through continuous \ac{imu} preintegration.
Thus, the initial conditions (gravity direction, linear velocity, and biases) are the only state variables to estimate, making the problem efficient to solve.
For motion consistency over longer horizons, the scan-to-map registration is performed using \ac{gp}-based distance fields.
Thanks to the use of efficient data structures and locally computed \acp{gp}, the scan alignment can be done in real time.
In odometry/mapping mode, the map is incremented after every scan-to-map registration.
An optional \edit{online} loop-closure detection and correction step, based on a pose-graph optimization, gives 2Fast-2Lamaa the ability to estimate globally consistent trajectories if required.

Throughout an extensive experimental analysis using data from automotive and handheld platforms, 2Fast-2Lamaa demonstrates state-of-the-art accuracy for odometry, localization, and \ac{slam}.
Results were consistent across the different test environments, even with challenging routes going through feature-limited tunnels where baselines' performance significantly declined.
Note that these environments are not fully feature-deprived, as regular features such as service doors are present along the traffic lane.
To address fully feature-deprived scenarios, our future work includes three different directions that are: the integration of intensity information in both the undistortion and localization modules, the fusion with vision sensing, and the exploration of Doppler velocity measurement with \ac{fmcw} lidars.

\bibliographystyle{SageH}
\bibliography{bibliography}

\appendix

\section{Appendix}

\subsection{Distance field uncertainty proxy}
\label{app:uncertainty}

It is important to note that the distance field from previous work \citep{legentil2024accurate} is not a \ac{gp} as it is obtained via non-linear operations over a \ac{gp} latent space.
Accordingly, the naive \ac{gp} variance inference \citep{Rasmussen2006} does not provide satisfying uncertainty information about the distance values $\dist(\mathbf{x})$ away from the surface.
Here, after reviewing the proxy mechanisms previously introduced \citep{legentil2024accurate}, we present a novel uncertainty proxy for the distance field.

\begin{figure}
    \centering
    \def\imgwidth{0.49\columnwidth}
    \def\subfigdist{0.05cm}
    \def\imgdist{0.01\columnwidth}
    \def\vdist{0.6cm}
    \def\subfigstyle{\footnotesize}
    \begin{tikzpicture}
        \tikzstyle{img} = [fill=white, rectangle, align = center, execute at begin node=\setlength{\baselineskip}{8pt}, inner sep=0, outer sep=0]
        \tikzstyle{subfig} = [fill=white, rectangle, align = center, text width = \imgwidth,  minimum width = \imgwidth, execute at begin node=\setlength{\baselineskip}{8pt}, inner sep=0, outer sep=0]

        \node[img](field){\includegraphics[width=\imgwidth,clip]{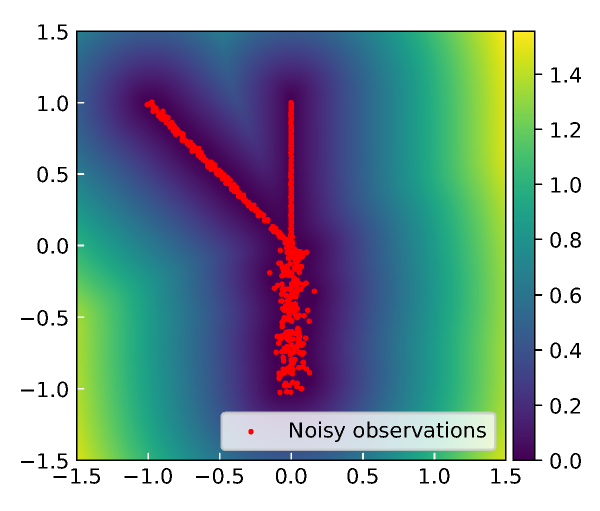}};
        \node[subfig, below=\subfigdist of field]{\subfigstyle(a) Inferred distance [m]};

        \node[img, right=\imgdist of field](error){\includegraphics[width=\imgwidth,clip]{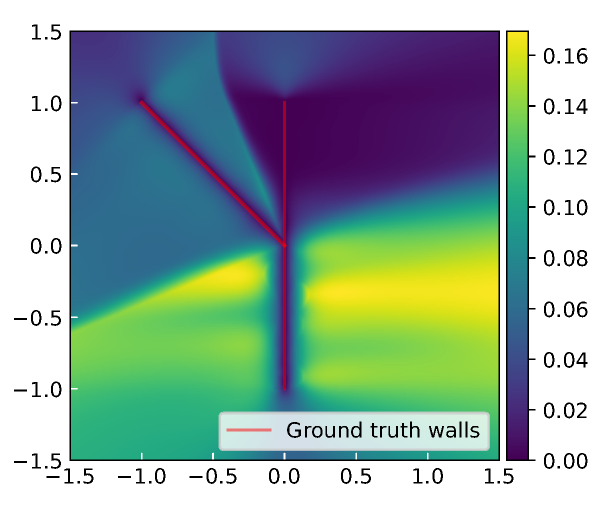}};
        \node[subfig, below=\subfigdist of error]{\subfigstyle(b) Distance field error [m]};
        
        \node[img, below=\vdist of field](old){\includegraphics[width=\imgwidth,clip]{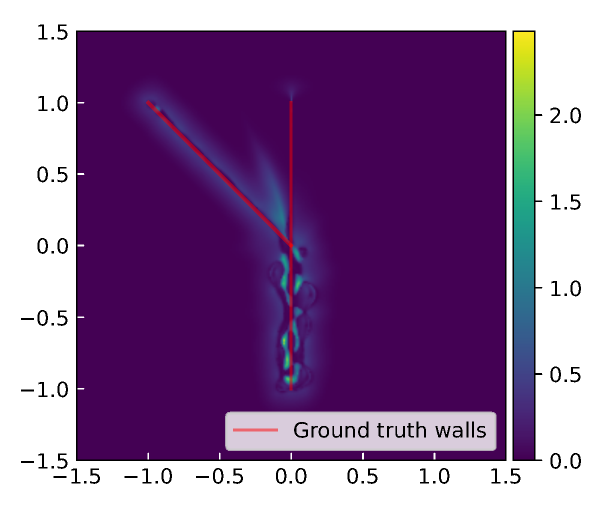}};
        \node[subfig, below=\subfigdist of old]{\subfigstyle(c) Existing uncertainty proxy \citep{legentil2024accurate}};
        
        \node[img, below=\vdist of error](ours){\includegraphics[width=\imgwidth,clip]{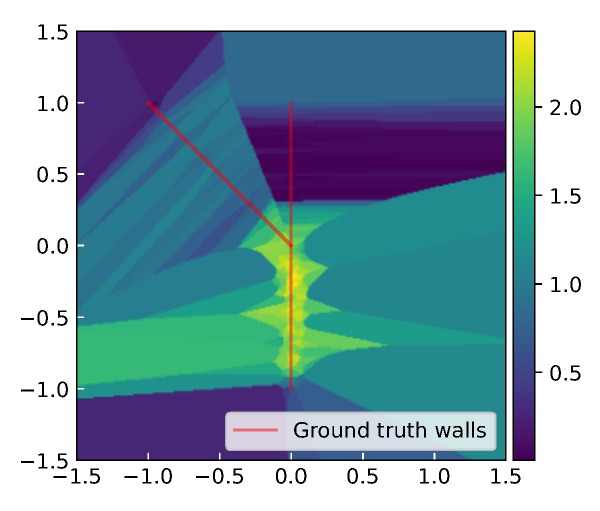}};
        \node[subfig, below=\subfigdist of ours]{\subfigstyle(d) Novel uncertainty proxy};

    \end{tikzpicture}
    \caption{Visualization of the proposed uncertainty proxy when considering non-stationary observation noise. (a) shows the estimated distance field with the observation parameters tuned for the least noisy of the wall observations. (b) shows the error compared with the ground-truth. (c) is the uncertainty proxy introduced in previous work \citep{legentil2024accurate}. (d) is the proposed uncertainty proxy, which patterns match the area of low and high error in (b).}
    \label{figure:uncertainty}
\end{figure}

To account for the uncertainty of the input point cloud, previous work introduced a method for optimizing the observation noise $\sigma$ based on a typical data sample \citep{legentil2024accurate}.
While this is a principled way to address the issue of noisy inputs, it presents limitations.
One is the uniqueness of the noise parameter $\sigma$, which hinders accuracy when considering non-stationary noise (both in time and space).
In realistic lidar-based odometry scenarios, some surfaces will appear `thicker' than others in the incoming point cloud due to various factors such as the noise in the lidar pose estimate, the angle of incidence of the lidar beams, etc.
Regarding the uncertainty of the distance estimates, \cite{legentil2024accurate} provides an elegant uncertainty proxy based on the degree of similarity between the field's gradient and the expected gradient.
However, this mostly addresses the issue of over-interpolation in tight spaces (corridor-like pattern with a width in the order of magnitude of the kernel's lengthscale).
We illustrate the aforementioned drawbacks in Fig.~\ref{figure:uncertainty}(c).
Note that this proxy is not really relevant to the proposed efficient distance field, as the \acp{gp} are computed locally, considering a small neighbourhood, thus rarely presenting corridor-like patterns.

Conceptually, the use of \ac{gp} regression for point-cloud-based distance field estimation does not fit the original purpose of standard \ac{gp} regression: we extrapolate the field far from the data observations, whereas \ac{gp} regression is an interpolation technique.
In other words, we are seeking information and its variance in areas of space far from where the information is actually observed.\footnote {Please note that the GP-inferred variance is still relevant close to the surface and can directly be leveraged in applications such as surface reconstruction.}
The proposed distance uncertainty proxy `fetches' the information where it is relevant for a given distance query by first selecting the closest surface point to the query location (denoted $\spoint(\mathbf{x})$ in the following).
The location of $\spoint(\mathbf{x})$ can easily be deduced from the inferred distance and gradient $\spoint(\mathbf{x}) = \mathbf{x} - \dist(\mathbf{x})\nabla\dist(\mathbf{x})$.
The following step consists of comparing the local shape of the latent field $\occ$ around $\spoint(\mathbf{x})$ with that of a noiseless surface observation.
We propose to use a local integration of the latent field to summarise the local shape of the latent field $\nu(\spoint(\mathbf{x}))=\iiint_{\mathbf{u}\in\mathcal{S} _{\spoint(\mathbf{x})}} \occ(\mathbf{u})d\mathbf{u}$, with $\mathcal{S}_{\spoint(\mathbf{x})}$ a small sphere around $\spoint(\mathbf{x})$.
The integrals can be obtained with linear operators on the \ac{gp} inference.
Eventually, the proxy is defined as $\phi(\mathbf{x}) =|\nu(\spoint(\mathbf{x})) - \nu_{\rm{calib}}|$, where $\nu_{\rm{calib}}$ is computed with noiseless simulated data of a flat wall.
It is equal to zero when the distance prediction is trustworthy and grows when the local shape differs from the ideal noiseless wall data.

The proposed proxy can be computed with the standard global \ac{gp} formulation \citep{legentil2024accurate}, but can also be made efficient with the sparse structure used in this work.
First, the computation of $\spoint(\mathbf{x})$ can be omitted and approximated with the results of the spatial index query that already occurred to find the closest map voxel.
Then, as the \ac{gp} is computed locally, the integral over the sphere around $\spoint(\mathbf{x})$ can be replaced by the integral over the full $\mathbb{R}^3$ space.
As the integral over $\mathbb{R}^3$ of the kernel function is a constant $c$, $\nu(\spoint(\mathbf{x}))\approx c\sum_i \alpha_i$, with $\boldsymbol{\alpha} = \left(\kernelmat{}{\lidarpoints{}{}}{\lidarpoints{}{}} + \sigma\mathbf{I}\right)^{-1} \mathbf{1}$ (already computed in \eqref{eq:occ_inference}).
In the end, the additional computation cost of our novel uncertainty proxy is negligible.
An example of the proposed proxy is given in Fig.~\ref{figure:uncertainty}~(d).
Note that in its current state, the proxy does not directly provide the standard deviation of the inferred distance.
For this reason, we decided not to account for it in 2Fast-2Lamaa's distance-field-based scan-to-map registration. 
Investigating the physical meaning of $\phi$ is outside the scope of this paper and is part of future work.

\subsection{Voxel weighting analysis}
\label{app:field_accuracy}

An appealing aspect of \ac{gp} regression is its ability to handle input noise elegantly.
With the simplifying assumption that lidar points are subject to \ac{iid} spherical noise, considering that the `surface measurements' in~\eqref{eq:occ_inference} are also \ac{iid} is a sensible approach.
The underlying intuition is that every lidar point is as informative as any other lidar point.
However, when voxelizing geometric data for computational efficiency, this assumption of equal information for every input point is not valid.
Each voxel's centroid provided a different amount of information depending on the number of points that occurred in it.
Thus, we presented in Section~\ref{sec:data_structure} a weighting mechanism that adapts the \ac{gp}'s measurement noise model to account for discrepancies in the individual voxel's information.
In this appendix, we provide a brief comparison of accuracy with and without the proposed weighting scheme.
We also compare the results of the proposed distance field against a simple nearest-neighbour search and the dense \ac{gp} formulation from previous work \citep{legentil2024accurate}.

\begin{figure}
    \centering
    \def\imgwidth{0.49\columnwidth}
    \def\subfigdist{0.1cm}
    \def\imgdist{0.01\columnwidth}
    \def\vdist{1.0cm}
    \def\subfigstyle{\footnotesize}
    \begin{tikzpicture}
        \tikzstyle{img} = [fill=white, rectangle, align = center, execute at begin node=\setlength{\baselineskip}{8pt}, inner sep=0, outer sep=0]
        \tikzstyle{subfig} = [fill=white, rectangle, align = center, text width = \imgwidth,  minimum width = \imgwidth, execute at begin node=\setlength{\baselineskip}{8pt}, inner sep=0, outer sep=0]

        \node[img](field){\includegraphics[width=\imgwidth,clip, trim=1.1cm 1.0cm 0.7cm 1.3cm]{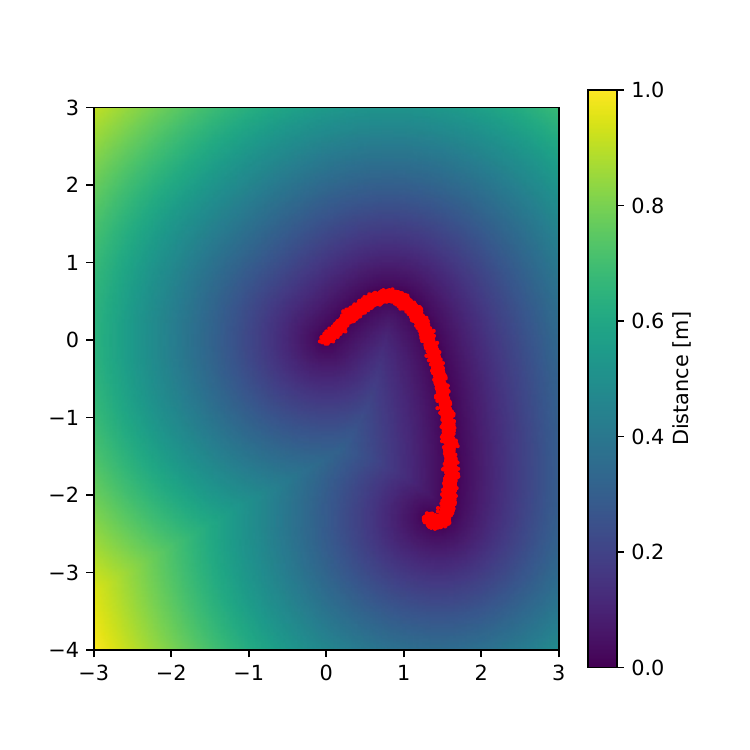}};
        \node[subfig, below=\subfigdist of field]{\subfigstyle(a) Ground-truth distance field and noisy observations};

        \node[img, right=\imgdist of field](errordense){\includegraphics[width=\imgwidth,clip, trim=1.1cm 1.0cm 0.7cm 1.3cm]{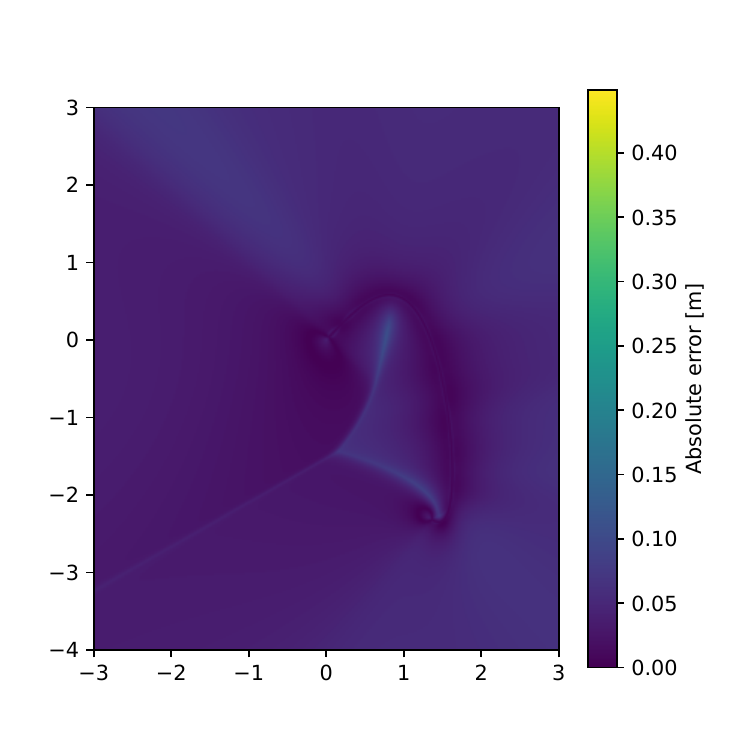}};
        \node[subfig, below=\subfigdist of errordense]{\subfigstyle(b) Absolute inference error dense GP \citep{legentil2024accurate}};
        
        \node[img, below=\vdist of field](errorunweight){\includegraphics[width=\imgwidth,clip, trim=1.1cm 1.0cm 0.7cm 1.3cm]{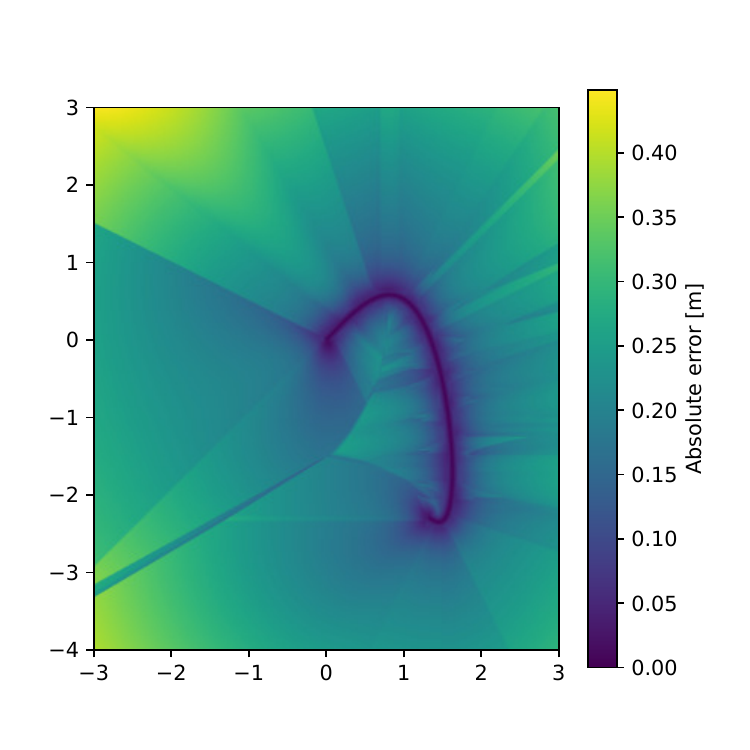}};
        \node[subfig, below=\subfigdist of errorunweight]{\subfigstyle(c) Absolute inference error ours unweighted};
        
        \node[img, below=\vdist of errordense](errorweight){\includegraphics[width=\imgwidth,clip, trim=1.1cm 1.0cm 0.7cm 1.3cm]{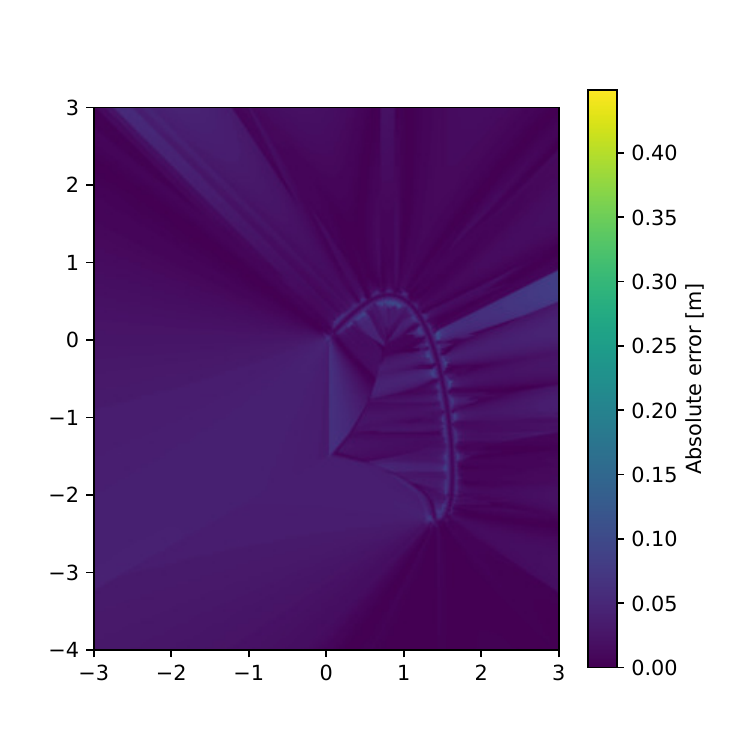}};
        \node[subfig, below=\subfigdist of errorweight]{\subfigstyle(d) Absolute inference error ours weighted};

    \end{tikzpicture}
    \caption{Visualization of the distance field inference error with \emph{dense GP}, and our proposed efficient distance function with and without the proposed weighting mechanism.}
    \label{fig:app_gp_weight}
\end{figure}

Our 2D simulated setup emulates lidar points collected along a randomly-shaped wall.
The wall is defined as a sum of sine functions.
A \SI{3}{\cm} spherical Gaussian noise is applied to each of the 10k points of the scene.
Fig.~\ref{fig:app_gp_weight}(a) shows one of the randomly generated scenes.
The distance field's ground-truth is computed based on a nearest-neighbour search with high-density noiseless wall measurements.
Table~\ref{tab:app_gp_weight} shows the average distance \ac{rmse} over 100 runs.
Note that the parameters of the \emph{Dense GP} version are tuned specifically for the simulated density and noise distribution.
The \emph{unweighted} (considering each voxel equally informative with an \ac{iid} observation noise) and \emph{weighted} versions share the same generic parameters (cell size, lengthscale, etc) to highlight the benefits of the proposed weighting mechanism.
A smaller error could be achieved with the \emph{unweighted} variant by specifically tuning the parameter to the point density and noise distribution of this setup, but losing genericity.
As illustrated in the error plot in Fig.~\ref{fig:app_gp_weight}, the \emph{unweighted} approach creates a virtually `thicker' surface, offsetting the distance values inferred away from the wall.
Interestingly, as the proposed efficient distance field method considers small local neighbourhoods to compute the \acp{gp}, the method is somewhat unaffected by the phenomenon of over-interpolation in narrow spaces that degrades the performance of the \emph{dense} formulation.
Accordingly, our method performs the best, at least on this simulated setup.
As expected, the proposed approach performs significantly better than a simple nearest-neighbour search (\emph{Voxel NN}), which correlates with the final odometry findings in our ablation study (Section~\ref{sec:exp_ablation}).

\begin{table}
    \centering
    \caption{Average RMSE error of the proposed efficient distance field and various baselines.}
    \setlength{\tabcolsep}{2pt}
    \begin{tabularx}{\linewidth}{lYY}
        \toprule
        \textbf{Variant} & \textbf{Average RMSE [m]} & \textbf{Std. [m]}
        \\
        \midrule
        \textbf{Voxel NN} & 0.072 & 0.006
        \\
        \textbf{Dense GP} \scriptsize\citep{legentil2024accurate} & 0.047 & 0.007
        \\
        \textbf{Ours unweighted} & 0.252 & 0.019
        \\
        \textbf{Ours weighted} & \textbf{0.029} & 0.006
        \\
        \bottomrule
    \end{tabularx}
    \label{tab:app_gp_weight}
\end{table}

\subsection{Spatial index performance}
\label{app:phtree}

\begin{figure}
    \centering
    \includegraphics[width=0.99\columnwidth]{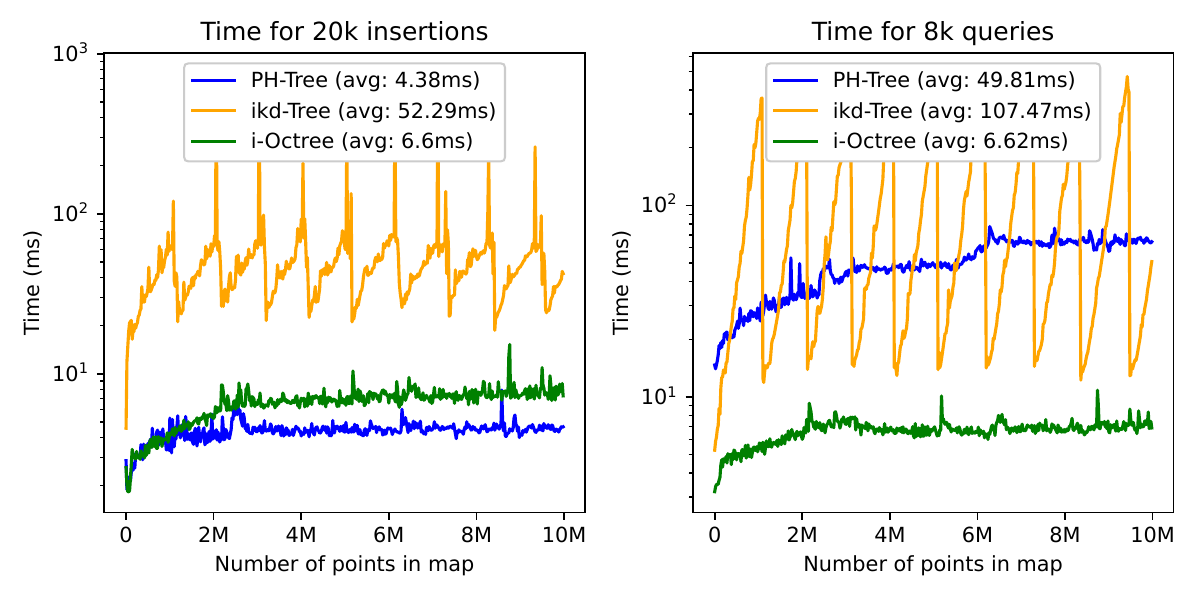}
    \caption{Timing comparison, using a log-scale, between ikd-Tree, PH-Tree, and i-Octree for point insertion and nearest neighbour search in a growing map (using scans made of random points in a virtual \SI{5}{\km} straight line trajectory).}
    \label{fig:spatial_index}
\end{figure}

In this appendix, we benchmark spatial indexes that fulfill 2Fast-2Lamaa's requirements, which are the ability to add or remove points without rebuilding the index from scratch, allowing k-nearest-neighbour and radius searches, and scalability with over 10 million points in the index.
We use \edit{three} data structures with publicly available C++ implementations: ikd-Trees \citep{cai2021ikd}, PH-Trees \citep{zaschke2014phtree}, \edit{and i-Octrees \citep{zhu2024ioctree}.}
Our toy example emulates 2Fast-2Lamaa's use case in driving scenarios with an incrementally built map of 10 million points created along a \SI{5}{\km} trajectory.
We simulate the lidar collection with `scans' of 20k points at random locations within the lidar range (\SI{100}{\m}).
Every time a scan is inserted in the spatial index, 8k nearest-neighbour queries are performed to imitate the registration process.
Dividing the goal of 10 million points in the final map by the size of the scans, we obtain 500 steps of \SI{10}{\m} each.
The trajectory is defined as a straight line parameterized by $x=y$ and $z=0$.
Fig.~\ref{fig:spatial_index} shows the timing obtained with the three methods using a single core of our Intel i7-1370p CPU.\footnote{Code available here: \url{https://github.com/clegenti/spatial_index_benchmark}}
The results show an advantage for the PH-Tree \edit{for insertion, but the i-Octree clearly outperforms the other two methods for queries}.
Note that using multi-threading on the queries can greatly reduce the computation time \edit{of i-Octree queries down to \SI{1.51}{\ms} (with 8 cores).
The jagged nature of the ikd-Tree reflects its need for regular balancing.}

\edit{
\subsection{Entire Boreas-RT dataset results}
\label{app:boreasrt}
This appendix provides the odometry and localization results obtained with 2Fast-2Lamaa on the entirety of the Boreas-RT dataset~\citep{lisus2026boreasrt}.

\subsubsection{Odometry}
Tables~\ref{tab:boreasrt_odom_structured},~\ref{tab:boreasrt_odom_rural}, and~\ref{tab:boreasrt_odom_freeway} corresponds to the \emph{Structured}, \emph{Rural}, and \emph{Regional} sequence types, respectively.
In total, nearly \SI{650}{\km} of data is used to generate these results.
2Fast-2Lamaa robustly provides high-accuracy estimates, consistently outperforming LTR~\citep{burnett2022arewe}.

\begin{table}
    \centering
    \caption{Per-sequence odometry results on the \emph{Structured} sequences of the Boreas-RT dataset.}
    \setlength{\tabcolsep}{2pt}
    \begin{tabularx}{\linewidth}{lYYY}
\label{tab:boreasrt_odom_structured}\\
\toprule
\textbf{Sequence type} & \textbf{ID} & \textbf{LTR w/ gyro} & \textbf{2Fast-2Lamaa} \\
\midrule

\texttt{suburbs} & \scriptsize{2024-12-03-12-54} & 0.30/0.10 & 0.24/0.08 \\
& \scriptsize{2024-12-05-14-25} & 0.29/0.09 & 0.21/0.08 \\
& \scriptsize{2025-01-08-10-59} & 0.30/0.09 & 0.23/0.08 \\
& \scriptsize{2025-01-08-11-22} & 0.28/0.09 & 0.21/0.07 \\
& \scriptsize{2025-01-08-12-28} & 0.26/0.08 & 0.21/0.08 \\
& \scriptsize{2025-02-15-16-58} & 0.27/0.09 & 0.20/0.07 \\
& \scriptsize{2025-02-15-17-19} & 0.27/0.09 & 0.22/0.08 \\
& \scriptsize{2025-02-21-14-51} & 0.34/0.10 & 0.24/0.08 \\
& \scriptsize{2025-02-22-11-32} & 0.29/0.09 & 0.25/0.09 \\
& \scriptsize{2025-02-22-12-26} & 0.30/0.10 & 0.22/0.08 \\
\cmidrule{2-4}
& \textbf{Average} & 0.29/0.09 & \textbf{0.22/0.08} \\
\midrule
\texttt{industrial} 
& \scriptsize{2024-12-05-14-12} & 0.23/0.08 & 0.21/0.08 \\
& \scriptsize{2024-12-23-16-27} & 0.27/0.10 & 0.21/0.08 \\
& \scriptsize{2024-12-23-16-44} & 0.24/0.09 & 0.20/0.08 \\
& \scriptsize{2024-12-23-17-01} & 0.26/0.09 & 0.19/0.07 \\
& \scriptsize{2024-12-23-17-18} & 0.24/0.09 & 0.19/0.07 \\
\cmidrule{2-4}
& \textbf{Average} & 0.25/0.09 & \textbf{0.20/0.07} \\
\midrule
\texttt{urban} 
& \scriptsize{2025-08-06-06-33} & 0.41/0.16 & 0.35/0.11 \\
& \scriptsize{2025-08-06-07-05} & 0.46/0.17 & 0.40/0.11 \\
& \scriptsize{2025-08-06-07-41} & 0.43/0.17 & 0.37/0.11 \\
& \scriptsize{2025-08-06-08-35} & 0.55/0.22 & 0.53/0.11 \\
& \scriptsize{2025-08-06-10-48} & 0.48/0.21 & 0.31/0.10 \\
& \scriptsize{2025-08-06-11-32} & 0.57/0.23 & 0.39/0.10 \\
& \scriptsize{2025-08-06-12-20} & 0.51/0.24 & 0.29/0.09 \\
\cmidrule{2-4}
& \textbf{Average} & 0.49/0.20 & \textbf{0.38/0.10} \\
\bottomrule
\multicolumn{4}{l}{\scriptsize KITTI odometry metric reported as \textit{XX / YY} with \textit{XX} [\%] and \textit{YY} [$\si{\degree}/100\,\si{\m}$]}
\vspace{-0.15cm}
\\
\multicolumn{4}{l}{\scriptsize the translation and orientation errors, respectively.}
\end{tabularx}
\end{table}

\begin{table}
    \centering
    \caption{Per-sequence odometry results on the \emph{Rural} sequences of the Boreas-RT dataset.}
    \setlength{\tabcolsep}{2pt}
    \begin{tabularx}{\linewidth}{lYYY}
\label{tab:boreasrt_odom_rural}\\
\toprule
\textbf{Sequence type} & \textbf{ID} & \textbf{LTR w/ gyro} & \textbf{2Fast-2Lamaa} \\
\midrule

\texttt{forest} 
& \scriptsize{2025-07-18-10-33} & 0.66/0.14 & 0.46/0.18 \\
& \scriptsize{2025-07-18-11-00} & 0.71/0.10 & 0.46/0.18 \\
& \scriptsize{2025-07-18-11-25} & 0.64/0.10 & 0.48/0.18 \\
& \scriptsize{2025-07-18-11-53} & 0.78/0.10 & 0.47/0.19 \\
\cmidrule{2-4}
& \textbf{Average} & 0.70/\textbf{0.11} & \textbf{0.47}/0.18 \\
\midrule

\texttt{farm} 
& \scriptsize{2025-07-18-14-55} & 0.45/0.11 & 0.35/0.10 \\
& \scriptsize{2025-07-18-15-12} & 0.44/0.10 & 0.37/0.10 \\
& \scriptsize{2025-07-18-15-30} & 0.42/0.11 & 0.38/0.10 \\
& \scriptsize{2025-07-18-15-48} & 0.60/0.11 & 0.39/0.11 \\
& \scriptsize{2025-07-18-16-05} & 0.49/0.10 & 0.40/0.11 \\
& \scriptsize{2025-08-13-09-01} & 0.40/0.13 & 0.33/0.10 \\
& \scriptsize{2025-08-13-09-21} & 0.40/0.13 & 0.34/0.09 \\
& \scriptsize{2025-08-13-09-46} & 0.42/0.13 & 0.34/0.09 \\
& \scriptsize{2025-08-13-10-12} & 0.42/0.13 & 0.35/0.10 \\
& \scriptsize{2025-08-13-10-36} & 0.40/0.12 & 0.34/0.10 \\
\cmidrule{2-4}
& \textbf{Average} & 0.44/0.12 & \textbf{0.36/0.10} \\
\bottomrule
\multicolumn{4}{l}{\scriptsize KITTI odometry metric reported as \textit{XX / YY} with \textit{XX} [\%] and \textit{YY} [$\si{\degree}/100\,\si{\m}$]}
\vspace{-0.15cm}
\\
\multicolumn{4}{l}{\scriptsize the translation and orientation errors, respectively.}
\end{tabularx}
\end{table}

\begin{table}
    \centering
    \caption{Per-sequence odometry results on the \emph{Highway} sequences of the Boreas-RT dataset.}
    \setlength{\tabcolsep}{2pt}
    \begin{tabularx}{\linewidth}{lYYY}
\label{tab:boreasrt_odom_freeway}\\
\toprule
\textbf{Sequence type} & \textbf{ID} & \textbf{LTR w/ gyro} & \textbf{2Fast-2Lamaa} \\
\midrule
\texttt{regional} 
& \scriptsize{2024-12-03-13-13} & 0.34/0.09 & 0.18/0.07 \\
& \scriptsize{2024-12-03-13-34} & 0.34/0.09 & 0.20/0.08 \\
& \scriptsize{2024-12-10-12-07} & 0.34/0.09 & 0.20/0.08 \\
& \scriptsize{2024-12-10-12-24} & 0.35/0.10 & 0.21/0.08 \\
& \scriptsize{2024-12-10-12-38} & 0.32/0.08 & 0.18/0.08 \\
& \scriptsize{2024-12-10-12-56} & 0.38/0.10 & 0.21/0.08 \\
\cmidrule{2-4}
& \textbf{Average} & 0.35/0.09 & \textbf{0.20/0.08} \\
\midrule

\texttt{tunnel}
& \scriptsize{2024-12-04-14-28} & 2.26/0.12 & 0.27/0.08 \\
& \scriptsize{2024-12-04-14-34} & X         & 0.60/0.17 \\
& \scriptsize{2024-12-04-14-38} & 2.82/0.11 & 0.29/0.10 \\
& \scriptsize{2024-12-04-14-44} & 1.69/0.08 & 0.40/0.14 \\
& \scriptsize{2024-12-04-14-50} & X         & 0.30/0.10 \\
& \scriptsize{2024-12-04-14-59} & 2.05/0.09 & 0.37/0.11 \\
& \scriptsize{2024-12-04-15-04} & 2.36/0.11 & 0.26/0.09 \\
& \scriptsize{2024-12-04-15-10} & 1.02/0.09 & 0.33/0.11 \\
& \scriptsize{2024-12-04-15-19} & 2.87/0.11 & 0.28/0.09 \\
& \scriptsize{2024-12-04-15-24} & 1.41/0.09 & 0.39/0.14 \\
\cmidrule{2-4}
& \textbf{Average} & 2.06/\textbf{0.10} & \textbf{0.35}/0.11 \\
\midrule

\texttt{skyway} 
& \scriptsize{2024-12-04-11-45} & 1.62/0.08 & 0.30/0.10 \\
& \scriptsize{2024-12-04-11-56} & 0.47/0.07 & 0.32/0.10 \\
& \scriptsize{2024-12-04-12-08} & 0.63/0.08 & 0.26/0.09 \\
& \scriptsize{2024-12-04-12-19} & 0.51/0.08 & 0.34/0.10 \\
& \scriptsize{2024-12-04-12-34} & 0.60/0.08 & 0.27/0.09 \\
\cmidrule{2-4}
& \textbf{Average} & 0.77/\textbf{0.08} & \textbf{0.30}/0.09 \\
\midrule

\texttt{freeway} 
& \scriptsize{2025-07-18-16-24} & 0.53/0.13 & 0.26/0.10 \\
& \scriptsize{2025-08-13-07-54} & 0.42/0.11 & 0.20/0.07 \\
& \scriptsize{2025-08-13-11-52} & 0.44/0.13 & 0.26/0.10 \\
\cmidrule{2-4}
& \textbf{Average} & 0.46/0.12 & \textbf{0.24/0.09} \\
\bottomrule
\multicolumn{4}{l}{\scriptsize KITTI odometry metric reported as \textit{XX / YY} with \textit{XX} [\%] and \textit{YY} [$\si{\degree}/100\,\si{\m}$]}
\vspace{-0.15cm}
\\
\multicolumn{4}{l}{\scriptsize the translation and orientation errors, respectively.}

\end{tabularx}
\end{table}

\subsubsection{Localization}
The localization results are provided in Table~\ref{tab:boreasrt_loc}.
Note that the \emph{Forest} and \emph{Urban} routes are omitted as the ground-truth trajectories are not sufficiently consistent between repeated sequences to enable centimetre-level localization evaluation.
As explained by~\cite{lisus2026boreasrt}, a lack of unobstructed view of the sky, due to trees or high buildings, and the distance to the RTK base station lead to larger errors in the sequence-to-sequence ground-truth accuracy (greater than \SI{0.4}{\m}).

2Fast-2Lamaa and LTR (see~\cite{lisus2026boreasrt} for results) provide similar levels of accuracy and robustness in structured environments such as \emph{Suburbs}, \emph{Industrial}, and \emph{Regional}.
However, in more challenging environments, LTR reports larger errors and even fails on some sequences.
The variability of the vertical accuracy on the \emph{Suburbs} and \emph{Farm} is explained by significant changes in the environment between the time of mapping and the time of localization.
The former is due to large snowbanks that were created by snowstorms.
The latter is due to the vegetation in the field that grew during the month separating the two data-collection days.
Overall, 2Fast-2Lamaa provides an average RMSE lateral error under \SI{0.05}{\m}.

\begin{table*}
    \centering
    \caption{Per-sequence RMSE localization results for 2Fast-2Lamaa on the Boreas-RT dataset.}
    \setlength{\tabcolsep}{2pt}
    \begin{tabularx}{\linewidth}{lcYYYYYY}
        \toprule
        \textbf{Seq. type \& map seq.} & \textbf{Seq. ID} & \textbf{Long.} \scriptsize[m] & \textbf{Lat.} \scriptsize[m] & \textbf{Vert.} \scriptsize[m] & \textbf{Roll} \scriptsize[$^\circ$] & \textbf{Pitch} \scriptsize[$^\circ$] & \textbf{Yaw} \scriptsize[$^\circ$] 
        \\
        \midrule
\textbf{Suburbs}    & 2024-12-05-14-25 & 0.031 & 0.023 & 0.041 & 0.039 & 0.032 & 0.030 \\
2024-12-03-12-54    & 2025-01-08-10-59 & 0.027 & 0.024 & 0.040 & 0.042 & 0.029 & 0.029 \\
                    & 2025-01-08-11-22 & 0.024 & 0.021 & 0.025 & 0.038 & 0.028 & 0.025 \\
                    & 2025-01-08-12-28 & 0.026 & 0.025 & 0.057 & 0.045 & 0.031 & 0.027 \\
                    & 2025-02-15-16-58 & 0.025 & 0.027 & 0.230 & 0.078 & 0.040 & 0.030 \\
                    & 2025-02-15-17-19 & 0.025 & 0.031 & 0.234 & 0.078 & 0.044 & 0.032 \\
                    & 2025-02-21-14-51 & 0.030 & 0.028 & 0.280 & 0.082 & 0.052 & 0.030 \\
                    & 2025-02-22-11-32 & 0.030 & 0.027 & 0.278 & 0.084 & 0.047 & 0.029 \\
                    & 2025-02-22-12-26 & 0.032 & 0.026 & 0.278 & 0.083 & 0.067 & 0.028 \\
\cmidrule{2-8}
& \textbf{Average} & 0.028 & 0.026 & 0.163 & 0.063 & 0.041 & 0.029 \\

\midrule
\textbf{Regional north} & 2024-12-10-12-07 & 0.032 & 0.028 & 0.031 & 0.037 & 0.026 & 0.030 \\
2024-12-03-13-13        & 2024-12-10-12-38 & 0.035 & 0.043 & 0.041 & 0.038 & 0.033 & 0.030 \\
\cmidrule{2-8}
& \textbf{Average} & 0.034 & 0.035 & 0.036 & 0.038 & 0.030 & 0.030 \\

\midrule
\textbf{Regional south} & 2024-12-10-12-24 & 0.031 & 0.026 & 0.046 & 0.035 & 0.026 & 0.025 \\
2024-12-03-13-34        & 2024-12-10-12-56 & 0.031 & 0.029 & 0.034 & 0.041 & 0.028 & 0.029 \\
\cmidrule{2-8}
& \textbf{Average} & 0.031 & 0.028 & 0.040 & 0.038 & 0.027 & 0.027 \\

\midrule
\textbf{Tunnel east}    & 2024-12-04-14-38 & 0.066 & 0.025 & 0.060 & 0.077 & 0.025 & 0.031 \\
2024-12-04-14-28        & 2024-12-04-14-50 & 0.119 & 0.028 & 0.032 & 0.089 & 0.024 & 0.029 \\
                        & 2024-12-04-15-04 & 0.112 & 0.030 & 0.046 & 0.087 & 0.026 & 0.032 \\
                        & 2024-12-04-15-19 & 0.095 & 0.025 & 0.046 & 0.080 & 0.024 & 0.029 \\
\cmidrule{2-8}
& \textbf{Average} & 0.098 & 0.027 & 0.046 & 0.083 & 0.024 & 0.030 \\

\midrule
\textbf{Tunnel west}    & 2024-12-04-14-34 & 0.136 & 0.024 & 0.043 & 0.150 & 0.028 & 0.029 \\
2024-12-04-14-44        & 2024-12-04-14-59 & 0.230 & 0.028 & 0.022 & 0.081 & 0.020 & 0.028 \\
                        & 2024-12-04-15-10 & 0.079 & 0.043 & 0.061 & 0.084 & 0.024 & 0.038 \\
                        & 2024-12-04-15-24 & 0.176 & 0.053 & 0.039 & 0.113 & 0.041 & 0.033 \\
\cmidrule{2-8}
& \textbf{Average} & 0.155 & 0.037 & 0.041 & 0.107 & 0.028 & 0.032 \\

\midrule
\textbf{Skyway} & 2024-12-04-11-56 & 0.126 & 0.047 & 0.050 & 0.186 & 0.040 & 0.083 \\
2024-12-04-11-45 & 2024-12-04-12-08 & 0.062 & 0.035 & 0.052 & 0.058 & 0.038 & 0.031 \\
& 2024-12-04-12-19 & 0.074 & 0.038 & 0.047 & 0.067 & 0.039 & 0.035 \\
& 2024-12-04-12-34 & 0.067 & 0.035 & 0.044 & 0.062 & 0.038 & 0.040 \\
\cmidrule{2-8}
& \textbf{Average} & 0.082 & 0.038 & 0.048 & 0.093 & 0.039 & 0.048 \\

\midrule
\textbf{Industrial} & 2024-12-23-16-27 & 0.024 & 0.023 & 0.027 & 0.063 & 0.040 & 0.034 \\
2024-12-05-14-12 & 2024-12-23-16-44 & 0.023 & 0.022 & 0.038 & 0.066 & 0.039 & 0.030 \\
& 2024-12-23-17-01 & 0.023 & 0.022 & 0.038 & 0.063 & 0.040 & 0.038 \\
& 2024-12-23-17-18 & 0.021 & 0.021 & 0.036 & 0.065 & 0.042 & 0.034 \\
\cmidrule{2-8}
& \textbf{Average} & 0.023 & 0.022 & 0.035 & 0.064 & 0.040 & 0.034 \\

\midrule
\textbf{Farm} & 2025-07-18-15-12 & 0.046 & 0.040 & 0.058 & 0.052 & 0.035 & 0.033 \\
2025-07-18-14-55 & 2025-07-18-15-30 & 0.045 & 0.040 & 0.063 & 0.054 & 0.036 & 0.030 \\
& 2025-07-18-15-48 & 0.048 & 0.044 & 0.071 & 0.052 & 0.039 & 0.038 \\
& 2025-07-18-16-05 & 0.049 & 0.041 & 0.068 & 0.053 & 0.037 & 0.033 \\
& 2025-08-13-09-01 & 0.085 & 0.068 & 0.169 & 0.175 & 0.060 & 0.069 \\
& 2025-08-13-09-21 & 0.079 & 0.073 & 0.165 & 0.167 & 0.057 & 0.042 \\
& 2025-08-13-09-46 & 0.085 & 0.067 & 0.176 & 0.173 & 0.058 & 0.047 \\
& 2025-08-13-10-12 & 0.079 & 0.060 & 0.187 & 0.174 & 0.057 & 0.049 \\
& 2025-08-13-10-36 & 0.071 & 0.059 & 0.185 & 0.174 & 0.056 & 0.062 \\
\cmidrule{2-8}
& \textbf{Average} & 0.065 & 0.055 & 0.127 & 0.119 & 0.048 & 0.045 \\

\midrule
\textbf{Freeway} & 2025-08-13-11-52 & 0.052 & 0.038 & 0.082 & 0.067 & 0.033 & 0.037 \\
\cmidrule{2-8}
2025-07-18-16-24 & \textbf{Average} & 0.052 & 0.038 & 0.082 & 0.067 & 0.033 & 0.037\\
        \bottomrule
    \end{tabularx}
    \label{tab:boreasrt_loc}
\end{table*}

}

\end{document}